%% file: main.tex
\documentclass[runningheads]{llncs}

\usepackage{eccv}

\usepackage{eccvabbrv}

\usepackage{graphicx}
\usepackage{booktabs}
\usepackage{multirow}
\usepackage{adjustbox} 

\usepackage[accsupp]{axessibility}
\usepackage[english]{babel}
\usepackage[T1]{fontenc}
\usepackage[utf8]{inputenc}
\usepackage[final]{microtype}
\usepackage{hyphenat}
\usepackage{enumitem}
\usepackage[all]{nowidow}
\usepackage{fontawesome5}
\usepackage{amsmath}
\usepackage{amssymb}
\usepackage{amsfonts}
\usepackage{xfrac}

\usepackage{tabularx}
\usepackage{tikz}

\usepackage[labelsep=period,labelfont=bf,font={small}]{caption}
\usepackage{hyperref}

\hypersetup{pdfcreator={Apple}}
\hypersetup{pdfproducer={Apple}}
\hypersetup{pdftitle={Large-Scale High-Quality 3D Gaussian Head Reconstruction from Multi-View Captures}}
\hypersetup{pdfsubject={ECCV 2026}}
\hypersetup{pdfkeywords={ECCV, Machine Learning, Computer Vision, 3D Reconstruction, Gaussian Splatting}}

\usepackage{float}
\usepackage{graphicx}
\usepackage[table]{xcolor}
\usepackage{longtable}
\usepackage{soul}
\usepackage{subcaption}
\usepackage{lipsum}
\usepackage{siunitx}
\usepackage{wrapfig}                                                                        

\usepackage[capitalize]{cleveref}
\crefname{section}{Sec.}{Secs.}
\Crefname{section}{Section}{Sections}
\crefname{table}{Tab.}{Tabs.}
\Crefname{table}{Table}{Tables}

\definecolor{highlightcolor}{RGB}{255,237,191}
\definecolor{textboxcolor}{RGB}{155,155,155}
\definecolor{paperboxcolor}{RGB}{204,148,73}
\definecolor{highlightred}{RGB}{224,103,103}
\definecolor{highlightblue}{RGB}{112,211,255}
\definecolor{highlightgreen}{RGB}{209,255,189}
\definecolor{highlightpurple}{RGB}{218,177,218}

\definecolor{tableheadergray}{RGB}{240,240,240}
\definecolor{tablecellgreen}{RGB}{202, 242, 199}

\makeatletter
\def\thanks#1{\protected@xdef\@thanks{\@thanks\protect\footnotetext{#1}}}
\makeatother

\makeatletter
\renewcommand\subsubsection{\@startsection{subsubsection}{3}{\z@}%
  {-8\p@ \@plus -4\p@ \@minus -4\p@}%
  {-0.5em \@plus -0.22em \@minus -0.1em}%
  {\normalfont\normalsize\bfseries\boldmath}}
\makeatother

\let\titleold\title
\renewcommand{\title}[1]{\titleold{#1}\newcommand{\thetitle}{#1}}
\makeatletter
\def\maketitlesupplementary{%
  \newpage
  \newcounter{savedfigure}\setcounter{savedfigure}{\value{figure}}%
  \newcounter{savedtable}\setcounter{savedtable}{\value{table}}%
  \newcounter{savedsection}\setcounter{savedsection}{\value{section}}%
  \newcounter{savedequation}\setcounter{savedequation}{\value{equation}}%
  \ifeccv@review
    \title{\thetitle{}}%
    \author{Anonymous ECCV \eccv@year{} Supplementary}%
    \titlerunning{ECCV \eccv@year{} Supplementary \#\eccv@ID}%
    \authorrunning{ECCV \eccv@year{} Supplementary \#\eccv@ID}%
    \institute{Paper ID \#\eccv@ID}%
    \maketitleold
  \else
    \title{\thetitle{}}%
    \author{Supplementary}%
    \maketitleold
  \fi
  \setcounter{figure}{\value{savedfigure}}%
  \setcounter{table}{\value{savedtable}}%
  \setcounter{section}{\value{savedsection}}%
  \setcounter{equation}{\value{savedequation}}%
}
\makeatother

\begin{document}

\titlerunning{HeadsUp: Large-scale Gaussian Head Reconstruction}
\title{Large-Scale High-Quality 3D Gaussian Head Reconstruction from Multi-View Captures}

\author{
  Evangelos Ntavelis %
  \and
  Sean Wu$^{\dagger}$\thanks{$^\dagger$Work done during internship at Apple.} \and
  Mohamad Shahbazi \and
  Fabio Maninchedda \and
  Dmitry Kostiaev \and
  Artem Sevastopolsky \and
  Vittorio Megaro \and
  Trevor Phillips \and
  Alejandro Blumentals \and
  Shridhar Ravikumar \and
  Mehak Gupta \and
  Reinhard Knothe \and
  Jeronimo Bayer \and
  Matthias Vestner \and
  Simon Schaefer \and
  Thomas Etterlin \and
  Christian Zimmermann \and
  Alexey Artemov \and
  Mathias Deschler \and
  Peter Kaufmann \and
  Stefan Brugger \and
  Sebastian Martin \and
  Brian Amberg \and
  Tom Runia 
}

\authorrunning{E. Ntavelis et al.}
\institute{Apple \\[0.2cm] \texttt{\href{https://apple.github.io/ml-headsup/}{\textcolor{MidnightBlue}{https://apple.github.io/ml-headsup/}}}}

\vspace{-1em} %
\maketitle
\begin{figure*}[t]
  \centering
\includegraphics[width=\linewidth,trim=0 20cm 1.5cm 0,clip]{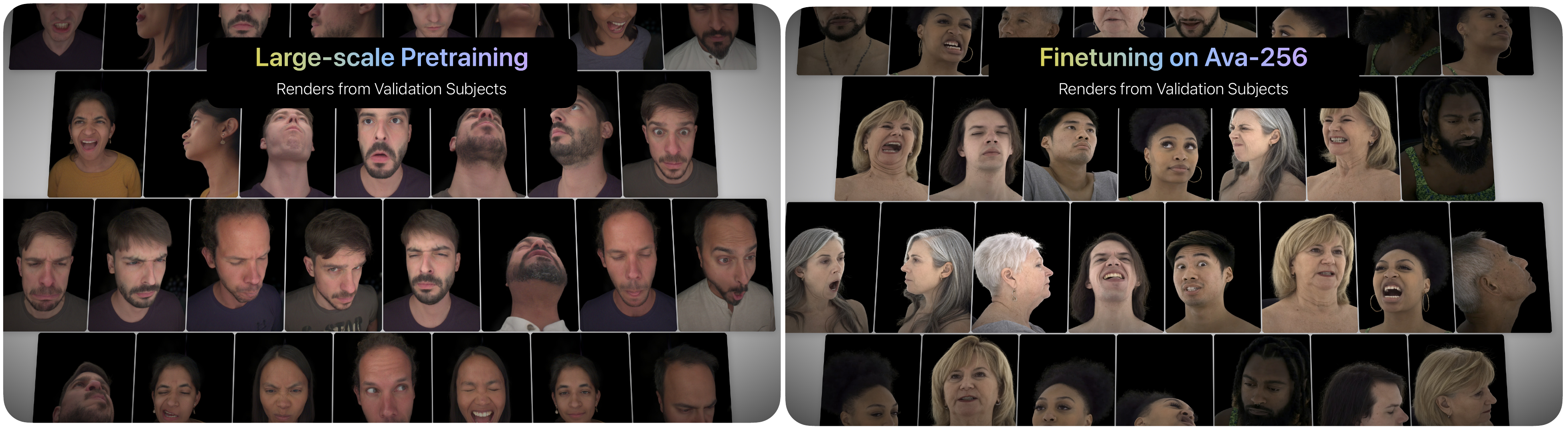}
 \caption{We introduce \emph{HeadsUp}, a novel feed-forward approach leveraging 3D Gaussians to predict high-quality avatars. By scaling to thousands of subjects and diverse expressions, our method achieves exceptional rendering quality on completely held-out subjects. Notice the accurate, high-resolution recovery of intricate fine details, such as eyelashes, complex earrings, teeth and tongue. The figure displays renders from novel subjects unseen during training.}
  \label{fig:teaser}
\end{figure*}

\begin{abstract}
  \input{00_abstract}
\end{abstract}

\input{01_introduction}

\input{02_related_work}

\input{03_method}

\input{04_experiments}

\input{05_conclusion}

\section*{Acknowledgements}
We thank Simon Biland, Armin Kappeler and Rick Zhang for their support and valuable feedback.

\bibliographystyle{splncs04}
\bibliography{main}

\input{06_supplementary}

\end{document}

%% file: 00_abstract.tex
We propose \textit{HeadsUp}, a scalable feed-forward method for reconstructing high-quality 3D Gaussian heads from large-scale multi-camera setups. Our method employs an efficient encoder-decoder architecture that compresses input views into a compact latent representation. This latent representation is then decoded into a set of UV-parameterized 3D Gaussians anchored to a neutral head template. This UV representation decouples the number of 3D Gaussians from the number and resolution of input images, enabling training with many high-resolution input views. We train and evaluate our model on an internal dataset with more than \num{10000} subjects, which is an order of magnitude larger than existing multi-view human head datasets. \textit{HeadsUp} achieves state-of-the-art reconstruction quality and generalizes to novel identities without test-time optimization. We extensively analyze the scaling behavior of our model across identities, views, and model capacity, revealing practical insights for quality-compute trade-offs. Finally, we highlight the strength of our latent space by showcasing two downstream applications: generating novel 3D identities and animating the 3D heads with expression blendshapes. %

%% file: 01_introduction.tex
\section{Introduction\label{sec:introduction}}

High-fidelity 3D head assets are foundational to photorealistic digital humans, enabling convincing and authentic social co-presence in immersive digital environments. Such assets are particularly valuable for view-consistent, close-up rendering in applications such as telepresence, actor digitization, and content creation~\cite{Lawrence2021}. To achieve this level of quality, multi-camera capture systems have become a standard approach for producing dense calibrated images of human heads~\cite{debevec_acquiring_2000}. However, it remains a challenge to reliably and efficiently transform these detailed captures into compact 3D reconstructions at scale~\cite{kirschstein2025avat3r,ye2024real3dportrait}.

Existing solutions for converting multi-view head captures into renderable assets span a spectrum that reflects a recurring tension between reconstruction fidelity and throughput. At one end, instance-specific optimization methods, such as per-subject fitting using Neural Radiance Fields~\cite{mildenhall2020nerf,kirschstein2023nersemble,buhler2024cafca_siggraphasia} or 3D Gaussian Splatting (3DGS)~\cite{kerbl3Dgaussians,saito2024rgca}, serve as strong baselines for high-quality view-consistent rendering. However, the computational cost of per-identity optimization makes their large-scale deployment challenging. At the other end, recent feed-forward reconstruction methods~\cite{kirschstein2025avat3r,charatan2024pixelsplat,chen2024mvsplat,gpsGaussian} amortize computation across datasets and enable fast inference, but their compute and memory costs typically scale with the number and resolution of input views. This hinders the full utilization of dense, high-resolution capture setups. Orthogonally, another line of work targets animatable avatars by predicting geometry and appearance in a canonical, model-conditioned space~\cite{chu2024gagavatar,ROME,ye2024real3dportrait}. While this supports temporal coherence and control, it often sacrifices the representation capacity needed for high-quality rendering. Together, these trends suggest a clear need for methods that can fully leverage multi-camera rigs (i.e., many views, high resolution, many identities) while producing compact head assets suitable for photorealistic rendering. To address this need, we prioritize high-fidelity reconstruction quality from high-resolution multi-view captures spanning thousands of identities and millions of frames.

We propose \emph{HeadsUp}, a scalable feed-forward approach for reconstructing high-fidelity 3D Gaussian head assets from large-scale multi-camera studio captures. Specifically, our method transforms input multi-view images to the UV parameterization of a set of 3D Gaussians 
attached to a neutral head template. In contrast to pixel-aligned prediction~\cite{kirschstein2025avat3r,ji2026fastgha}, the UV-space formulation decouples the number of output Gaussians from the number and resolution of input images. This allows our method to scale gracefully with the number of high-resolution input views, which is crucial for resolving fine details such as hair strands and jewelry. Our feed-forward model is based on an encoder-decoder architecture: a cross-attention transformer efficiently encodes the input images into a compact 2D latent representation, which is then decoded into the corresponding Gaussian UV maps. To better capture the high-frequency details, we introduce a background model to improve fine boundary structures such as hair, and a high-resolution finetuning stage that leverages multi-scale and region-specific losses to enhance overall render quality and the fidelity of core facial features like the eyes and mouth.

To demonstrate the reconstruction quality and scalability of our method, we train and evaluate our model on a large-scale internal dataset that is an order of magnitude larger than existing multi-view face datasets, comprising over $10\,000$ subjects with diverse appearances and facial expressions. For comparison with prior work, we also report results on the publicly available Ava-256 dataset~\cite{martinez2024codec}. \emph{HeadsUp} achieves state-of-the-art reconstruction quality on novel identities and expressions (examples in~\cref{fig:teaser}). We further characterize the scaling behavior of our model with respect to the number of training subjects and input views, as well as the model's capacity. Finally, we highlight the strength of our learned latent space by showcasing two downstream applications: generating novel 3D identities with latent diffusion models and animating the 3D heads from expression blendshapes.\\[0.1mm]

\noindent In summary, our main contributions are as follows:
\begin{itemize}
    \item We address the challenge of large-scale, high-fidelity 3D head reconstruction from multi-camera studio captures spanning thousands of identities and millions of frames.
    \item We introduce \emph{HeadsUp}, a feed-forward model that predicts UV maps that encode the parameters of 3D Gaussians anchored to a neutral head template. Our model decouples the output representation from the input resolution and view count, enabling efficient scaling on dense, high-resolution data.
    \item We introduce explicit background modeling and a tailored training strategy to improve reconstruction in high-frequency regions (hair, eyes, and mouth).
    \item We achieve state-of-the-art reconstruction quality on novel identities and expressions, and characterize the scaling behavior of our model across identities, views, and model capacity.
\end{itemize}

%% file: 02_related_work.tex
\section{Related Work}
\label{sec:related-work}

\subsubsection{Novel View Synthesis of Human Heads.}

Various 3D representations have been explored for head reconstruction and novel view synthesis. Earlier attempts based on Neural Radiance
Fields~\cite{gafni2021nerface,park2021nerfies,athar2022rignerf,kirschstein2023nersemble,zielonka2023instant} have succeeded at photorealistic rendering but required lengthy per-scene optimization. Implicit surface methods~\cite{zheng2022imavatar} and point-based approaches~\cite{zheng2023pointavatar,zhao2024psavatar} offered alternatives with different trade-offs between quality and efficiency. Hybrid methods like MonoNPHM~\cite{giebenhain2024mononphm} combine neural parametric models with implicit representations for monocular reconstruction. Works on Codec Avatars~\cite{lombardi2018deep,ma2021pixelcodec,saito2024rgca,li2024uravatar} and others~\cite{teotia2025audio} achieve high-fidelity head avatars through learning on multiple dense multi-view captures. Often, these methods involve per-subject optimization or fine-tuning at test time. 3DGS optimization-based approaches have quickly become popular due to their efficiency and more explicit representation, albeit limited in scalability because of the large number of Gaussians required. GaussianAvatars~\cite{qian2024gaussianavatars} ties 3D Gaussians to a parametric model~\cite{li2017flame} for fully controllable heads but requires minutes of optimization per identity, excluding multi-view tracking time.  SplattingAvatar~\cite{shao2024splattingavatar} embeds Gaussians within triangle meshes for hybrid mesh-Gaussian avatars. GAF~\cite{tang2025gaf} distills multi-view diffusion priors using pseudo-ground truths to enhance monocular avatar reconstruction. These works, along with~\cite{xiang2024flashavatar,teotia2024gaussianheads} and others, demonstrate that 3DGS is an excellent representation for detailed head avatars. However, the reliance on slow optimization makes scaling to thousands of subjects with high-resolution studio captures challenging.

\subsubsection{Feed-forward 3D Head Reconstruction.}

Following the advances in Large Reconstruction Models (LRMs)~\cite{hong2023lrm,chen2024mvsplat,charatan2024pixelsplat,zhang2024gslrm,xu2024grm,szymanowicz2024splatterimage,wang2025vggt}, recent works develop feed-forward 3DGS head models to bypass per-subject optimization.

\textit{Multi-view feed-forward methods} address the well-constrained setting of reconstruction from multiple input images. GPAvatar~\cite{chu2024gpavatar} learns efficient Gaussian projections from multi-view inputs by employing deep feature extractors. 
HeadGAP~\cite{zheng2025headgap} learns generalizable Gaussian priors from multi-view data and personalizes to new identities from a few images, though it still requires a per-subject adaptation step.
Avat3r~\cite{kirschstein2025avat3r} regresses animatable 3D head avatars from as few as four images using DUSt3R position maps~\cite{wang2024dust3r} and Sapiens features~\cite{khirodkar2024sapiens}. While template-free, its pixel-aligned Gaussians scale linearly with input views, causing memory constraints and potential temporal inconsistencies. The reliance on DUSt3R and Sapiens makes the pipeline time- and memory-constrained in a practical setting. Similarly, FastGHA~\cite{ji2026fastgha} achieves few-shot real-time animation via pixel-aligned features, yet faces comparable scaling limitations. The concurrent work MATCH~\cite{prinzler2026match} efficiently predicts Gaussian textures within the fixed UV layout of a template mesh, whereas our method aggregates multi-view features into a compact latent representation prior to UV decoding. Finally, Pippo~\cite{kant2025pippo} achieves high quality results by employing a diffusion transformer trained on a large-scale dataset. While practical for single images, inference scales to minutes for multi-view captures, and strict 3D view consistency is not guaranteed. Parallel works on full-body reconstruction explore similar concepts, such as efficiently reconstructing animatable humans from pose-free images~\cite{qiu2025lhm,qiu2025pflhm}, regressing Gaussians in UV space~\cite{kwon2024ghg}, or training DiT-based generators~\cite{yang2025sigman}. Despite achieving impressive full-body results, these methods remain bottlenecked by the limited facial resolution of available datasets.

\textit{Single-view feed-forward methods} tackle the ill-posed problem of reconstructing human heads from a single input image. GAGAvatar~\cite{chu2024gagavatar} is one of the first generalizable triplane-based Gaussian head models. Portrait4D~\cite{deng2024portrait4d,deng2024portrait4dv2} learns novel view synthesis and driving from synthetic 2D data and pseudo multi-view supervision, generated from a triplane-based generator. However, triplane representations often struggle with extreme novel viewpoints and are sensitive to image cropping. LAM~\cite{he2025lam} and PanoLAM~\cite{li2025panolam} perform one-shot head synthesis based on UV-aligned Gaussians, similar to our method but with a focus on rigging. FastAvatar~\cite{liang2025fastavatar} predicts residual Gaussians from a canonical template in $<$10\,ms, specifically targeting the fast rigging setting. PercHead~\cite{oroz2025perchead} uses perceptual supervision from DINOv2~\cite{oquab2023dinov2} and SAM~\cite{kirillov2023segmentanything} for disentangled geometry and appearance control. Our method partially borrows inspiration from some of these methods and uses a UV-aligned Gaussian map that naturally supports an arbitrary number of views, including a single view, while remaining fast and supporting a wide range of novel viewpoints. Rather than optimizing for real-time rigging, in this work we prioritize maximum reconstruction quality. Like Pippo~\cite{kant2025pippo}, several recent methods rely on diffusion priors to reconstruct human heads~\cite{lyu2025facelift,yang2025pshead,zhang2024rodinhd,zhang2025high,lu2025gas,zhang2024humanref,li2025pshuman,xue2024human,chen2024generalizable}, but this reliance comes at computational cost and compromises rendering fidelity.

\textit{3D GAN inversion} represents a parallel track for single-image 3D head synthesis. PanoHead~\cite{an2023panohead} and SphereHead~\cite{li2024spherehead} extend EG3D~\cite{chan2022eg3d} to $360^\circ$ head synthesis with back-of-head supervision. Encoder-based latent inversion methods like TriPlaneNet~\cite{bhattarai2024triplanenet} and GOAE~\cite{yuan2023goae}, alongside optimization-based PTI~\cite{roich2022pti}, facilitate reconstruction from real images, while DiffPortrait3D~\cite{gu2024diffportrait3d} and VOODOO3D\cite{tran2024voodoo3d} focus on one-shot reenactment. InvertAvatar~\cite{zhao2024invertavatar} adapts this inversion strategy to multiple images for incremental improvement. However, triplanes can fundamentally limit the level of achievable resolution and expressivity. Furthermore, triplane-based generators rely on external 2D head pose estimation that can introduce compounding errors; these methods also usually impose strict cropping requirements on the input image and camera assumptions~\cite{chan2022eg3d, skorokhodov20233d}.\vspace{2mm}

\noindent\textbf{Summary.} Ultimately, our work distinguishes itself from relevant prior work by combining several key properties: (1) a fully feed-forward architecture that avoids lengthy optimization to a novel subject and decouples memory from the input view count, enabling massive scalability; (2) universal support for monocular, sparse, or dense multi-view inputs; (3) independence from strict parametric face models, allowing for natural reconstruction of accessories and diverse expressions; and (4) state-of-the-art photorealism and subject recognizability in novel views.

%% file: 03_method.tex
\section{Method}
\label{sec:method}
 
\begin{figure}[t]
  \centering
  \includegraphics[width=1.0\linewidth,trim={0 14.5cm 24cm 0},clip]{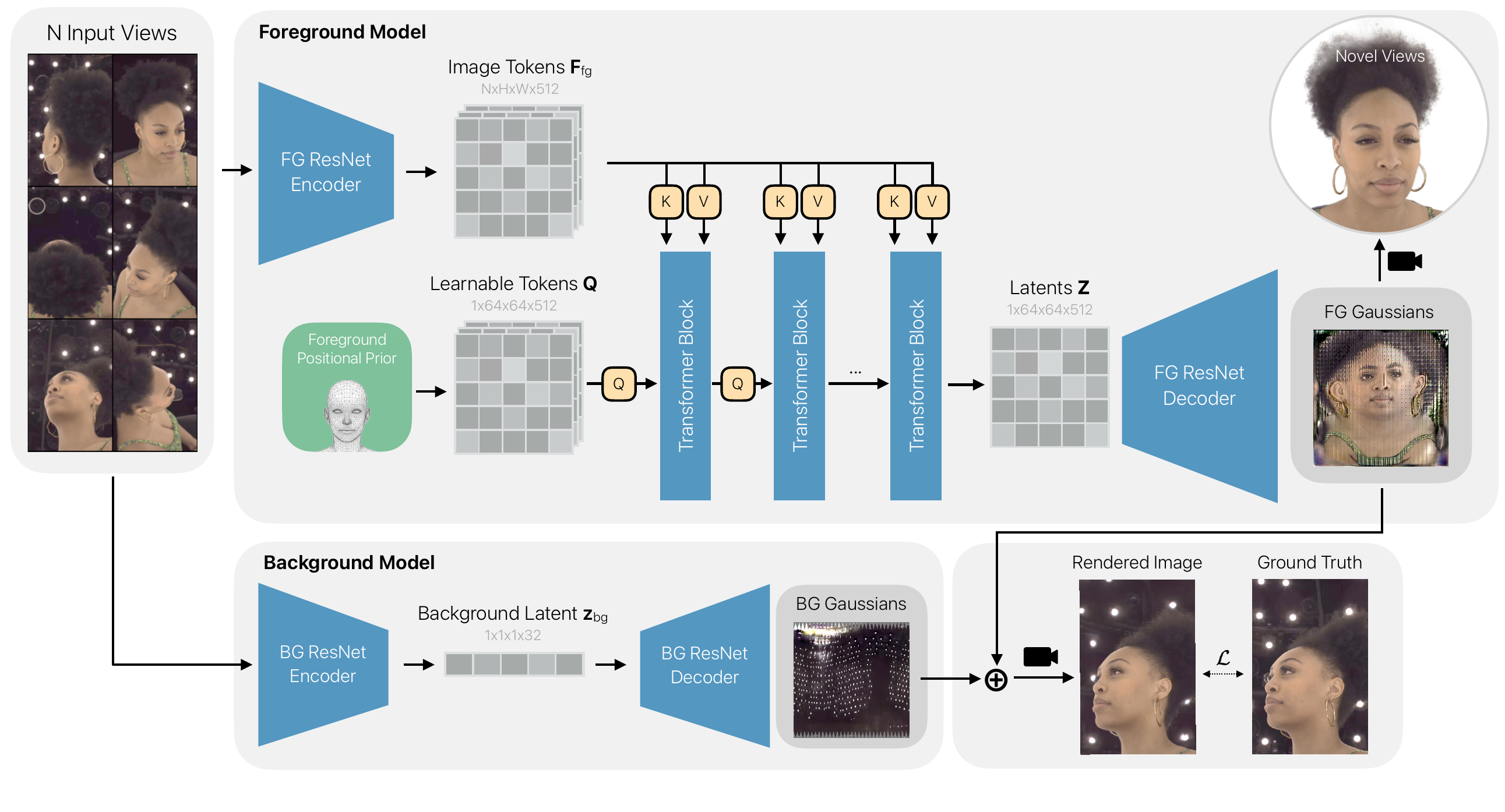}
  \caption{\textbf{Overview of \emph{HeadsUp}.} Our method reconstructs high-fidelity 3D Gaussian heads from multi-view images. Given a set of input views, our model utilizes a transformer-based encoder and a 3D Gaussian decoder to predict UV-parameterized 3D Gaussians for both the foreground and background. The model is trained end-to-end using a combination of photometric and perceptual supervision. \label{fig:headsup-overview}}
\end{figure}

Here we introduce our method \emph{HeadsUp}. \cref{fig:headsup-overview} shows the overall architecture of our model. Given $N$ calibrated time-synchronized images, our model jointly predicts 3D Gaussians for the foreground and background. Explicit background modeling bypasses the need for foreground matting, which often fails to accurately segment high-frequency boundary details like hair or jewelry.

Our model is based on  UV-parameterized 3D Gaussians anchored to a template mesh. Our feed-forward network consists of two main components: a \emph{multi-view encoder} based on a transformer architecture to extract latents from the input images, and a \emph{3D Gaussian decoder} to map the latents to the 3D Gaussians' UV parameters. To ensure high-fidelity rendering of core facial features without prohibitive computational costs, we employ a two-stage training strategy where a high-resolution finetuning stage leverages region-specific and multi-scale supervision. Below, we detail our model architecture and training methodology.

\subsection{UV-Parameterized 3D Gaussian Splatting}
\label{subsec:methods-uv-gaussians}

Our model predicts a multi-channel 2D UV map, representing a fixed number of Gaussians anchored to a template mesh. The channels of the UV map correspond to the Gaussian attributes: position $\boldsymbol{\mu} \in \mathbb{R}^3$, scale $\mathbf{s} \in \mathbb{R}^3_+$, quaternion rotation $\mathbf{q} \in \mathbb{R}^4$, opacity $\alpha \in [0, 1]$, and spherical harmonic color coefficients $\{\mathbf{c}^{(\ell,m)}\}$ up to degree $L=1$. The foreground Gaussians are anchored to a neutral head template shared among all identities in a canonical coordinate system. 
The head canonical coordinate system is defined with its origin at the mid-pupil point of the template and its orientation aligned to the Frankfurt plane~\cite{ISO7250-1:2017}. The background Gaussians are anchored to a sphere template fitted to the capture rig and are transformed into the canonical coordinate system for joint rendering. 

The UV formulation in our method offers several advantages, including: 

\noindent\emph{(1) Shared geometric prior}. The canonical mesh topology provides consistent spatial structure across subjects and expressions, allowing the network to focus on appearance and local variations rather than coarse global head geometry.

\noindent\emph{(2) Efficient multi-view aggregation.} The mesh-anchored UV parameterization decouples the number of output Gaussians from the number and resolution of input images, allowing for efficient aggregation of information from dense and high-resolution captures.

\noindent\emph{(3) Robustness to tracking errors.} As the 3D Gaussians are anchored to a fixed neutral template, our representation only requires rigid head pose tracking and does not rely on fragile and error-prone facial expression tracking for inference.

\subsection{Network Architecture}
\label{subsec:network-architecture}

Here we describe our network architecture, composed of a \emph{multi-view encoder} and a \emph{Gaussian UV decoder}.\\[2mm]

\noindent \textbf{Multi-View Encoder.}
Given $N$ calibrated input views $\{\mathbf{I}_i\}_{i=1}^N$, $\mathbf{I}_i \in \mathbb{R}^{3 \times H \times W}$, and their camera extrinsics $\{\mathbf{E}_i\}_{i=1}^N$, $\mathbf{E}_i \in \mathbb{R}^{4 \times 4}$, along with intrinsics $\{\mathbf{K}_i\}_{i=1}^N$, $\mathbf{K}_i \in \mathbb{R}^{3 \times 3}$, we first patchify the images and convert each into patch embeddings $\mathbf{e}^{i}_p \in \mathbb{R}^{d \times h \times w}$. To explicitly encode the camera geometry into the features, we concatenate per-patch Pl\"ucker embeddings~\cite{sitzmann2021lfns} with the corresponding patch embeddings along the channel dimension:
\begin{equation}
\mathbf{f}^{i}_p = \text{Concat}(\mathbf{e}^{i}_p, \mathcal{P}(\mathbf{K}_i, \mathbf{E}_i)) \quad \in \mathbb{R}^{(d+6) \times h \times w}
\end{equation}
where $\mathcal{P}(\cdot)$ computes the 6D Pl\"ucker coordinates. Two parallel convolutional encoders process these features to disentangle each input view into foreground and background features, $\mathbf{F}^{\text{fg}}_i \in \mathbb{R}^{c \times h_f \times w_f}$ and $\mathbf{F}^{\text{bg}}_i \in \mathbb{R}^{c' \times h_f \times w_f}$, respectively.

We employ a transformer architecture \cite{vaswani20217transformer,dosovitskiy2020vit} to map these unstructured multi-view foreground features to a low-resolution 2D latent representation of the target 3D head. The transformer converts a 2D grid of learnable tokens $\mathbf{Q} \in \mathbb{R}^{d_z \times h_z \times w_z}$ to the 2D latent $\mathbf{Z} \in \mathbb{R}^{d_z \times h_z \times w_z}$ by aggregating information from multi-view features through cross-attention layers:
\begin{equation}
\mathbf{Z} = \text{CrossAttnTransformer}(\mathbf{Q}, \mathbf{F}_{\text{fg}}).
\end{equation}
When discussing our latent representation throughout the paper, we refer to the foreground latent $\mathbf{Z}$.

To model the background, we use a shallow convolutional network with global average pooling to map the stacked multi-view background features $\{\mathbf{F}^{\text{bg}}_i\}_{i=1}^N$ to the background latent $\mathbf{z}_{\text{bg}} \in \mathbb{R}^{d_{bg}}$. This design choice is motivated by the fact that the background is mostly constant between different frames, apart from small variations such as lighting.

\subsubsection{3D Gaussian UV Decoder.}
We use a convolutional decoder to convert the resulting latent variable $\mathbf{Z}$ into a high-resolution UV feature map $\mathbf{U} \in \mathbb{R}^{H\times W \times 23}$. Specifically, for each UV location $(u,v)$, the output UV features are mapped to the corresponding 3D Gaussian attributes as follows:
\begin{align}
\boldsymbol{\mu}_{u,v} &= \mathbf{V}{(u,v)} + \mathbf{U}^{(\mu)}(u,v), \quad \|\mathbf{U}^{(\mu)}(u,v)\| \leq \delta_{\max} \\
\mathbf{s}_{u,v} &= \exp(\mathbf{U}^{(s)}(u,v))\\
\mathbf{q}_{u,v} &= \text{normalize}(\mathbf{U}^{(q)}(u,v))\\
\alpha_{u,v} &= \sigma(\mathbf{U}^{(\alpha)}(u,v)) \\
\{\mathbf{c}_{u,v}^{(\ell,m)}\} &= \mathbf{U}^{(c)}(u,v),
\end{align}
where $\mathbf{V}(u,v)$ is the corresponding 3D vertex position on the template mesh $\mathbf{V} \in \mathbb{R}^{(H \times W) \times 3}$, and $\delta_{\max}$ is the position offset bound (empirically set to $200\mathrm{mm}$). 
Similarly, $\mathbf{z}_{bg}$ is decoded into the background 3D Gaussians using a similar but separate decoder (with $\delta_{\max}=10\mathrm{mm}$). During training, we find it important to perform a warm-up period of 1000 iterations, where opacity and scale attributes are detached from the gradient backpropagation graph.

\subsection{Training Objectives}
\label{sec:training_objectives}
Our overall training loss consists of reconstruction and regularization terms:
\begin{equation}
    \mathcal{L}_{\mathrm{total}} = \lambda_{\mathrm{L1}} \mathcal{L}_{\mathrm{L1}} + \lambda_{\mathrm{LPIPS}} \mathcal{L}_{\mathrm{LPIPS}} + \lambda_{\mathrm{adv}} \mathcal{L}_{\mathrm{adv}} + \lambda_{\mathrm{pos}} \mathcal{L}_{\mathrm{pos}} + \lambda_{\mathrm{mask}} \mathcal{L}_{\mathrm{mask}} + \lambda_{\mathrm{TV}} \mathcal{L}_{\mathrm{TV}},
\end{equation}
where the $\lambda$ hyperparameters control the relative influence of each term.

\subsubsection{Reconstruction Losses.}
Let $I$ and $I_{\mathrm{gt}}$ denote the rendered and ground-truth composite images (foreground and background) from a randomly sampled view. We optimize an L1 photometric loss $\mathcal{L}_{\mathrm{L1}} = \| I - I_{\mathrm{gt}} \|_1$ and a multi-scale LPIPS perceptual loss $\mathcal{L}_{\mathrm{LPIPS}}$~\cite{johnson2016perceptual}. To enhance high-frequency details, we employ a Perceptual Discriminator~\cite{goodfellow2014generative, Sungatullina_2018_ECCV} to compute an adversarial loss $\mathcal{L}_{\mathrm{adv}}$. To maintain training stability, $\mathcal{L}_{\mathrm{adv}}$ is activated only after $240\mathrm{k}$ iterations, and the discriminator operates on random $256 \times 256$ spatial crops of the input.

\subsubsection{Regularization Terms.}
To guide the geometry during the warm-up phase, we utilize expression-tracked meshes to regularize the Gaussian positions. Specifically, the loss term $\mathcal{L}_{\mathrm{pos}}$ penalizes the distance between $\boldsymbol{\mu}(u,v)$ and the 3D point $\mathbf{V}^{e}(u,v)$, derived via barycentric interpolation of the Gaussian UV coordinates on a tracked mesh $\mathbf{V}^e$. Concurrently, a silhouette loss $\mathcal{L}_{\mathrm{mask}}$ minimizes the discrepancy between the rendered foreground alpha map and the ground-truth segmentation mask $M_{\mathrm{gt}}$. The weights for both $\mathcal{L}_{\mathrm{pos}}$ and $\mathcal{L}_{\mathrm{mask}}$ are decayed over the course of training. Finally, a Total Variation loss $\mathcal{L}_{\mathrm{TV}}$ is applied to the rendered UV-space colors~\cite{kirschstein2024gghead} to encourage spatial smoothness and prevent surface holes.

\subsection{Two-Stage Training}
\label{sec:highres_finetuning}

Training our model on a large dataset of high-resolution input images can be computationally expensive as the transformer scales quadratically with the number of input tokens. Therefore, we adopt a two-stage training strategy: we first train on $2\times$ downscaled images, followed by a high-resolution finetuning stage that uses the native image resolution. During high-resolution finetuning we make two key modifications:

\begin{itemize}
    \item \textbf{Region-Specific Losses.} We introduce region-specific losses for areas with high-frequency details. Specifically, we extract image crops around the eyes and mouth using the head canonical coordinate frame and supervise with additional multi-scale LPIPS perceptual losses.

    \item \textbf{Multi-Resolution Loss Strategy.} We observed that applying global perceptual and discriminator losses directly at native resolution leads to training instability, likely due to noisy high-frequency gradients. Therefore, to ensure stable convergence, we compute these losses on $2\times$ downsampled outputs.
    
\end{itemize}

\noindent More implementation details are provided in the Supplementary Material.

%% file: 04_experiments.tex
\section{Experiments}
\label{sec:experiments}

\subsection{Datasets}
\label{subsec:experiments-datasets}

\subsubsection{Internal Multi-View Head Dataset.}
Our internal dataset contains over \num{10000} unique participants recorded in a head-focused rig with \num{16} calibrated RGB cameras. The images have a resolution of $1000 \times 750$. Participants were asked to perform various facial expressions and speech sequences. We train on \num{10000} subjects with \num{100} frames per subject sampled for expression diversity. We evaluate our method on \num{50} validation subjects, using \num{20} frames per subject. All participants provided written informed consent for use of their data. The internal dataset will not be publicly released. Images shown in this paper are from subjects who provided explicit written consent for the use of their images in the publication and visualizations.

\subsubsection{Ava-256 Dataset.} We also finetune and evaluate our model on the public dataset Ava-256 ~\cite{martinez2024codec}, which contains high-resolution multi-view head captures. We use the \num{4}TB version of the dataset, containing \num{256} subjects recorded with \num{80} RGB cameras. The images have a resolution of $1024 \times 667$. Following Avat3r's setup~\cite{kirschstein2025avat3r}, we train on 244 subjects with 1000 frames per subject and evaluate on \num{12} validation subjects with around 2000 sampled frames in total. Unlike previous methods restricted to frontal views, we sample views from all \num{80} cameras.

\subsection{Experimental Setup}
\label{subsec:experiment-setup}

\subsubsection{Training.}
We train our model on the internal dataset for \num{900}K steps with $2\times$ downscaled images (batch size \num{64}) in the first stage,  followed by \num{200}K steps of full-resolution finetuning (batch size \num{32}). We use \num{10} input images for this setup. Training with the Adam optimizer~\cite{kingma2014adam} with a learning rate of $2 \times 10^{-4}$ converges in approximately \num{10} days on \num{16} H\num{100} GPUs with \texttt{bfloat16} precision. We perform full-resolution finetuning on Ava-256 for \num{200}K steps with \num{16} input images, converging in less than a day.

\subsubsection{Metrics.}
We evaluate rendering quality with three image metrics: Peak Signal-to-Noise Ratio (PSNR), Structural Similarity (SSIM), and Learned Perceptual Image Patch Similarity (LPIPS)~\cite{zhang2018unreasonable}. Following Avat3r~\cite{kirschstein2025avat3r}, we also report two face-specific metrics: Average Keypoint Distance (AKD) which measures the distance in pixels between 2D keypoints estimated from PIPNet~\cite{jin2021pixel}, and cosine similarity (CSIM) of ArcFace identity embeddings~\cite{deng2019arcface}.

\subsection{Baseline}
For our baseline comparison, we evaluate our method against Avat3r, representing the state-of-the-art in feed-forward 3D head reconstruction.

\subsubsection{Avat3r~\cite{kirschstein2025avat3r}.}
Avat3r uses DUSt3R~\cite{wang2024dust3r} position maps and Sapiens features~\cite{khirodkar2024sapiens} to reconstruct head avatars from sparse views. Since the official code is not released, we carefully reimplemented Avat3r and made some modifications to enable fair comparison to our work. Specifically, we adapt Avat3r from a sparse-view self-reenactment setting to a large-scale 3D head reconstruction pipeline by (1) removing the expression rigging module and (2) providing time-synchronized multi-view images as input. Furthermore, the original method precomputes DUSt3R and Sapiens features for only \num{10} frames per subject, whereas our experimental setup scales this to \num{1000} frames per subject. As extracting DUSt3R position maps at this scale is computationally prohibitive, we instead compute them using the more efficient VGGT~\cite{wang2025vggt}.

\input{figures/4_ava256_renders}
\input{tables/4_ava256_results}

\subsection{Baseline Comparison}
\label{subsec:experiments-results}
We compare our method with Avat3r~\cite{kirschstein2025avat3r} on our internal dataset (Internal10K) and Ava-256 dataset~\cite{martinez2024codec} using different metrics. In Ava-256 experiments, both our model and Avat3r are finetuned from the models pretrained on Internal10K. As shown in~\cref{tab:experiments}, our method significantly outperforms the baseline, with major gains in rendering quality (PSNR, LPIPS) and face fidelity (AKD, CSIM) metrics, while requiring more than an order of magnitude fewer Gaussians. Qualitative results in~\cref{fig:4_ava256_renders} also show renders with much higher fidelity and sharper details for our method, capturing fine hair and facial details, with sharper eye and mouth regions than the baseline. Notably, Avat3r is bottle-necked by GPU memory limits at 6 views, whereas our efficient formulation allows for many more inputs. More results and video visualizations are provided in the Supplementary Material.

\input{04_02_ablations}
\input{figures/4_downstream_applications}

\subsection{Downstream Applications}

\subsubsection{Text-driven Identity Generation.}

Our compact, yet information-rich latent space enables a range of downstream applications, such as novel identity generation. To show this, we train a text-conditioned DiT~\cite{peebles2023scalable} on a large dataset of latents $\mathbf{Z}$ precomputed from our base model. At inference time, we sample latents and decode them into Gaussians using our frozen decoder. \cref{subfig:novel-identities} shows randomly sampled identities from our trained model. Based on face-similarity analysis, we have confirmed these identities  do not appear in the training set. 

\subsubsection{Blendshape-driven Latent Animation.} We showcase facial animation controlled by expression blendshapes, operating entirely within our latent space. Given triplets $(Z_n,Z_b,b)$ of a neutral latent $Z_n$, an expression latent $Z_b$ of the same subject and the corresponding blendshape coefficients $b$, we train a transformer $F_\theta$ to predict the target expression latent $\hat{Z}_b = F_\theta(Z_n, b)$ with supervised losses on the latents, Gaussians and renders. Results are displayed in \cref{subfig:blendshape-rigging}.

%% file: figures/4_ava256_renders.tex
\begin{figure*}[t]
\centering
\setlength{\tabcolsep}{0pt}
\renewcommand{\arraystretch}{0}
\footnotesize
\resizebox{\linewidth}{!}{%
\begin{tabular}{@{}ccc@{\hspace{1mm}}ccc@{\hspace{1mm}}cc@{\hspace{1mm}}c@{}}
\multicolumn{3}{c}{$N=4$} & \multicolumn{3}{c}{$N=6$} & \multicolumn{2}{c}{$N=16$} & \\[6pt]
Input & Avat3r & Ours & Input & Avat3r & Ours & Input & Ours & GT \\[4pt]
\includegraphics[height=0.14\linewidth]{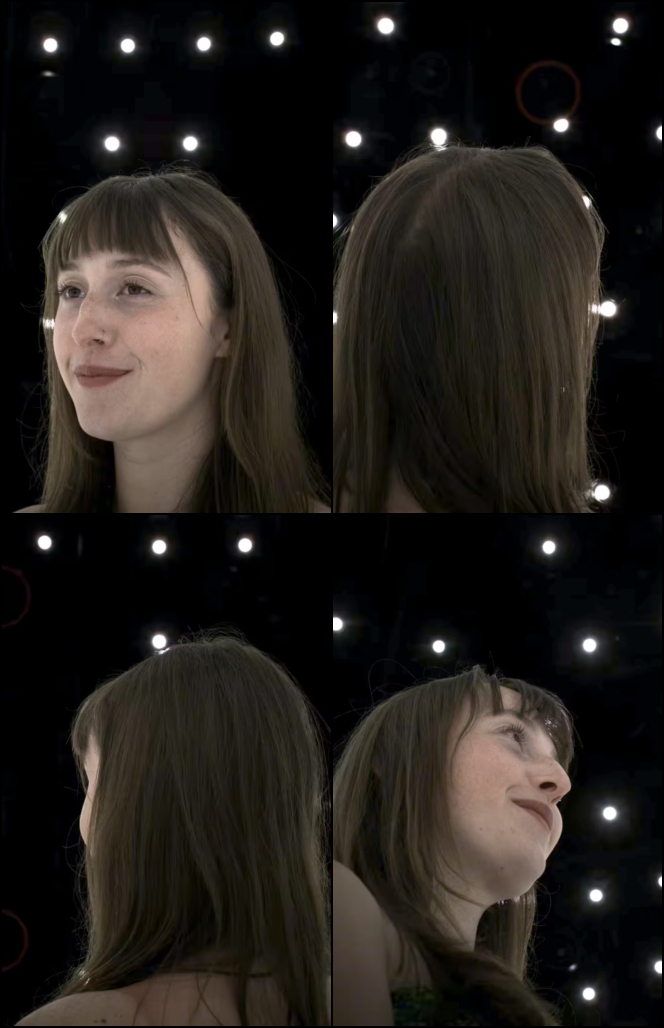} &
\includegraphics[height=0.14\linewidth]{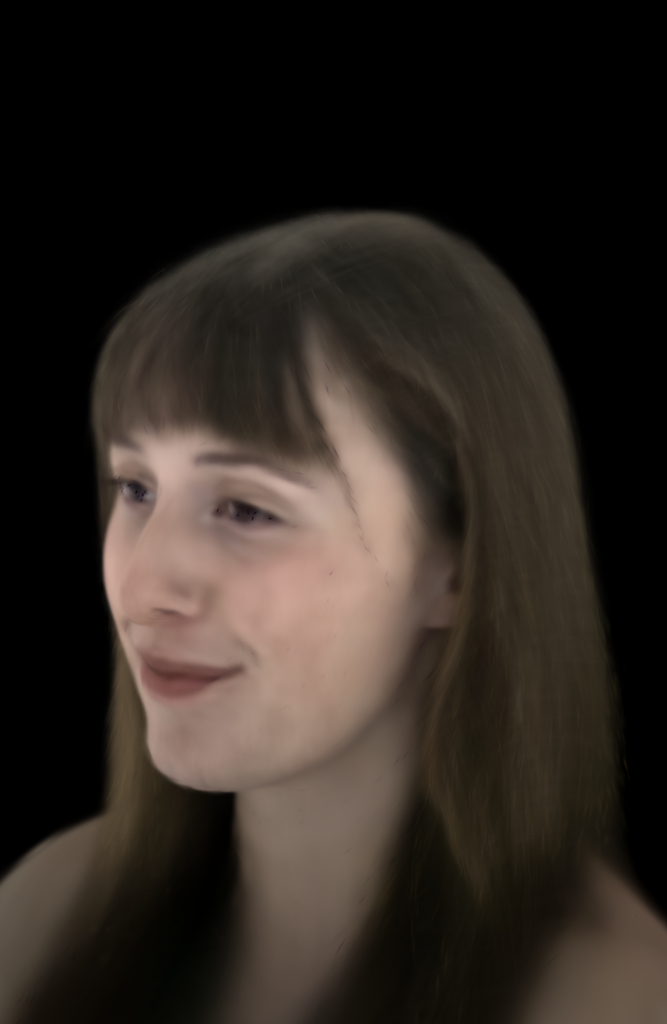} &
\includegraphics[height=0.14\linewidth]{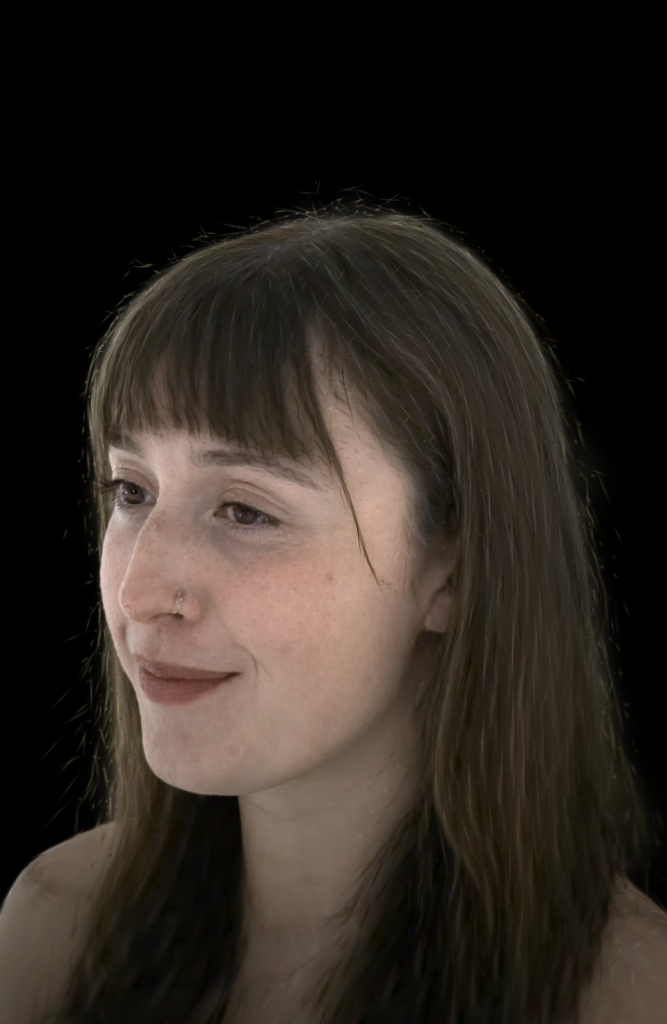} &
\includegraphics[height=0.14\linewidth]{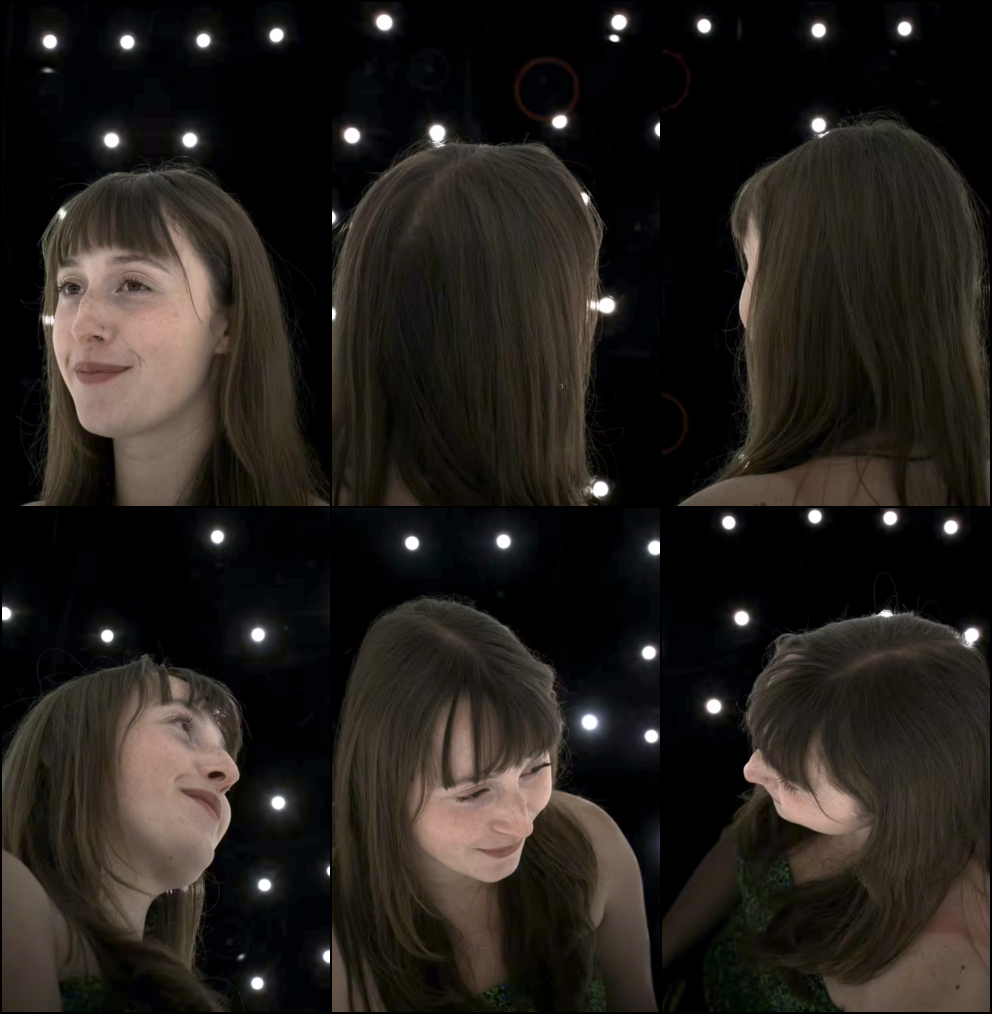} &
\includegraphics[height=0.14\linewidth]{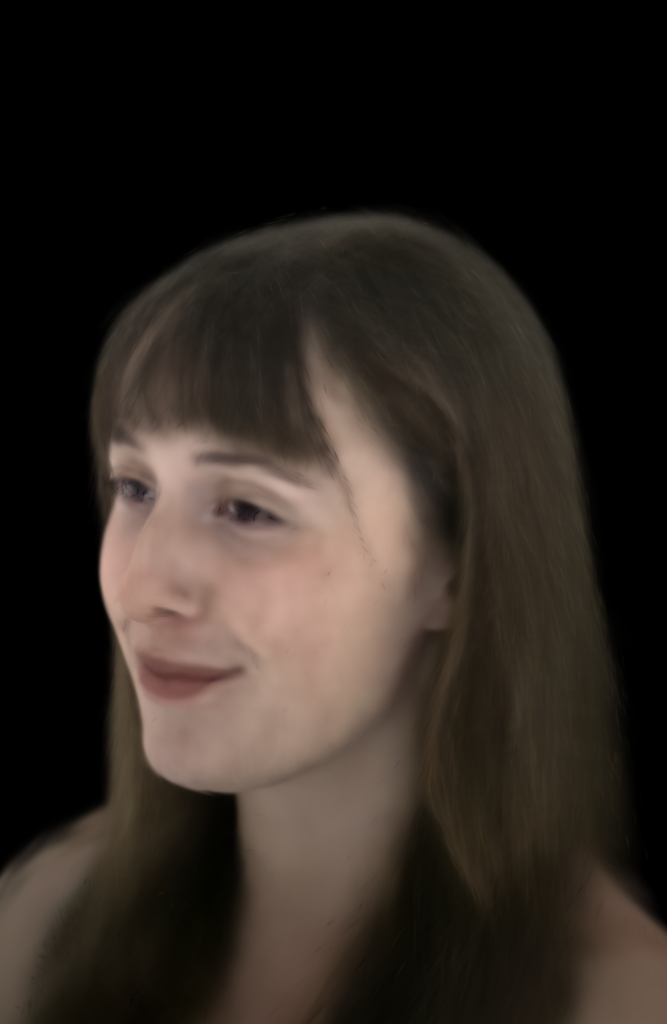} &
\includegraphics[height=0.14\linewidth]{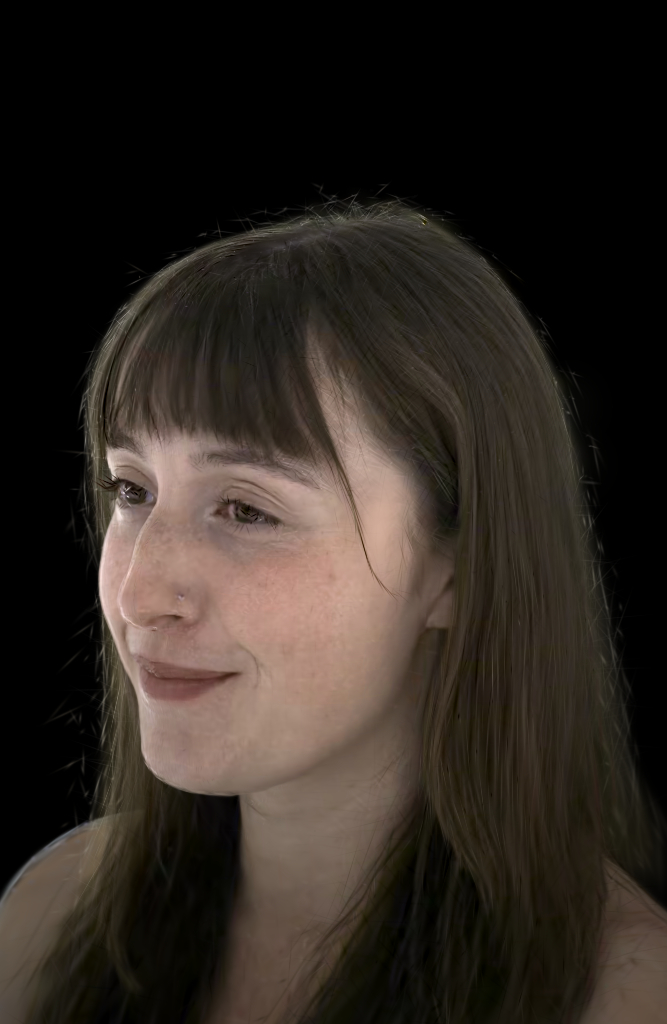} &
\includegraphics[height=0.14\linewidth]{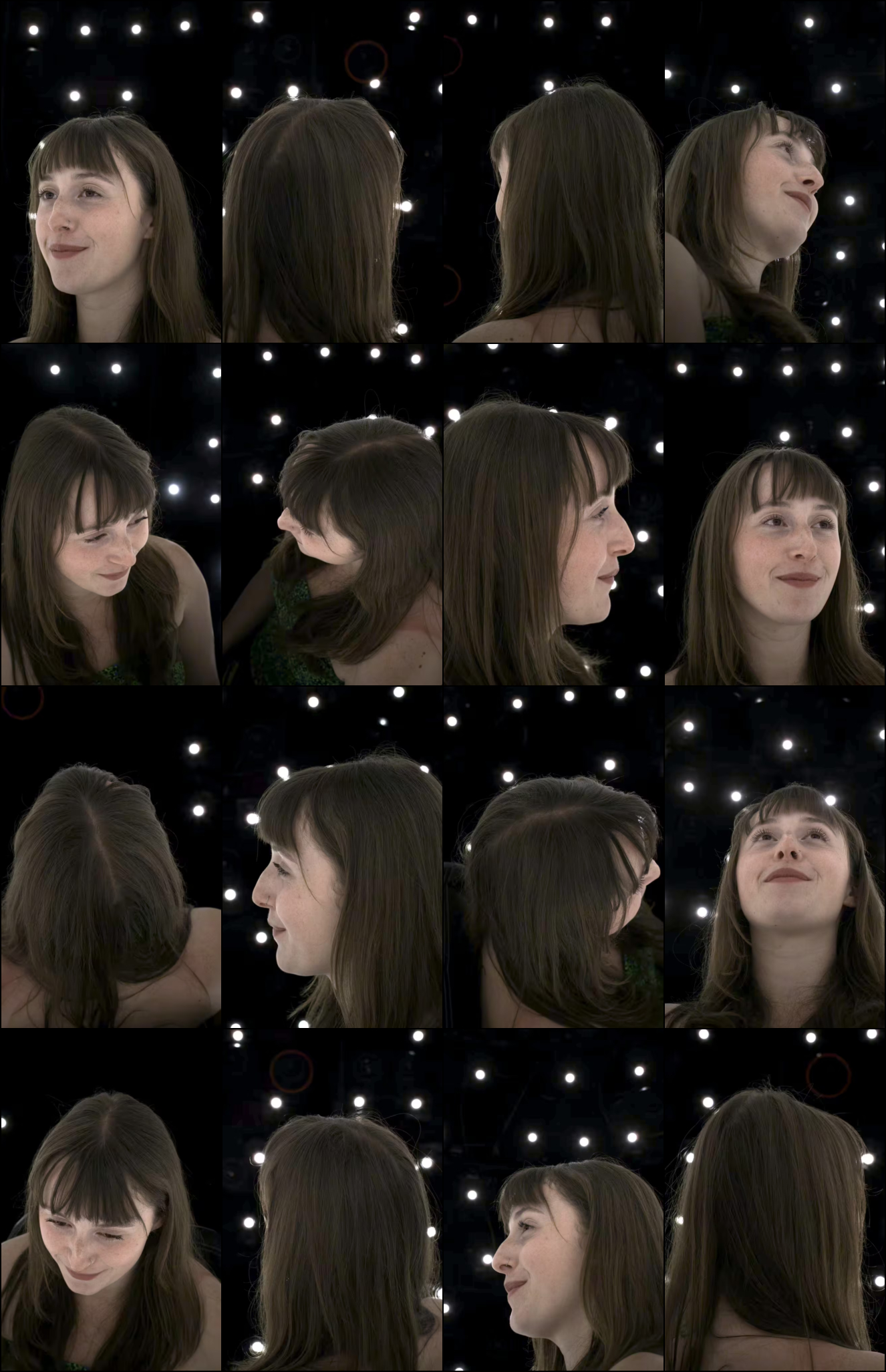} &
\includegraphics[height=0.14\linewidth]{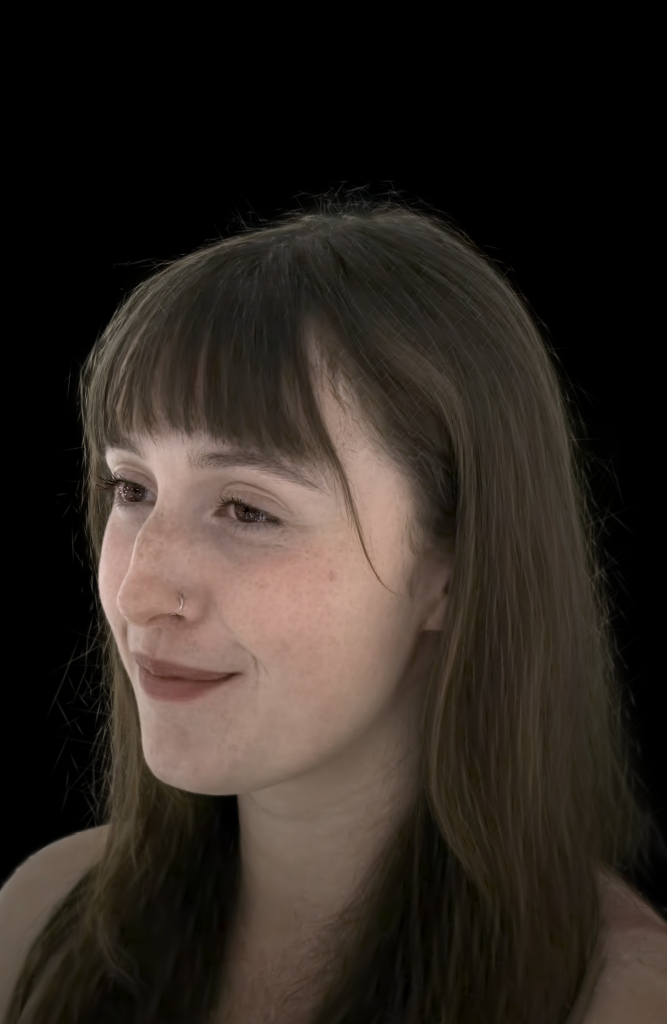} &
\includegraphics[height=0.14\linewidth]{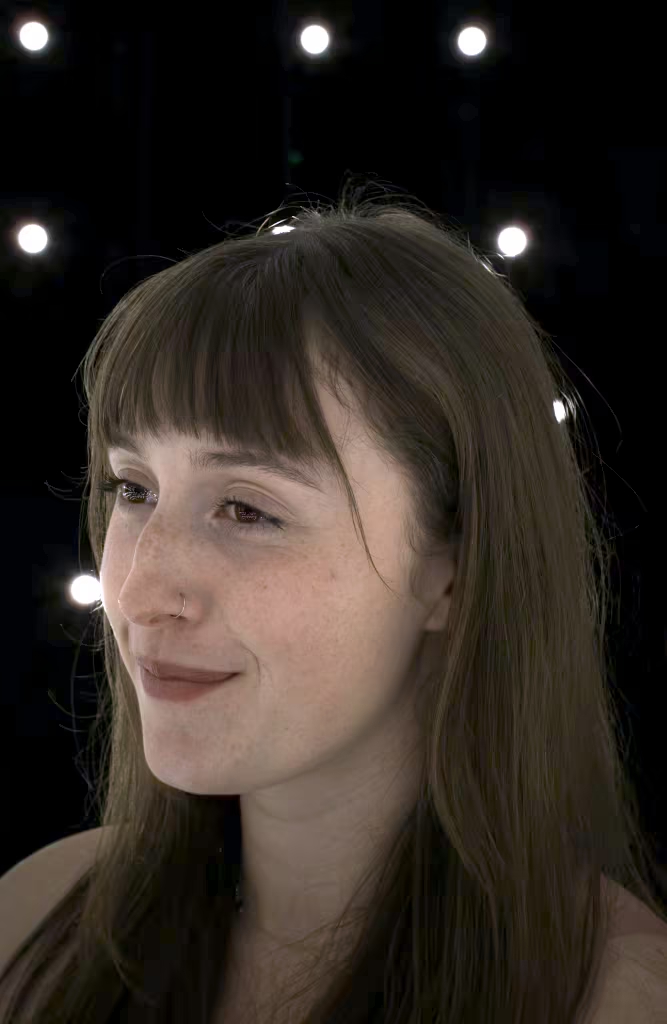} \\
\includegraphics[height=0.14\linewidth]{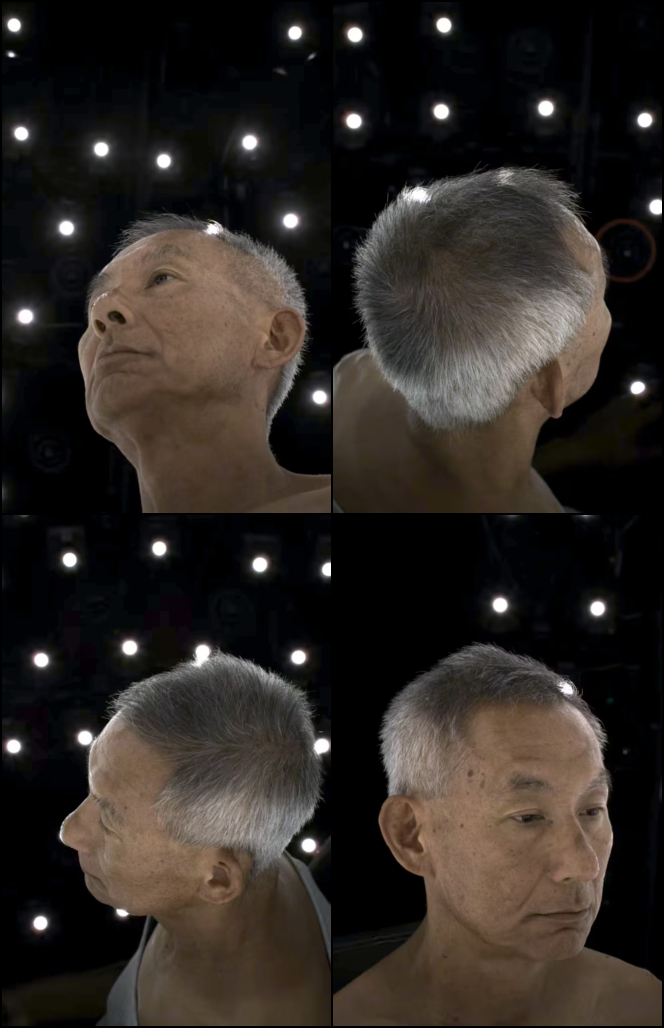} &
\includegraphics[height=0.14\linewidth]{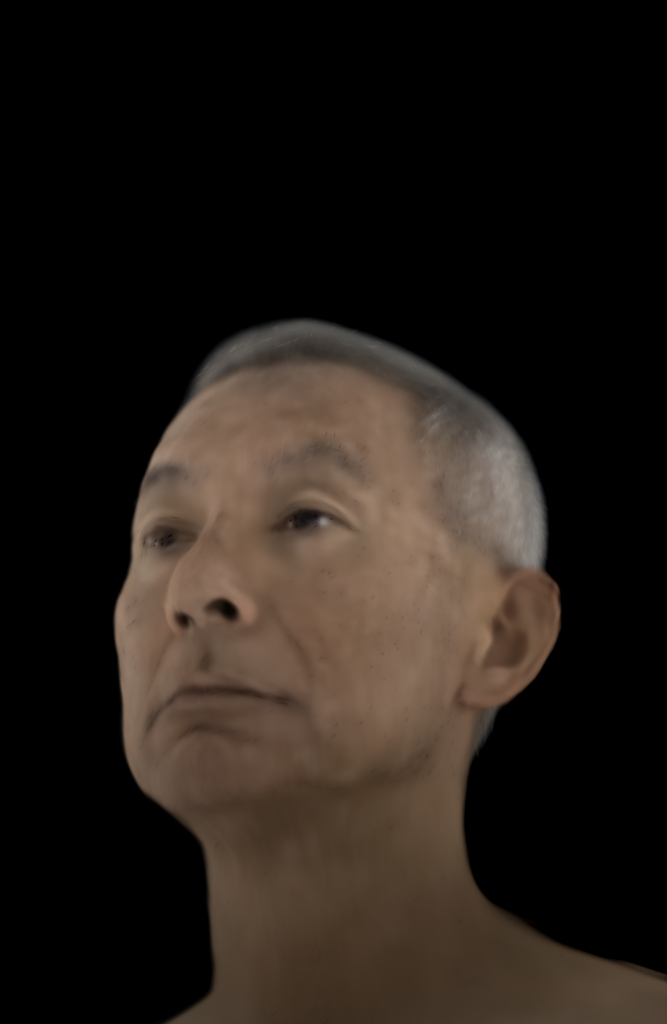} &
\includegraphics[height=0.14\linewidth]{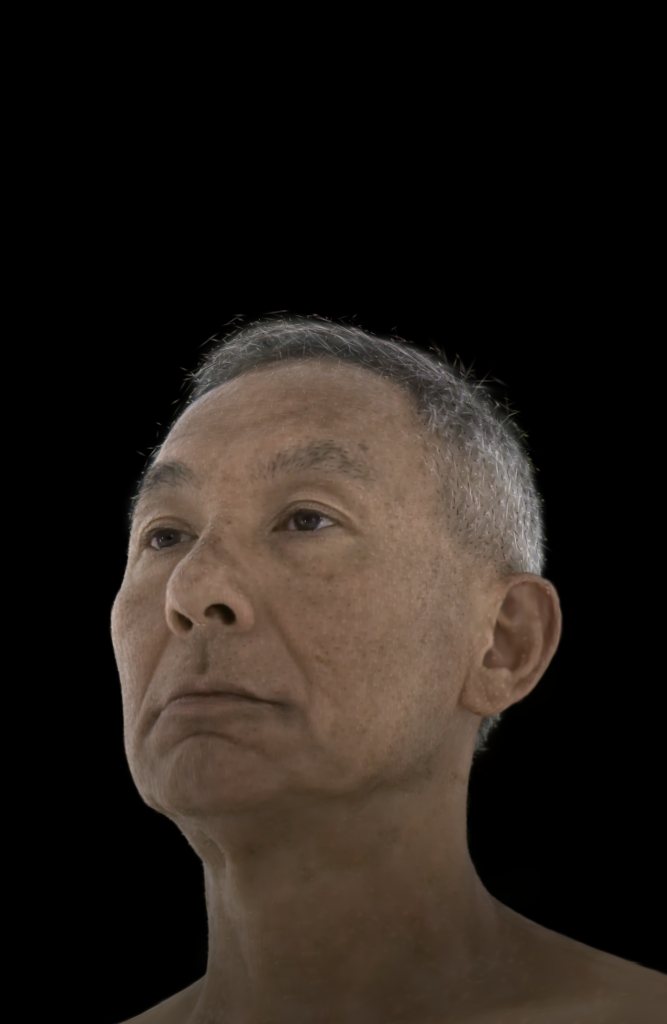} &
\includegraphics[height=0.14\linewidth]{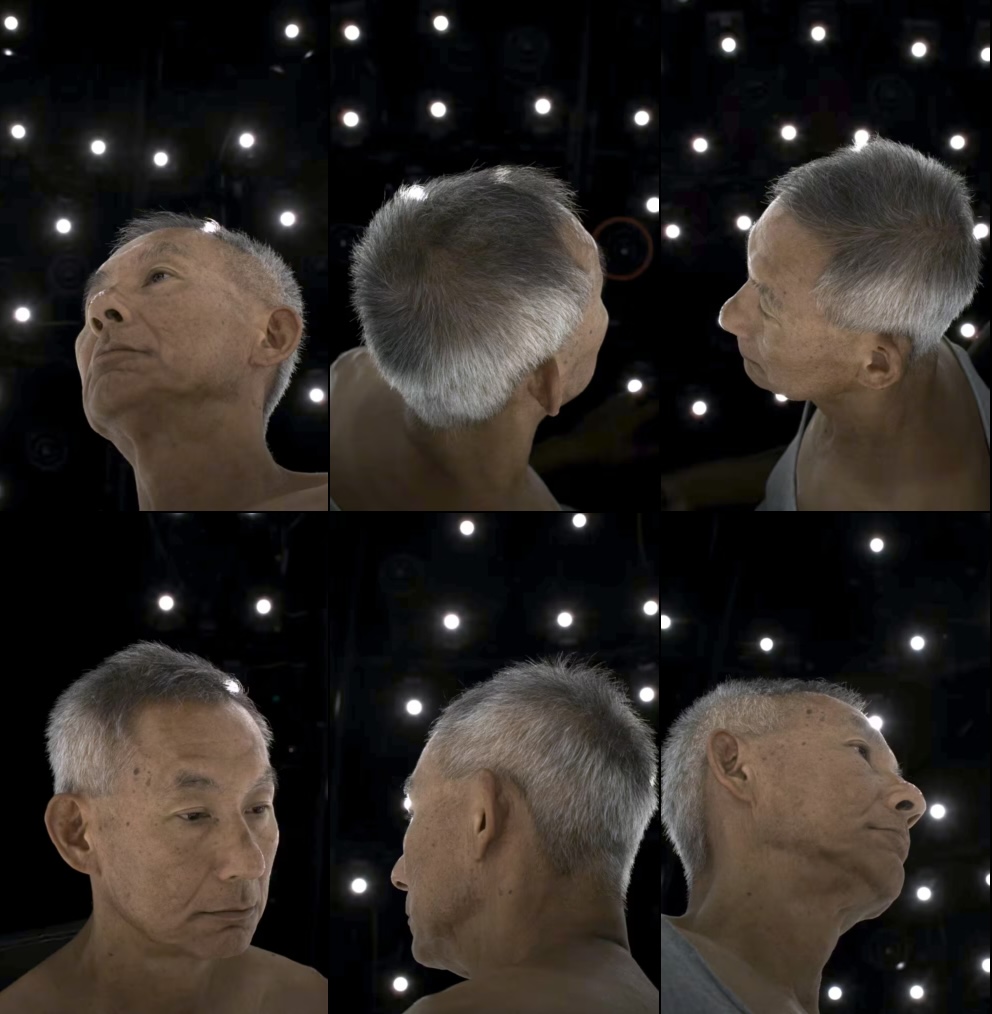} &
\includegraphics[height=0.14\linewidth]{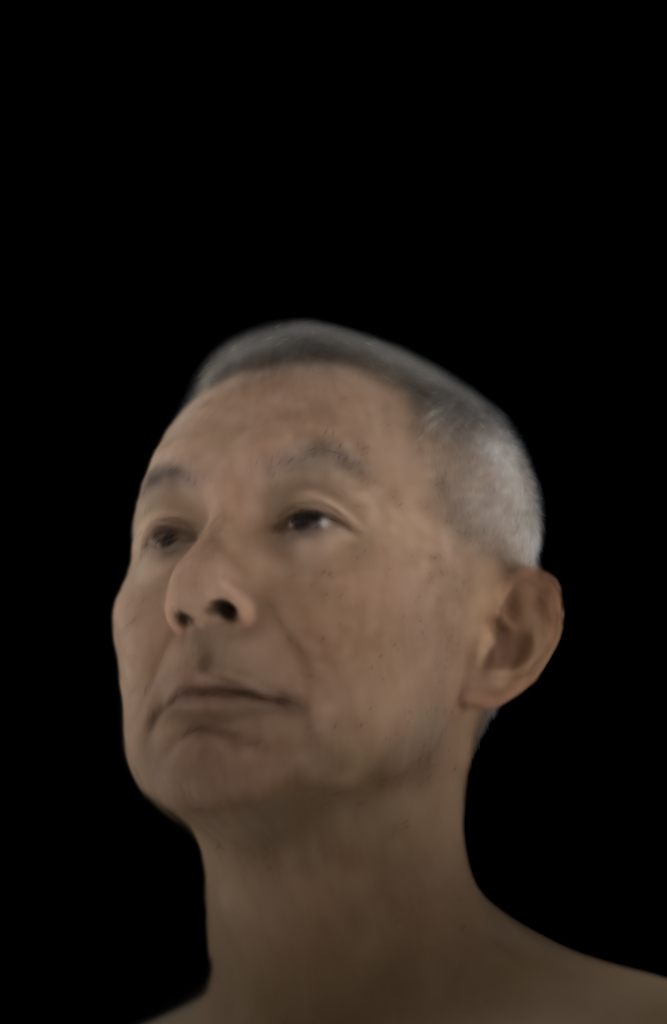} &
\includegraphics[height=0.14\linewidth]{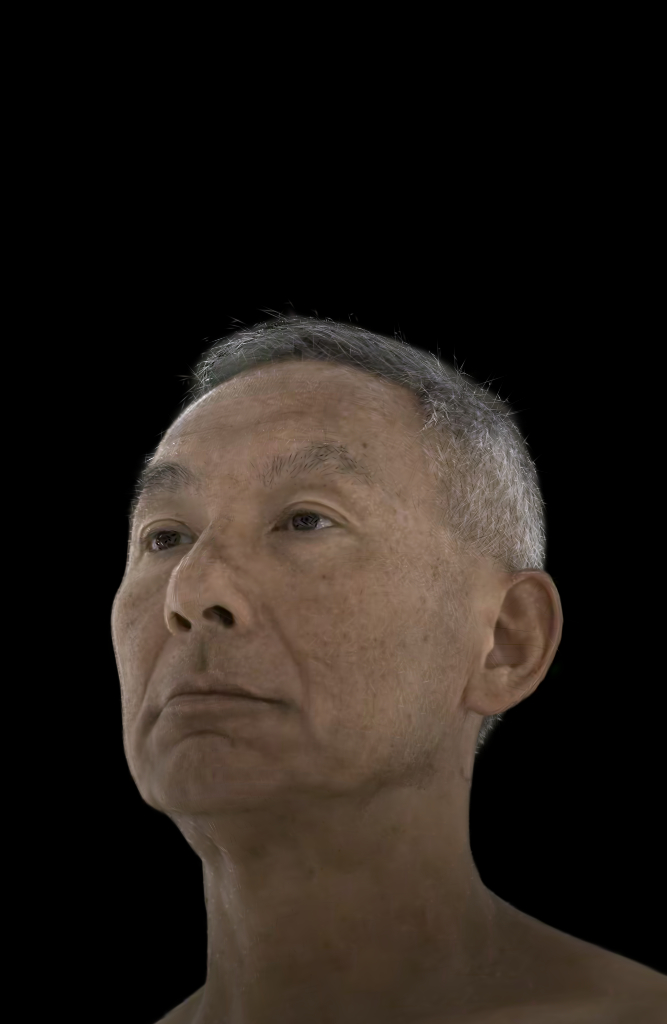} &
\includegraphics[height=0.14\linewidth]{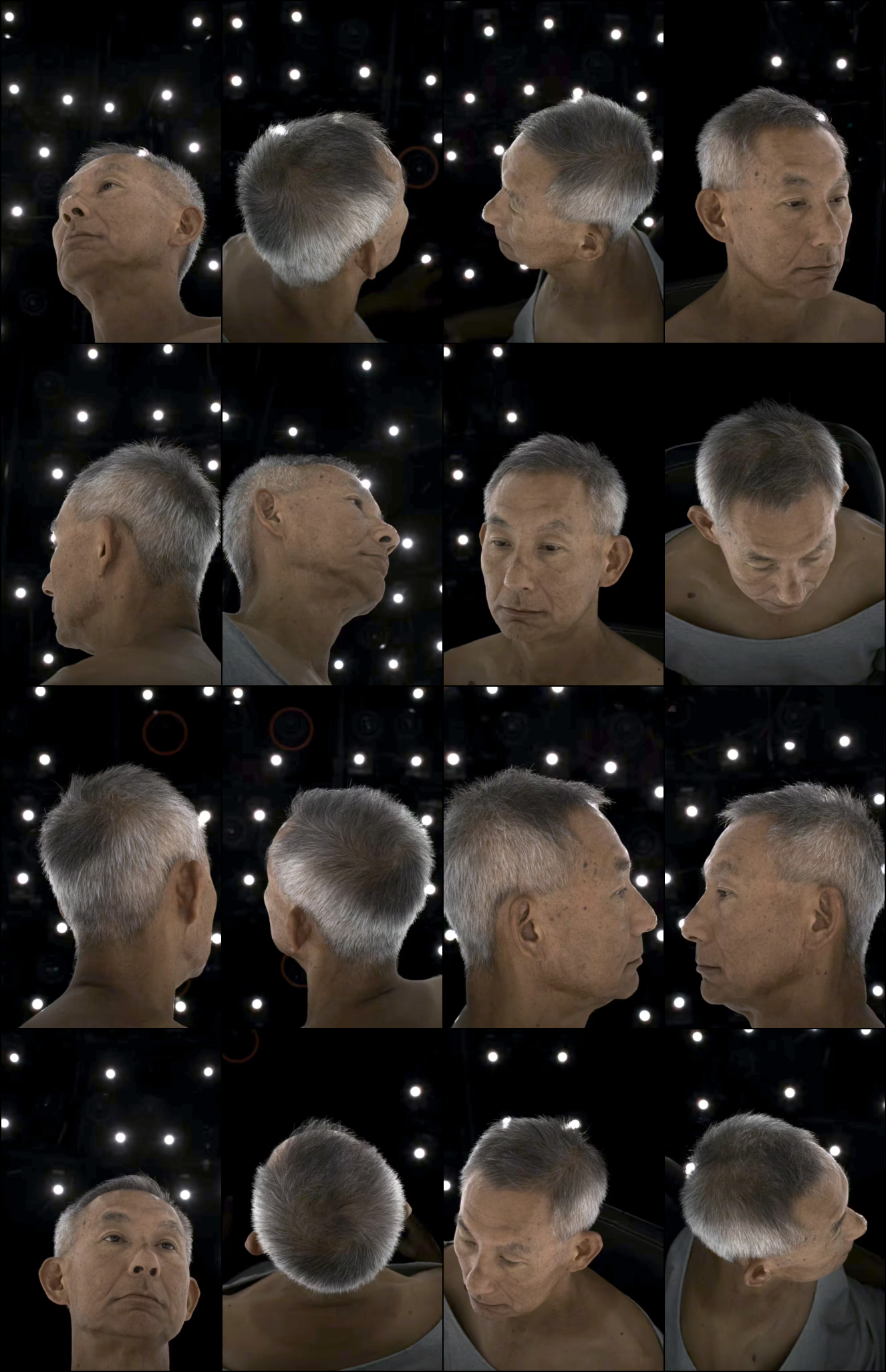} &
\includegraphics[height=0.14\linewidth]{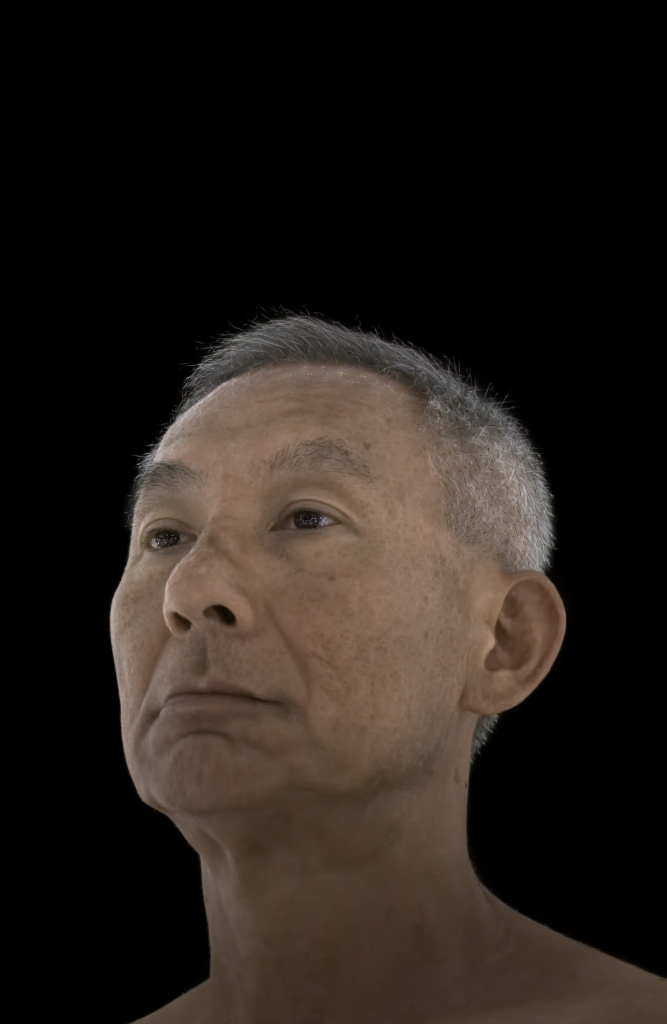} &
\includegraphics[height=0.14\linewidth]{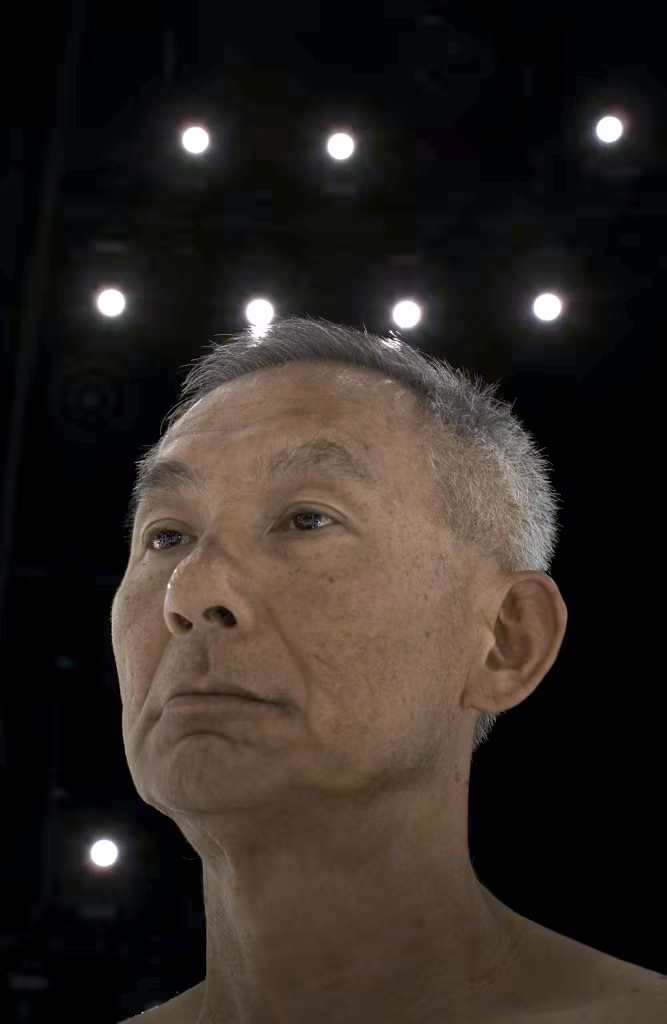} \\
\includegraphics[height=0.14\linewidth]{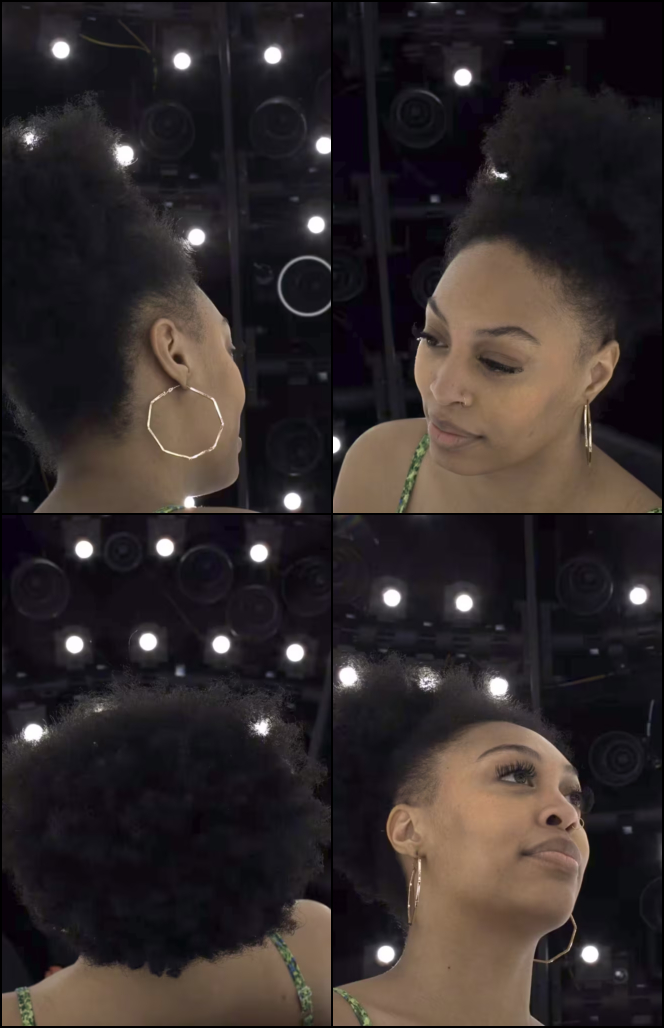} &
\includegraphics[height=0.14\linewidth]{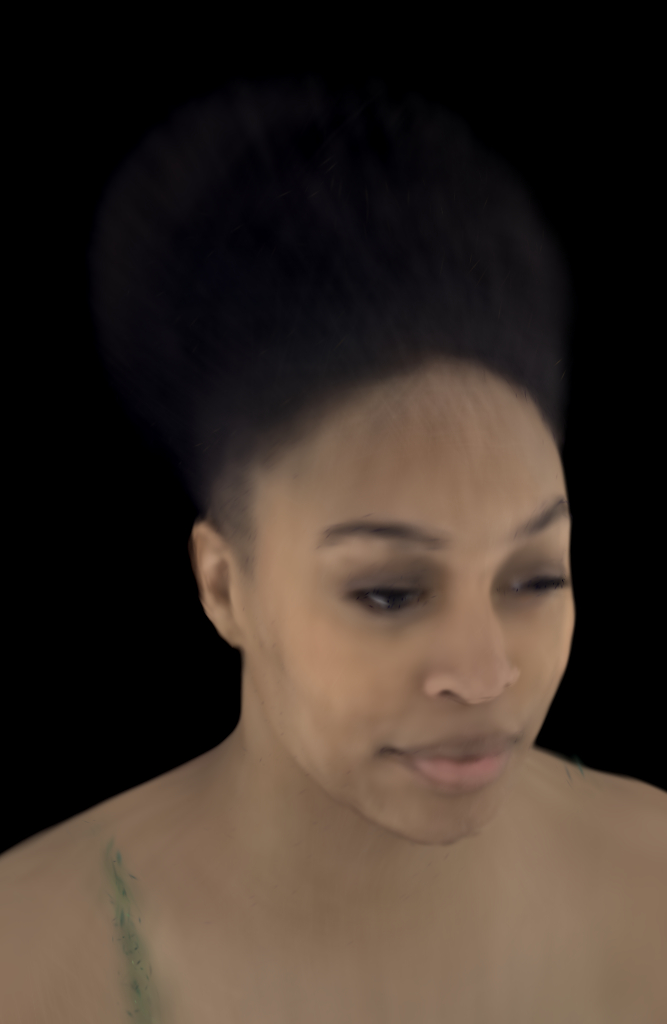} &
\includegraphics[height=0.14\linewidth]{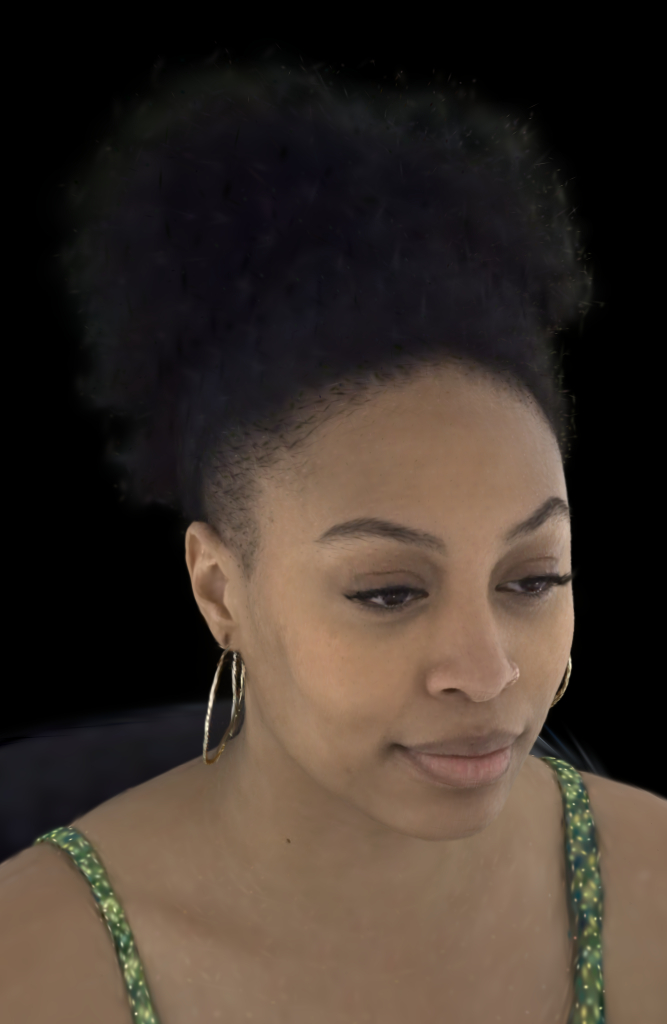} &
\includegraphics[height=0.14\linewidth]{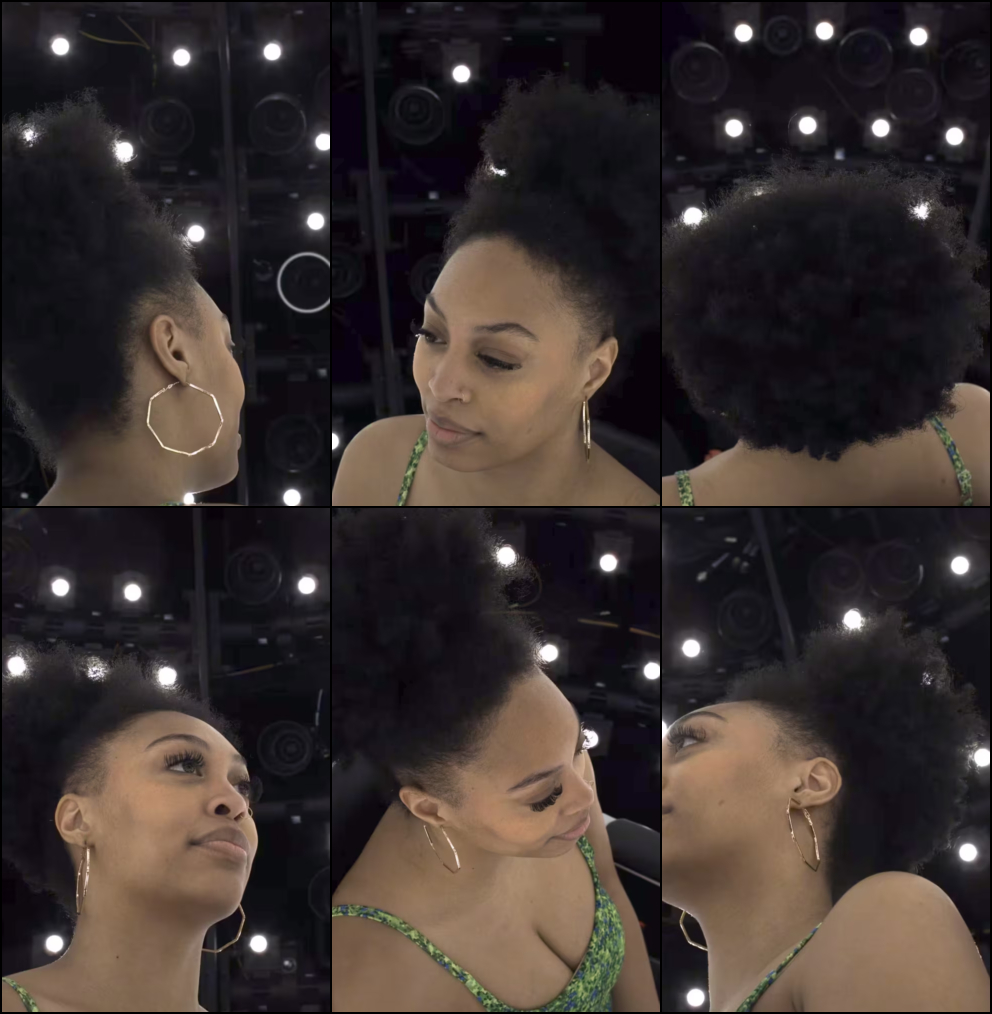} &
\includegraphics[height=0.14\linewidth]{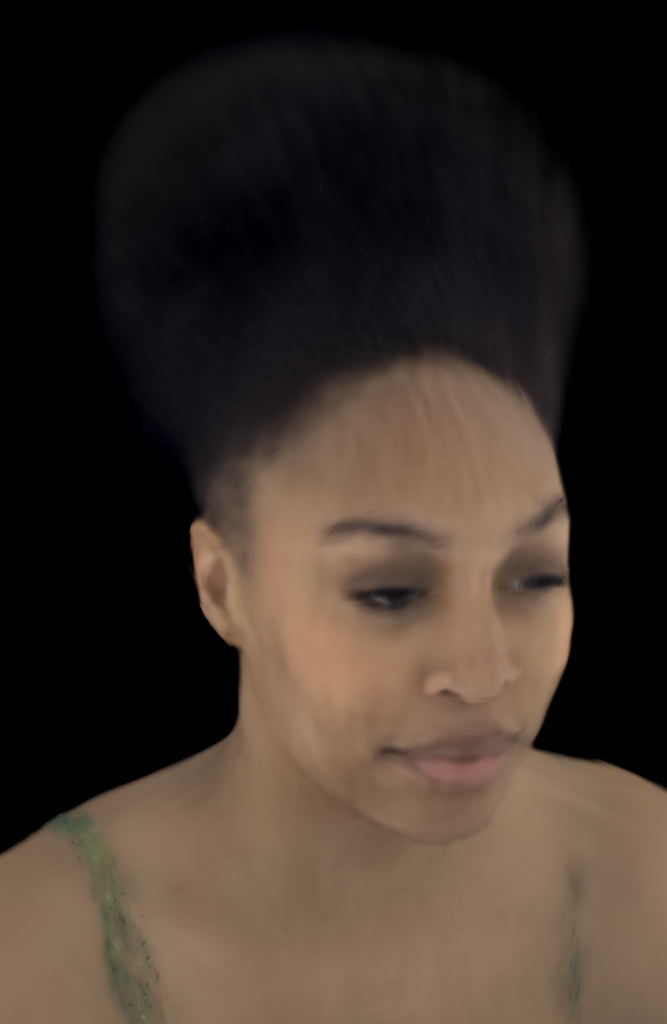} &
\includegraphics[height=0.14\linewidth]{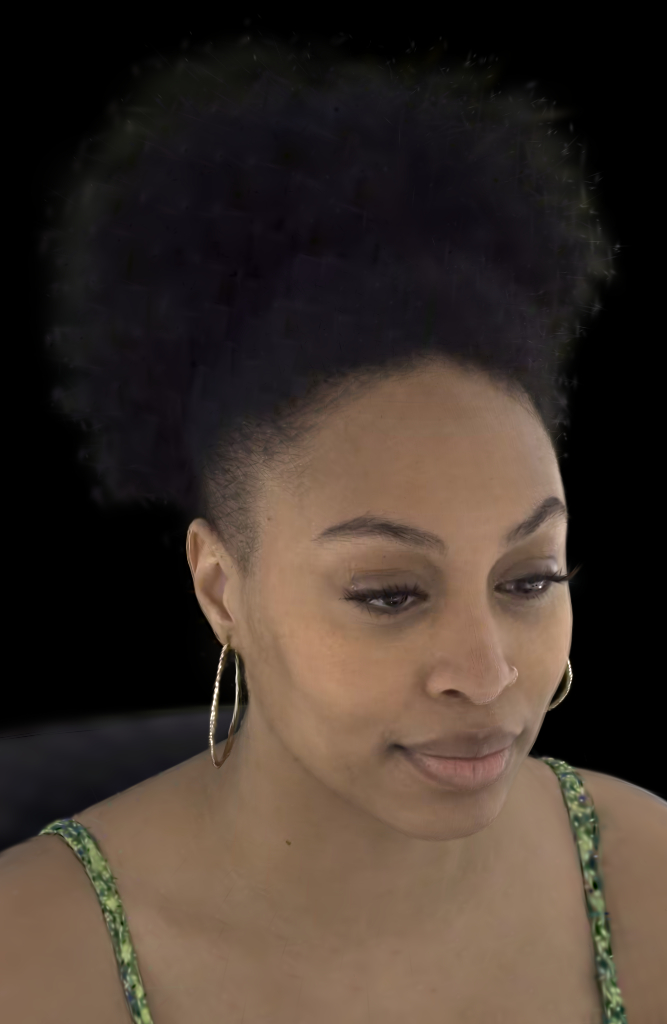} &
\includegraphics[height=0.14\linewidth]{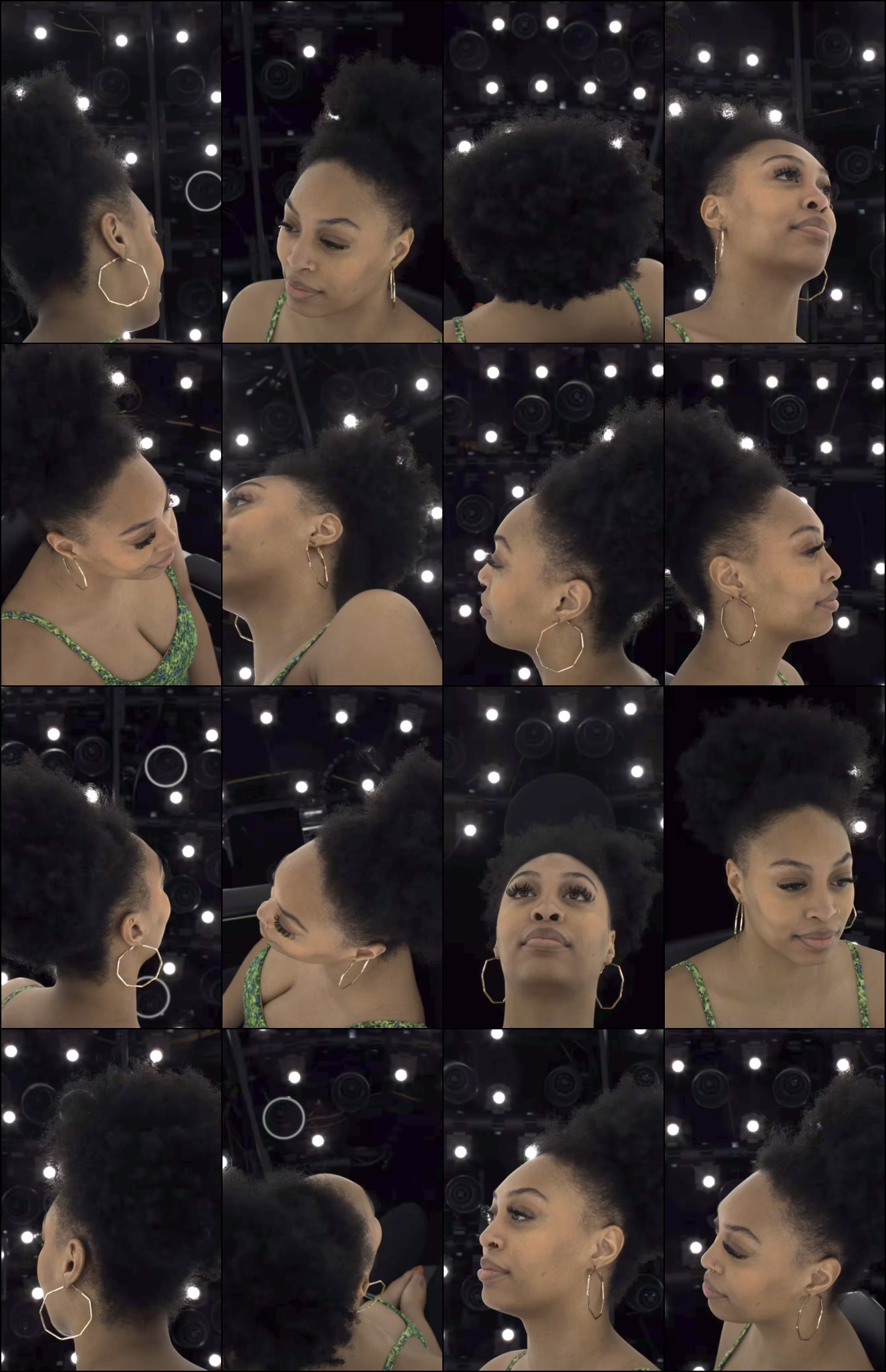} &
\includegraphics[height=0.14\linewidth]{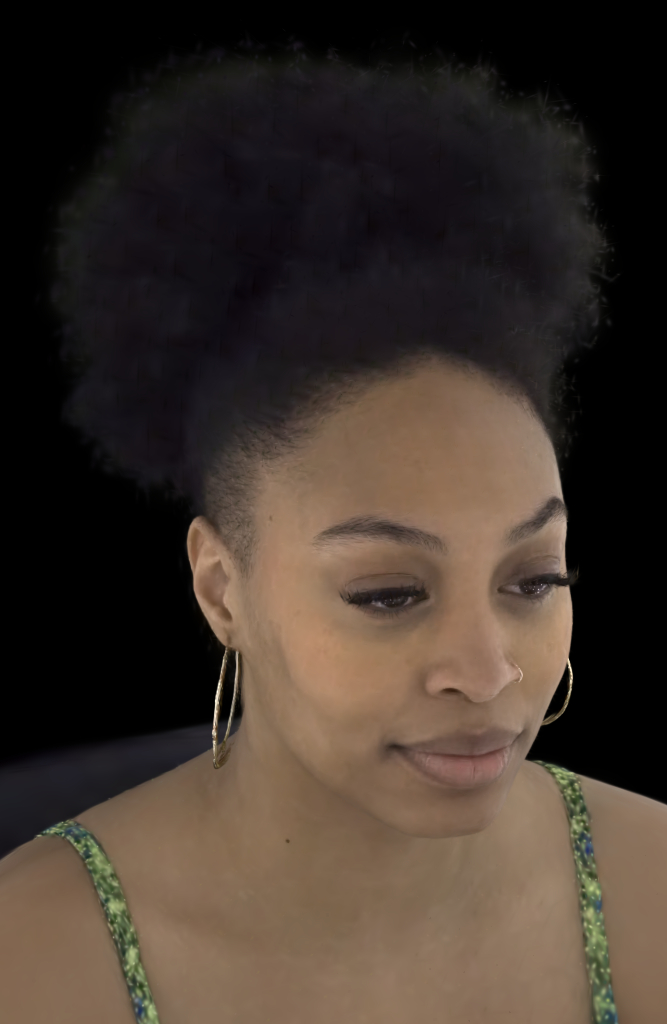} &
\includegraphics[height=0.14\linewidth]{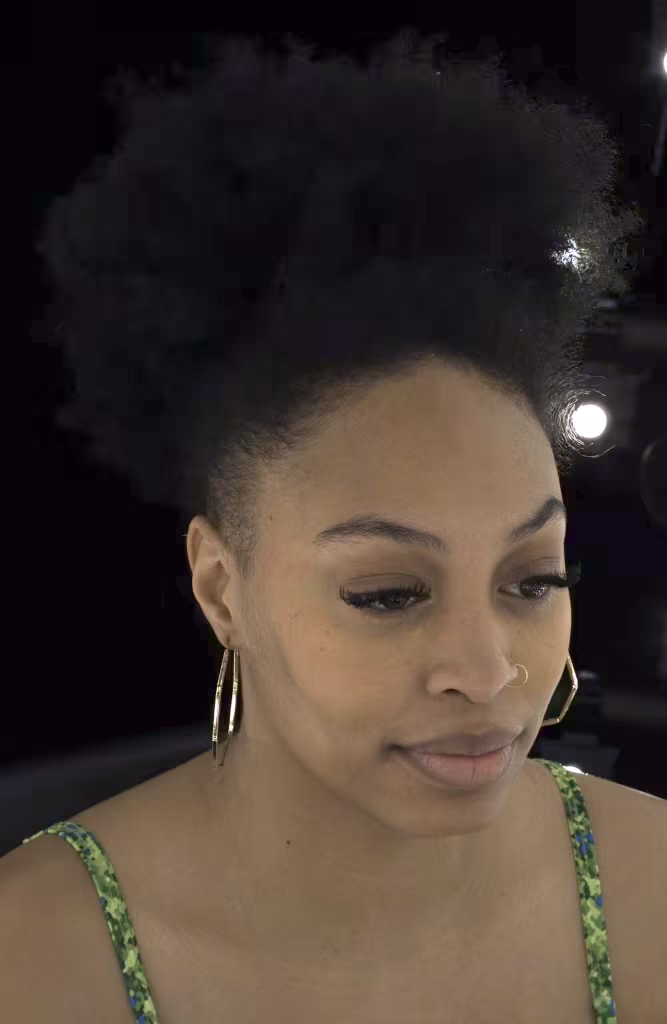} \\
\end{tabular}%
}
\caption{\textbf{Visual Comparison on Ava-256}. Our method produces sharper reconstructions with better identity preservation compared to prior work. Increasing the number of views permits reconstruction of details like earrings, hair and skin texture. Additionally, our background model successfully captures intricate head-boundary details that previous foreground-masking techniques typically discard; we only use the background model during training. Due to GPU memory constraints during training, Avat3r is limited to a maximum of 6 views.
}
\label{fig:4_ava256_renders}
\end{figure*}

%% file: tables/4_ava256_results.tex
\begin{table*}[t]
  \begin{minipage}[c]{0.54\textwidth}
    \fontsize{6.5pt}{8pt}\selectfont
    \setlength{\tabcolsep}{1.5pt}
    \setlength{\aboverulesep}{0pt}
    \setlength{\belowrulesep}{0pt}
    \renewcommand{\arraystretch}{1.25}

    \begin{tabular}{lrc ccccc}
    \toprule
    \rowcolor{tableheadergray}
    \multicolumn{8}{c}{\textbf{Internal10K}} \\
    \rowcolor{tableheadergray}
    & \#V & \#G
    & PSNR$\uparrow$ & SSIM$\uparrow$ & LPIPS$\downarrow$
    & AKD$\downarrow$ & CSIM$\uparrow$ \\
    \midrule
    Avat3r & 4 & 0.8M & 24.10 & 0.830 & 0.371 & 12.78 & 0.738 \\
    Avat3r & 6 & 1.1M & 24.37 & \textbf{0.831} & 0.362 & 12.99 & 0.787 \\
    \midrule 
    Ours & 10 & 65K & \textbf{29.25} & 0.821 & \textbf{0.117} & \textbf{10.39} & \textbf{0.922} \\
    \bottomrule
    \end{tabular}

    \vspace{10pt}

    \begin{tabular}{lrc ccccc}
    \toprule
    \rowcolor{tableheadergray}
    \multicolumn{8}{c}{\textbf{Ava-256}} \\
    \rowcolor{tableheadergray}
    & \#V & \#G
    & PSNR$\uparrow$ & SSIM$\uparrow$ & LPIPS$\downarrow$
    & AKD$\downarrow$ & CSIM$\uparrow$ \\
    \midrule
    Avat3r & 4 & 0.8M & 21.54 & 0.788 & 0.370 & 6.39 & 0.698 \\
    Avat3r & 6 & 1.1M & 22.31 & 0.795 & 0.359 & 6.10 & 0.775 \\
    \midrule
    Ours & 4  & 65K & 23.64 & 0.805 & 0.178 & 5.09 & 0.833 \\
    Ours & 6  & 65K & 24.62 & 0.815 & 0.161 & 4.70 & 0.873 \\
    Ours & 16 & 65K & \textbf{26.13} & \textbf{0.831} & \textbf{0.110} & \textbf{4.27} & \textbf{0.914} \\
    \bottomrule
    \end{tabular}
  \end{minipage}%
  \hfill
  \begin{minipage}[c]{0.45\textwidth}
    \small
    \caption{\textbf{Quantitative Comparison.} We evaluate both our method and Avat3r~\cite{kirschstein2025avat3r} on two datasets: our Internal10K dataset, and Ava-256~\cite{martinez2024codec} (fine-tuned from the Internal10K models), across varying numbers of input views (\#V). Our approach significantly outperforms the baseline with major gains in rendering quality (PSNR, LPIPS) and facial fidelity (AKD, CSIM) metrics. Notably, we achieve these improvements while requiring substantially fewer 3D Gaussians (\#G). Also, Avat3r is bottle-necked by GPU memory limits at 6 views on our training GPU budget.}\label{tab:experiments}
  \end{minipage}
\end{table*}

%% file: 04_02_ablations.tex
\subsection{Analysis and Ablation Study}
\label{subsubsec:ablation-study}

We ablate the proposed components of our overall method in~\cref{fig:ablations}.
We conduct extensive analysis to validate our design choices and characterize the scaling behavior of our method. 
We organize our analysis around five key aspects: number of training identities, number of input views, representation capacity, number of target views used for supervision, and our high-resolution finetuning strategy.
Moreover, 
we study the effect of the background model, region-specific perceptual losses, and multi-resolution loss.

\input{figures/4_ablation_all}

\input{figures/4_ablation_num_subjects}
\input{figures/4_figures_ablation_num_encoder_views}

\subsubsection{Number of training identities.}
As depicted in~\cref{fig:ablations}a, PSNR improves 1.7--1.8~dB per doubling of subjects up to 2K, with reduced performance gains after 4K. Identity preservation (CSIM) scales more steeply, from 0.441 (250 subjects) to 0.888 (8K). Models trained on $<$1K subjects fail catastrophically on out-of-distribution faces. Visual comparisons are provided in \cref{fig:data_scaling}.

\subsubsection{Number of input views.}
We train our model with an increasing number of random input views. For evaluation, we use a fixed set of views with varying size per experiment, e.g. the frontal view for the monocular case. As shown in~\cref{fig:ablations}b and~\cref{fig:view_scaling}, more views improve the quality, with diminishing gains after 8 views. Notably, our monocular model is able to convincingly reconstruct the subject.

\subsubsection{Latent and Gaussian UV resolution.}
As shown in~\cref{fig:ablations}c, increasing the latent resolution yields more significant improvements than the UV resolution.

\subsubsection{Template mesh type.} Our ablation in~\cref{fig:ablations}d shows that using a fixed neutral template significantly outperforms using an expression-tracked one.

\subsubsection{Number of target views.}
Adding more supervision views improves all the metrics by encouraging better view-consistency as shown in \cref{fig:ablations}e.

\subsubsection{High-resolution finetuning.}
As discussed in~\cref{sec:highres_finetuning}, after our low-resolution training has converged, we finetune our model on high-resolution target and input images. In \cref{fig:ablations}f we examine our design choices for the second training stage:

\noindent\emph{Region-specific losses.}
Our eye and mouth crop losses improve reconstruction quality in critical facial features where human perception is most sensitive.

\noindent\emph{Finetuning perceptual losses.}
Downsampling the rendered and real inputs to the Discriminator and LPIPS losses is important for the stability of the adversarial losses. Naively computing LPIPS on high resolution images leads to quality regressions.

\subsubsection{Background Gaussian model.}

As shown in \cref{fig:ablations}g, removing the background model degrades rendering quality for regions with semi-transparent areas or fine foreground elements. Without explicit background modeling, the model hallucinates background artifacts, such as discoloration in the hair. Since computed foreground masks are not pixel-perfect and view-consistent, background details are often combined with foreground elements within masked ground truth images used for supervision. By eliminating the need for these masks, our dedicated background model prevents artifacts from background regions, such as the multi-camera capture rig, and allows the foreground model to focus more Gaussians on facial details.

%% file: figures/4_ablation_all.tex
\begin{figure*}[t]
\centering
\begin{minipage}[c]{0.66\textwidth}
    \centering
    \includegraphics[width=\linewidth]{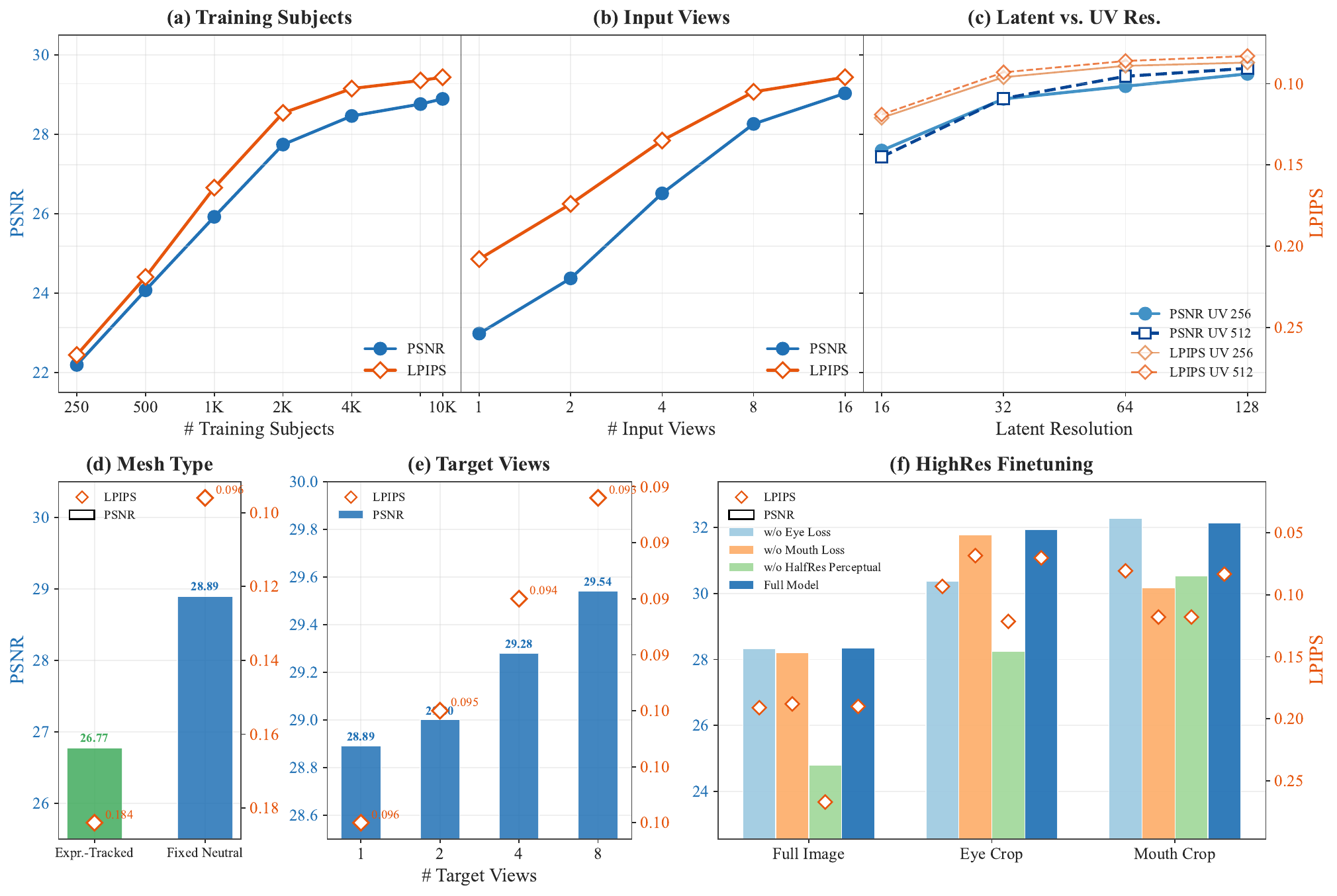}
\end{minipage}%
\hfill%
\begin{minipage}[c]{0.34\textwidth}
    \centering
                \vspace{-3mm}
        \caption*{\tiny\textbf{(g)} Background model ablation}
            \vspace{-3mm}
    \includegraphics[trim={0cm 1.1cm 34cm 0},clip,width=\linewidth]{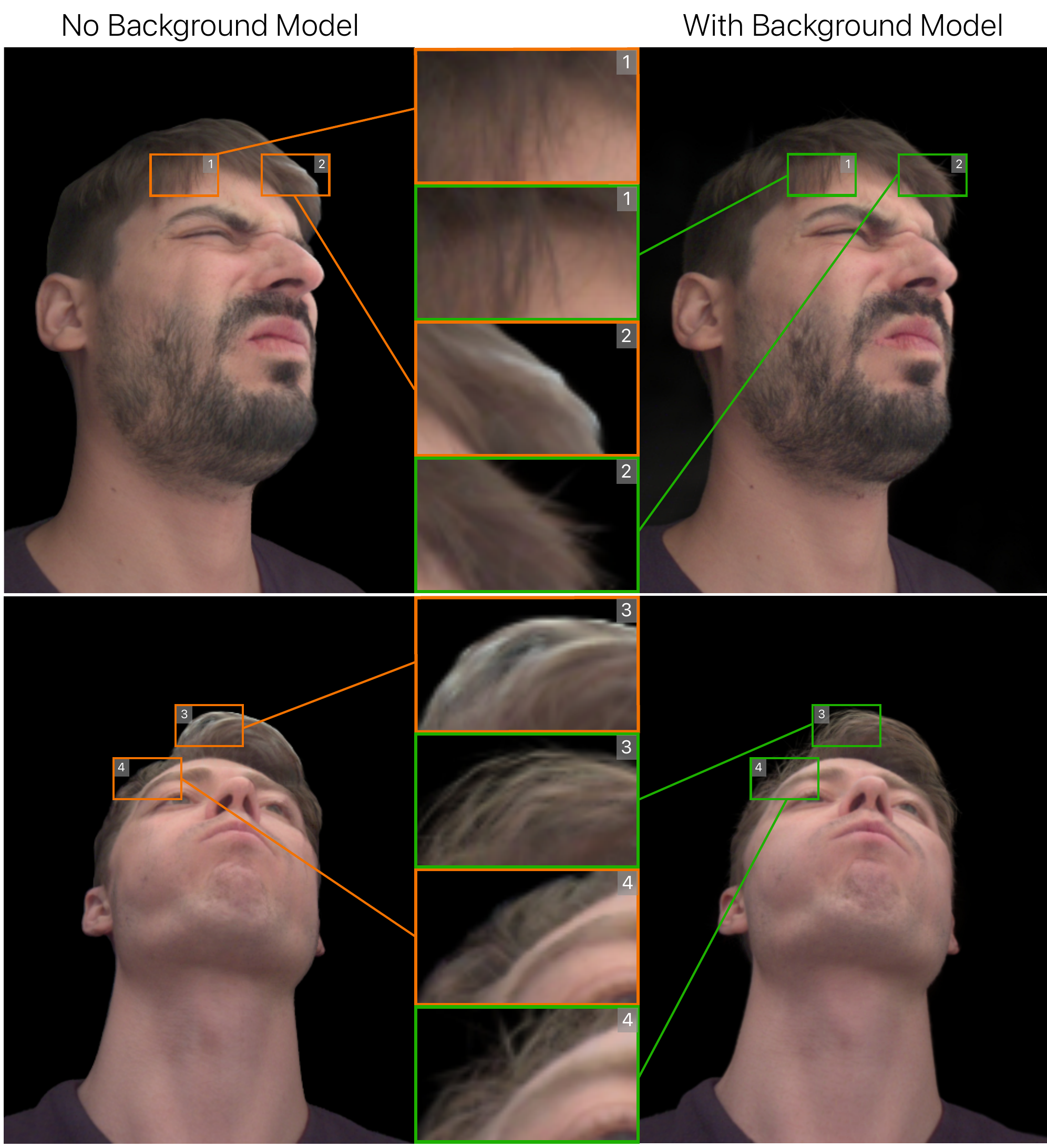}    %
\end{minipage}
\caption{~\textbf{Ablation study} on a single-stage model trained for 500K steps with 10K subjects,
10 input views, $32{\times}32$ latent and $256{\times}256$ Gaussian UV resolution unless stated otherwise.
\textbf{(a)}~\textbf{Training data scaling}: log-linear improvement up to 2K subjects, diminishing returns beyond 4K.
\textbf{(b)}~\textbf{Input view scaling}: quality improves with more views, with diminishing returns after 8.
\textbf{(c)}~\textbf{Model capacity}: increasing latent resolution yields larger gains than increasing Gaussian UV resolution.
\textbf{(d)}~\textbf{Template mesh type}: a fixed neutral mesh outperforms expression-tracked meshes.
\textbf{(e)}~\textbf{Number of target views}: more supervision views improve geometric consistency
\textbf{(f)}~\textbf{High-resolution finetuning}: Effect of different components in our second-stage training strategy.
\textbf{(g)}~\textbf{Background model ablation}: Our background modeling permits reconstruction of foreground details such as strands of hair, without background artifacts caused by imperfect image matting techniques.}
\label{fig:ablations}
\end{figure*}

%% file: figures/4_ablation_num_subjects.tex
\begin{figure*}[t]
\centering
\setlength{\tabcolsep}{0pt}
\renewcommand{\arraystretch}{0}
\footnotesize
\begin{tabular}{@{}ccccccc@{}}
& 250 & 500 & 1K & 4K & 10K & GT \\[4pt]
\rotatebox{90}{\hspace{0.5em}} &
\includegraphics[width=0.164\linewidth]{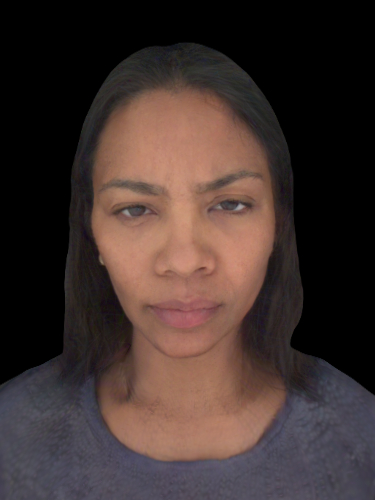} &
\includegraphics[width=0.164\linewidth]{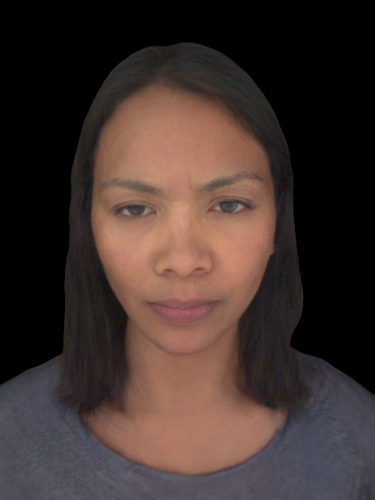} &
\includegraphics[width=0.164\linewidth]{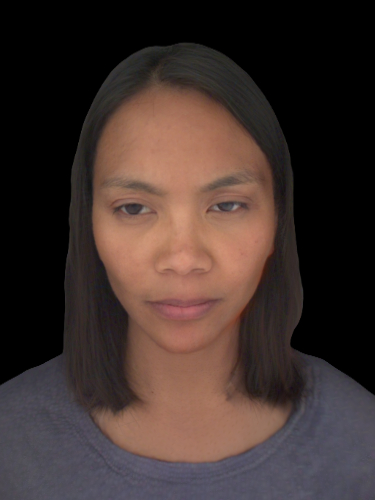} &
\includegraphics[width=0.164\linewidth]{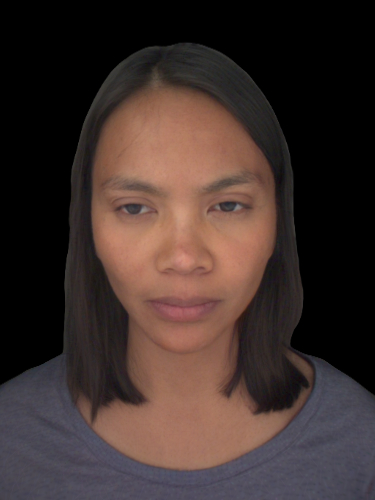} &
\includegraphics[width=0.164\linewidth]{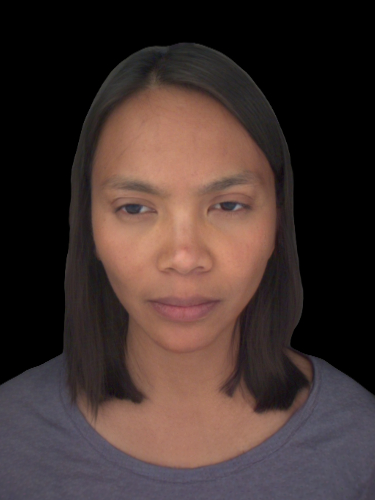}\hspace{1mm} &
\includegraphics[width=0.164\linewidth]{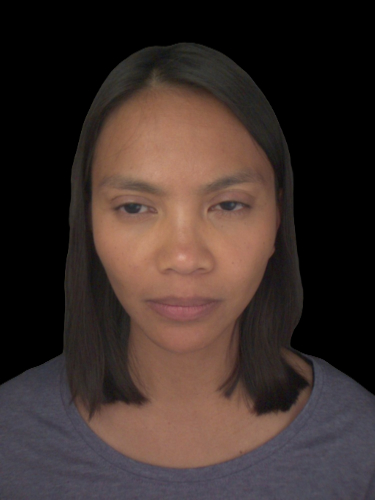} \\
\rotatebox{90}{\hspace{0.5em}} &
\includegraphics[width=0.164\linewidth]{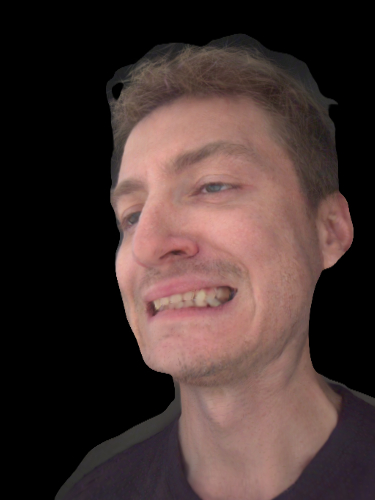} &
\includegraphics[width=0.164\linewidth]{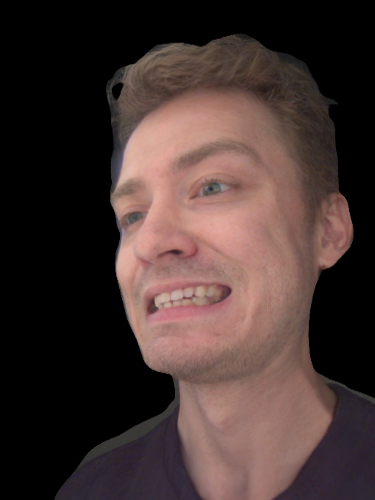} &
\includegraphics[width=0.164\linewidth]{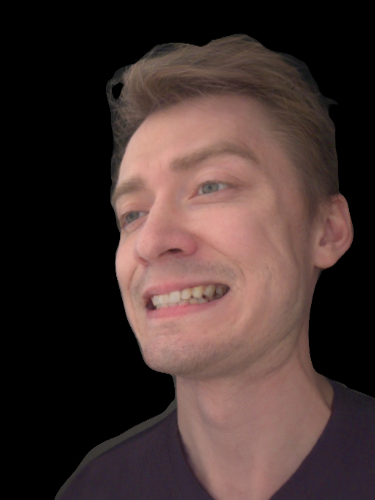} &
\includegraphics[width=0.164\linewidth]{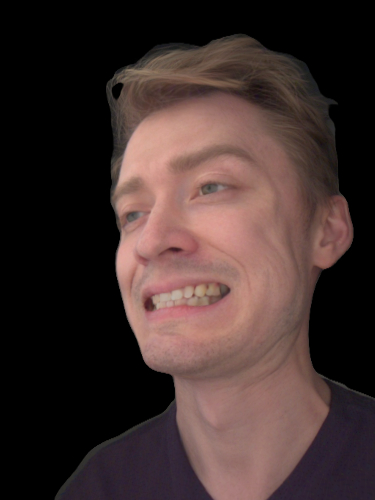} &
\includegraphics[width=0.164\linewidth]{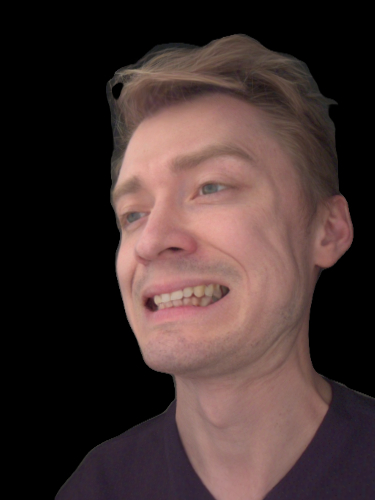}\hspace{1mm} &
\includegraphics[width=0.164\linewidth]{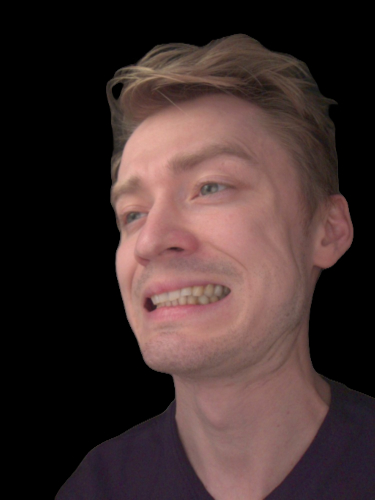} \\
\end{tabular}
\caption{\textbf{Training data scaling}. Models trained on fewer subjects fail to generalize to reconstruction of novel identities. At 250 subjects, facial features and hair color deviate significantly from ground truth. The reconstruction quality improves with more training data. On this validation set, the quality improves marginally after $4$K subjects. }
\label{fig:data_scaling}
\end{figure*}

%% file: figures/4_figures_ablation_num_encoder_views.tex
\begin{figure*}[h!]
\centering
\setlength{\tabcolsep}{0pt}
\renewcommand{\arraystretch}{0}
\footnotesize
\resizebox{\linewidth}{!}{%
\begin{tabular}{@{}ccccccc@{}}
$N=1$ & $N=2$ & $N=4$ & $N=6$ & $N=8$ & $N=16$ & GT \\[4pt]
\includegraphics[width=0.135\linewidth]{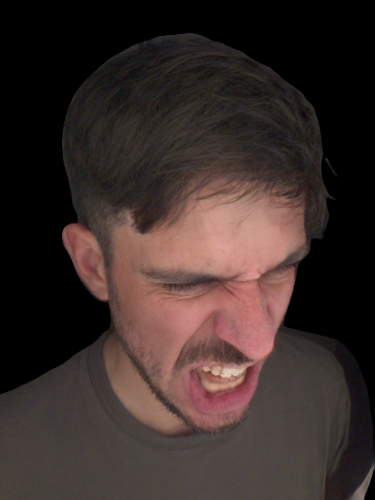} &
\includegraphics[width=0.135\linewidth]{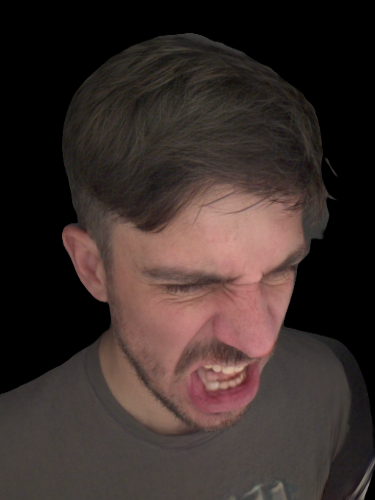} &
\includegraphics[width=0.135\linewidth]{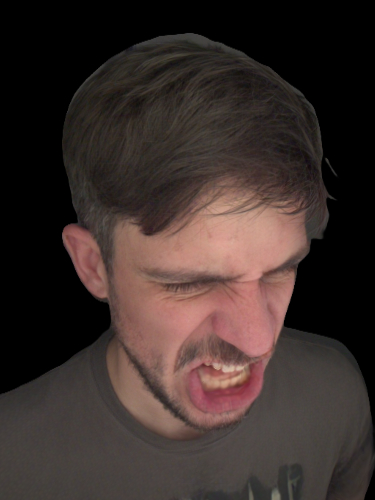} &
\includegraphics[width=0.135\linewidth]{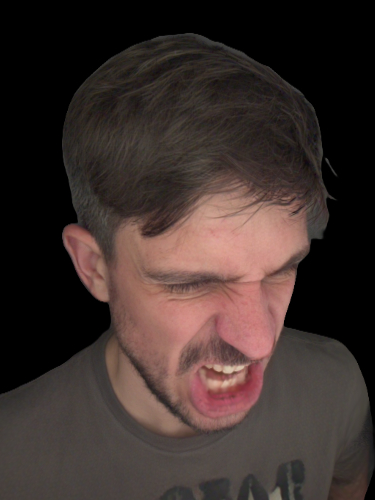} &
\includegraphics[width=0.135\linewidth]{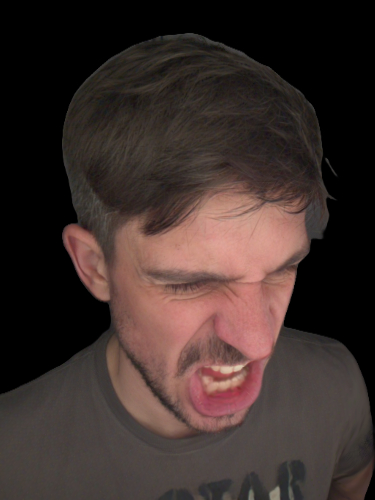} &
\includegraphics[width=0.135\linewidth]{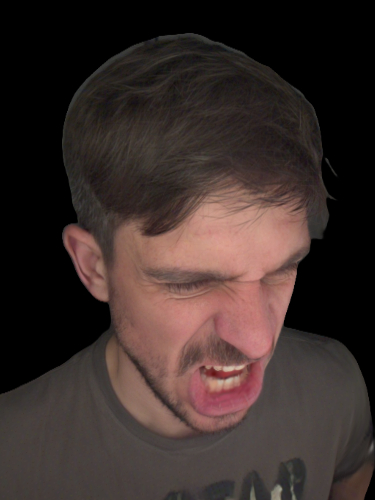} &
\includegraphics[width=0.135\linewidth]{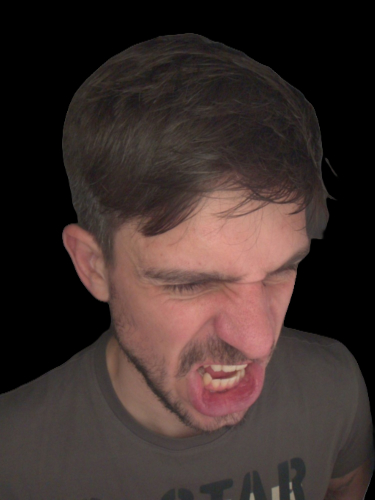} \\
\includegraphics[width=0.135\linewidth]{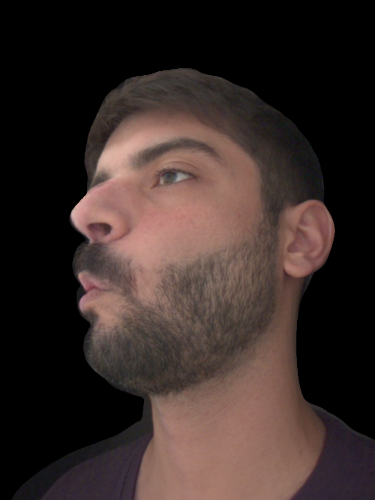} &
\includegraphics[width=0.135\linewidth]{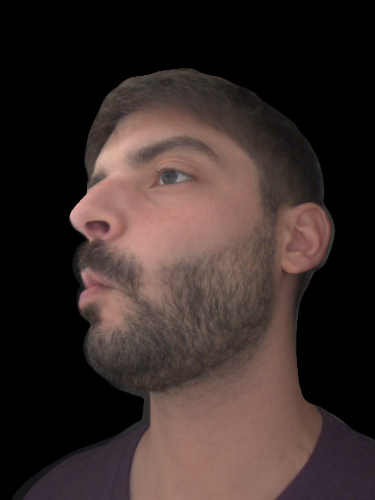} &
\includegraphics[width=0.135\linewidth]{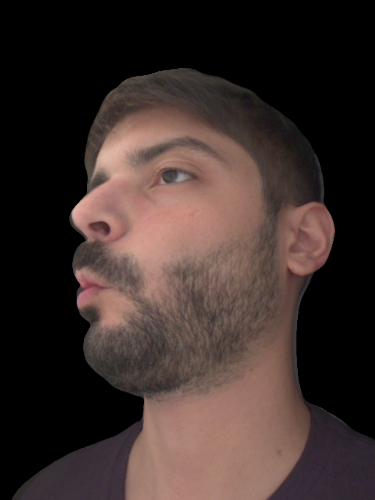} &
\includegraphics[width=0.135\linewidth]{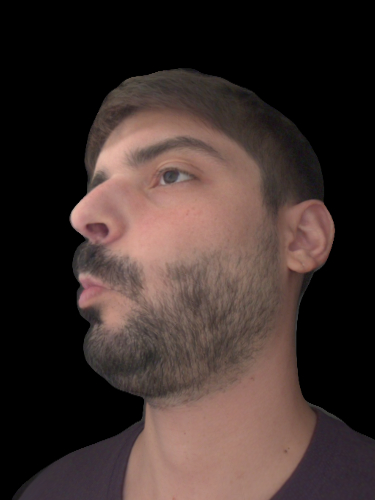} &
\includegraphics[width=0.135\linewidth]{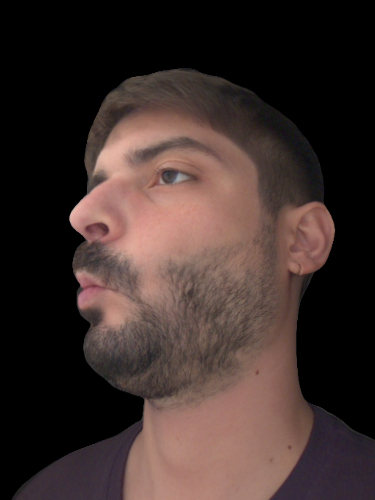} &
\includegraphics[width=0.135\linewidth]{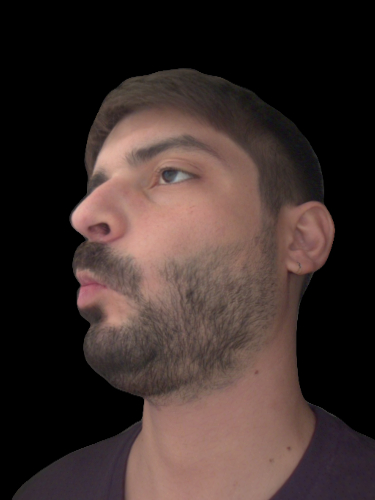} &
\includegraphics[width=0.135\linewidth]{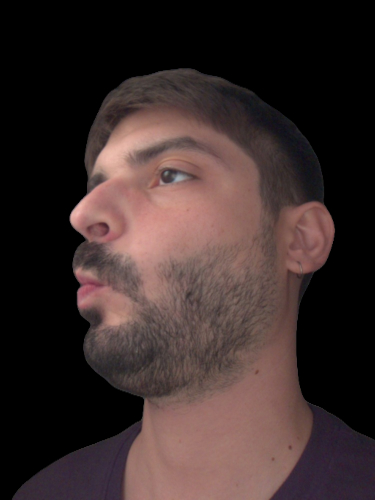} \\
\end{tabular}}
\vspace{-2mm}
\caption{\textbf{Impact of the number of input views.} Reconstruction quality scales naturally with the number of input images. A single frontal view ($N=1$) yields blurry results, identity drift, and fails to recover shirt text. However, adding more views progressively resolves these ambiguities, yielding clear improvements in fine details like the teeth and hair.
\label{fig:view_scaling}}
\end{figure*}

%% file: figures/4_downstream_applications.tex
  \begin{figure*}[t]
  \centering
  \begin{subfigure}[b]{0.48\linewidth}
  \centering      
  \includegraphics[width=\linewidth]{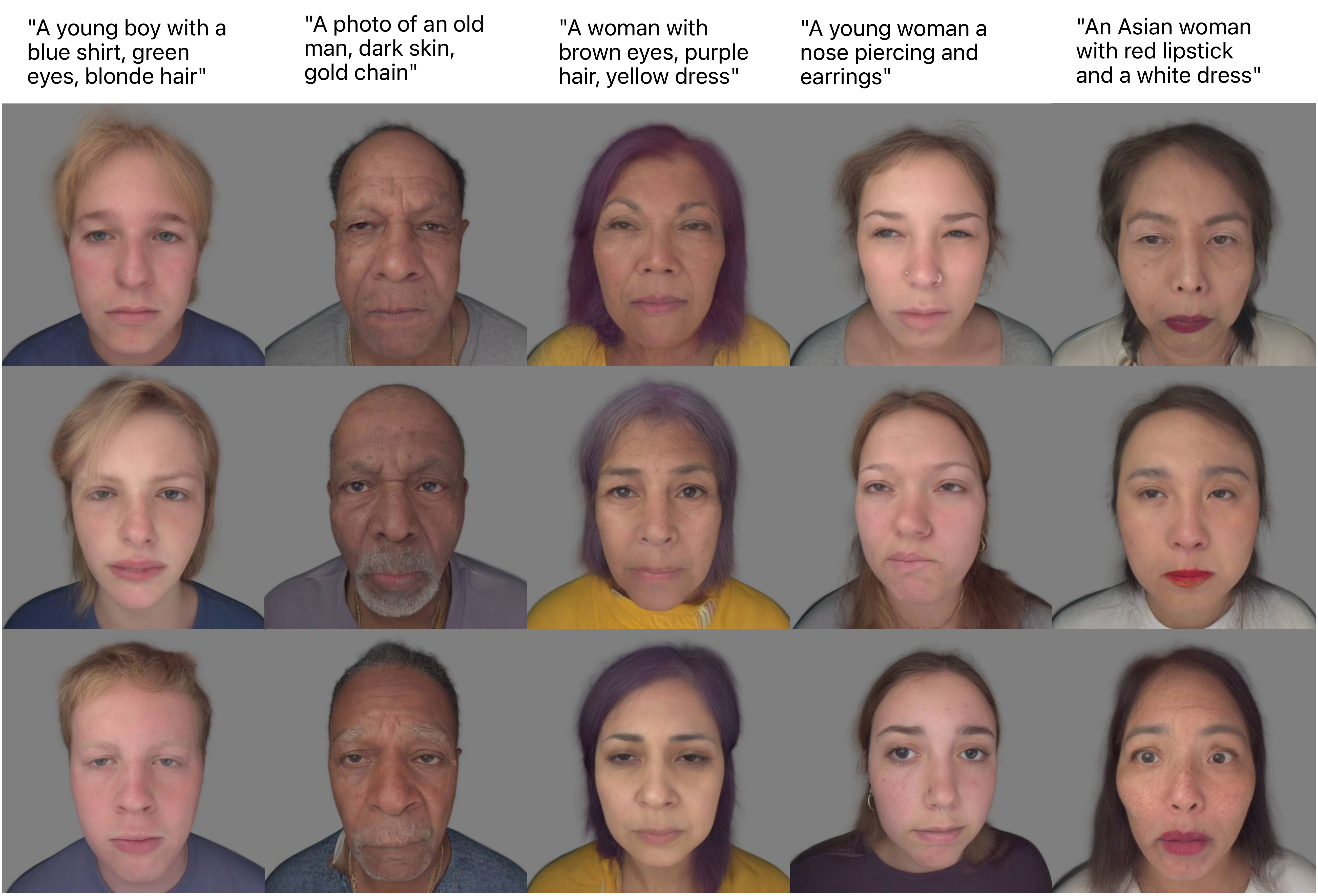}
  \subcaption{Text-driven identity generation \label{subfig:novel-identities}}
  \end{subfigure}%
  \hfill
  \begin{subfigure}[b]{0.48\linewidth}
  \centering
  \includegraphics[width=\linewidth,trim={0 0.5cm 14cm
  0},clip]{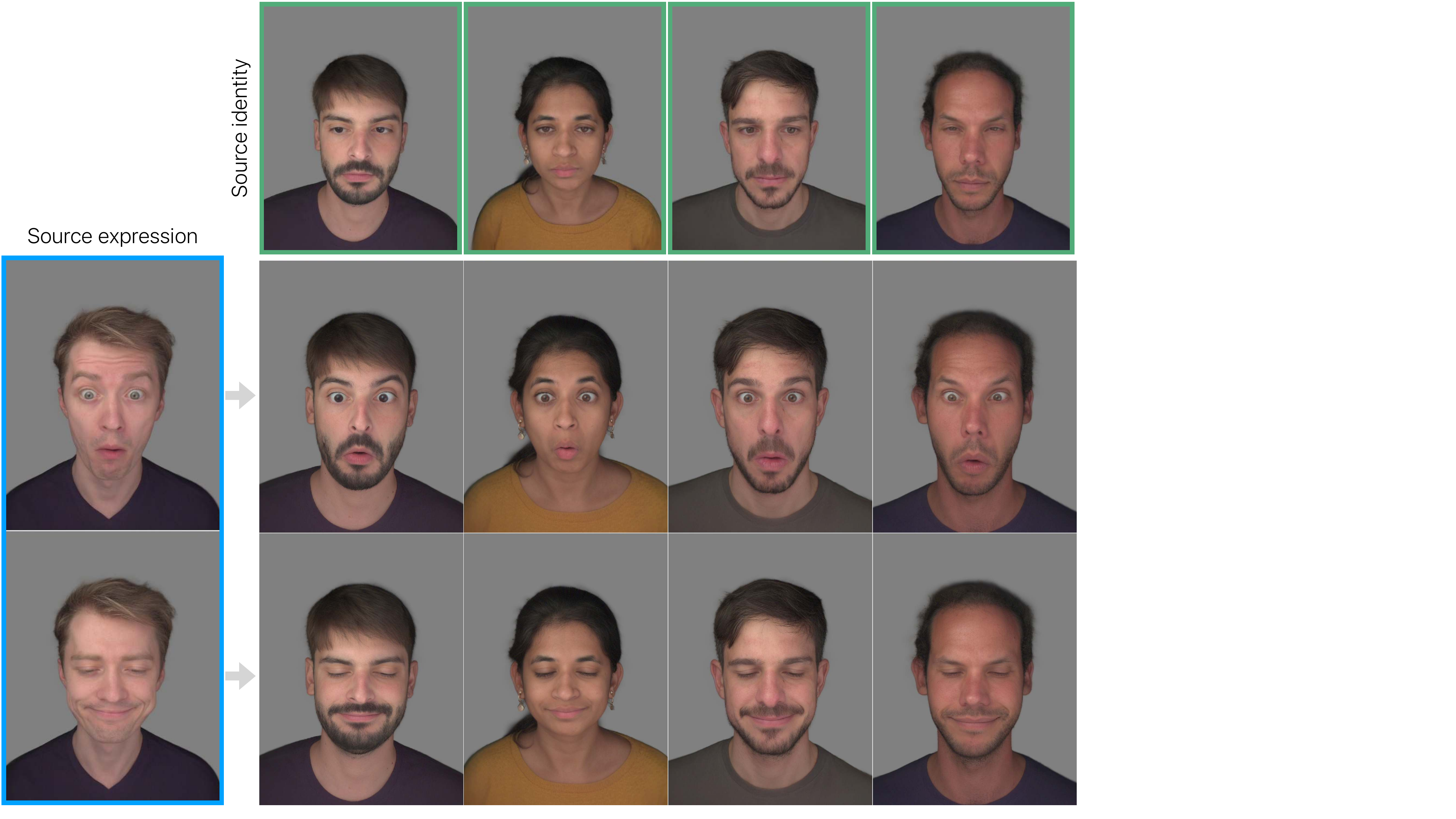}
  \subcaption{Blendshape-driven latent animation \label{subfig:blendshape-rigging}}
  \end{subfigure}
  \caption{\textbf{Downstream Applications.} \textbf{(a) Text-driven identity generation:} Novel identities generated by a text-conditioned diffusion model trained on our latents. Using nearest-neighbor face similarity, we verify that these synthesized subjects do not exist in our training set. \textbf{(b) Blendshape-driven latent animation:} We train a network conditioned on expression blendshapes, applying a target expression (blue box) to a source identity (green box). The model successfully animates the faces while preserving the subject's appearance. Both applications operate entirely within the latent space, requiring no per-subject fine-tuning.\label{fig:downstream_applications}}
  \end{figure*}

%% file: 05_conclusion.tex
\section{Conclusion\label{sec:conclusion}}

We present \emph{HeadsUp}, a highly scalable feed-forward approach for state-of-the-art reconstruction of 3D Gaussian heads from multi-camera studio captures. By anchoring a compact set of 3D Gaussians to a neutral UV template and employing a lightweight cross-attention transformer, our method achieves photorealistic rendering while gracefully scaling with respect to the number of input views. Crucially, our explicit background modeling eliminates the reliance on imperfect segmentation masks. Combined with a two-stage training strategy and region-specific perceptual losses, this enables the faithful reconstruction of challenging, high-frequency details such as hair strands and jewelry. We show that \emph{HeadsUp} successfully scales to a production-level dataset of \num{10000} unique identities, delivering state-of-the-art reconstruction quality and exhibiting robust generalization to unseen subjects. Finally, beyond feed-forward reconstruction, we show that our learned latent space enables downstream generative applications, including the text-driven synthesis of novel identities and blendshape-driven latent animation.

Additional details on the methodology and experimental setup, along with qualitative results including rendered images and videos, are provided in the supplementary material.

%% file: 06_supplementary.tex
\clearpage

\section*{Supplementary Material}

\renewcommand{\thesection}{\Alph{section}}
\renewcommand{\thesubsection}{\Alph{section}.\arabic{subsection}}
\renewcommand{\thefigure}{\Alph{section}.\arabic{figure}}
\renewcommand{\thetable}{\Alph{section}.\arabic{table}}
\renewcommand{\theequation}{\Alph{section}.\arabic{equation}}
\setcounter{section}{0}
\setcounter{figure}{0}
\setcounter{table}{0}
\setcounter{equation}{0}

\noindent In this Supplementary Material, we provide additional results, implementation details, and information on our training procedures. We encourage the reader to view the HeadsUp videos on our \href{https://apple.github.io/ml-headsup/}{project webpage}.
  
\section{Inference Speed}
\label{sec:runtime_comparison}

We analyze the reconstruction time and scalability of our approach compared to the Avat3r baseline given a varying number of input images, as detailed in~\cref{tab:supplementary-inference-speed}. Because the baseline architecture struggles with the dense aggregation of multi-view features, their processing time increases drastically as more views are added. Avat3r operates at sub-second frame rates for \num{4} and \num{6} views, and completely fails due to Out-of-Memory (OOM) errors during training when the input exceeds 6 views. Note that in
the reported results for Avat3r, while the Sapiens~\cite{khirodkar2024sapiens} feature maps are computed on-the-fly, the VGGT~\cite{wang2025vggt} position maps are computed offline.

Conversely, our method processes spatial features much more efficiently. While our reconstruction time naturally scales with the number of input views, it remains orders of magnitude faster than the baseline. Furthermore, our memory-efficient model easily accommodates 10 or more views during both training and inference without exhausting GPU memory, allowing for higher-fidelity reconstructions without the computational bottleneck. Moreover, our method does not rely on large pretrained backbones like DUSt3R, Sapiens, or DINOv3, as opposed to the recent feedforward reconstruction methods~\cite{li2024uravatar, prinzler2026match,ji2026fastgha}. 

To highlight the efficiency of our method compared to the large-scale video generation models: HeadsUp predicts 3D Gaussians from 16 views in $0.33$ seconds on a single A100 using only $82$M parameters. Additionally, our blendshape-driven latent animation (discussed in~\cref{subsubsec:blendshape_rigging}) runs at 65 FPS on a MacBook Pro. In comparison, Wan$2.2$ (14B parameters) and FlashWorld~\cite{li2025flashworld} (9B parameters) take $2810$ and $9$ seconds for inference, respectively, making them significantly slower than our method.

\begin{table}[b]
\centering
\caption{Inference speed comparison on a single A100 GPU for predicting 3D Gaussians from multi-view images. Our method is more than an order of magnitude faster than Avat3r,
enabling the 3D reconstruction of large-scale datasets. Note that in the reported results for Avat3r, the VGGT position maps are computed offline.
\label{tab:supplementary-inference-speed}}
\setlength{\tabcolsep}{6pt}
{%
\setlength{\aboverulesep}{0pt}%
\setlength{\belowrulesep}{0pt}%
\renewcommand{\arraystretch}{1.2}%
\begin{tabular}{lcccccc}
  \toprule
  \rowcolor{tableheadergray}
   & \multicolumn{3}{c}{\textbf{FPS} $\uparrow$} & \multicolumn{3}{c}{\textbf{Reconstruction Time (s)} $\downarrow$} \\
  \rowcolor{tableheadergray}
   & 4 Views & 6 Views & 16 Views & 4 Views & 6 Views & 16 Views \\
  \midrule
  Avat3r & 0.18 & 0.09 & OOM & 5.6 & 10.8 & OOM \\
  Ours   & \textbf{7.14} &\textbf{4.94} & \textbf{2.93} & \textbf{0.14} & \textbf{0.20} & \textbf{0.33} \\
  \bottomrule
\end{tabular}%
}
\end{table}

\section{Additional Baseline Details and Comparisons}
\label{sec:additional_baseline_comparisons}
We document our reimplented baselines and present an additional comparison against Avat3r~\cite{kirschstein2025avat3r} and FastGHA~\cite{ji2026fastgha} using the Internal10K dataset in~\cref{tab:supp_avat3r_fastgha_headsup_4views} and~\cref{fig:supp_avat3r_fastgha_headsup_4views}.
Our results show significantly improved rendering and facial fidelity.

\input{tables/supp_avat3r_fastgha_headsup_4views}

\subsection{Avat3r}
\label{sec:supplementary-avat3r_reimplementation}
As no official code is available, we reimplement Avat3r~\cite{kirschstein2025avat3r}, adapting it from its original sparse-view self-reenactment setting (256 subjects, 10 input frames each) to our large-scale multi-view setup (10K+ subjects, 100 input frames each). 

\paragraph{Modifications.} We make three key changes: (1) Removed expression rigging: We drop the cross-attention blocks for expression latents, as our setting provides time-synchronized multi-view images of the target expression. (2) Replaced DUSt3R with VGGT: Precomputing DUSt3R~\cite{wang2024dust3r} position maps at our scale (10K subjects $\times$ 100 frames $\times$ 16 views) is prohibitively expensive. We use VGGT~\cite{wang2025vggt}, which is faster and produces improved geometry. (3) Replaced Sapiens-2B with 1B: Precomputing Sapiens~\cite{khirodkar2024sapiens} requires 100\,TB of storage for Internal10K alone. Instead, we compute Sapiens-1B on-the-fly during training. It performs comparably to $2$B at half the compute cost.

\paragraph{Position Map Geometry Prior.} We align the normalized VGGT point maps to metric scale via a $7$-DoF similarity transform \cite{umeyama1991}. This is computed per-frame using 2D facial landmarks projected onto the VGGT point cloud and their corresponding 3D tracked mesh vertices. The aligned maps are transformed into the head canonical frame using the head pose. During the forward pass, we drop VGGT confidence map pixels below the $10$th percentile to create a binary mask, which is applied to the predicted Gaussian positions as in Avat3r.

\paragraph{Encoder View Sampling.} We adapt the two-step input view sampling of Avat3r by fixing $N_\text{candidate}=16$ diverse (not strictly frontal) cameras per frame during preprocessing to compute VGGT maps. During training, we uniformly subsample $N{=}4$ or $N{=}6$ views. This preserves viewpoint diversity while avoiding the cost of computing position maps for all possible view subsets. Unlike Avat3r we do not restrict the $N_\text{candidate}$ cameras to frontal-only cameras.

\paragraph{Training.} We faithfully match Avat3r's architecture, losses, and hyperparameters, adjusting only input resolution (full-resolution vs.\ $512 \times 512$ crops), batch size, and learning rate. Training proceeds in three stages: (1) $2\times$ downsampled Internal10K (batch size 64, $\lambda_\text{lpips}{=}0$), (2) full-resolution Internal10K (batch size 16, $\lambda_\text{lpips}{=}0.01$), and (3) full-resolution Ava-256 (batch size 16, $\lambda_\text{lpips}{=}0.01$). Models are trained until validation metrics plateau on 16 H100 GPUs.

\subsection{FastGHA}
FastGHA~\cite{ji2026fastgha} predicts pixel-aligned Gaussians with a feed-forward transformer conditioned on DINOv3~\cite{simeoni2025dinov3} and Stable Diffusion (SD) VAE features~\cite{sauer2024adversarial}.
We carefully reimplement FastGHA to the best of our ability as no code is released, subject to several changes which we mention here. 

\paragraph{Modifications.} We make five key changes: (1) To support learning task-specific features in addition to off-the-shelf features from pretrained foundational models, we extract $512 \times 54 \times 42$ feature maps from raw input images using a trained convolutional ResNet encoder. We concatenate these learned convolutional features with DINOv3, SD-VAE, and Pl\"ucker embeddings as input into the multi-view transformer. 
(2) We extract SD-VAE features using the model in~\cite{rombach2022high} instead of the SD-Turbo VAE in~\cite{sauer2024adversarial} used in the original FastGHA paper~\cite{ji2026fastgha}. 
(3) As our task does not require rigging the predicted Gaussians, we remove expression code conditioning and the respective shallow per-Gaussian deformation MLP. 
(4) Unlike Avat3r which integrates feature matching maps from DUSt3R as one of its inputs, FastGHA uses VGGT~\cite{wang2025vggt} position maps as geometric supervision in the form of a regularization. We empirically found this component to only marginally influence results and excluded it from full-scale trainings. 
(5) To improve convergence and generalization, after experimentation, we replace a frozen SD-VAE decoder with a learnable residual decoder. 

\paragraph{Encoder View Sampling.} We use the same two-step input view sampling as our Avat3r reimplementation (\cref{sec:supplementary-avat3r_reimplementation})

\paragraph{Training.}
Similarly to our reimplementation of Avat3r, for better convergence during the initial training epochs, we add a silhouette loss and a scale regularization.
We train our FastGHA reimplementation until validation metrics plateau on 16 H100 GPUs.

\input{figures/supp_avat3r_fastgha_headsup_4views}

\section{Additional Ablations}
\label{sec:additional_results}

\Cref{tab:supplementary-ablations} provides detailed numerical results for the ablation studies summarized in Fig. 4 of the main paper. All experiments use a single-stage model trained for 500K steps with 10K subjects, 10 input views, $32{\times}32$ latent and $256{\times}256$ Gaussian UV resolution unless stated otherwise.
  
\input{figures/6_ablation_avat3r_headsup_num_encoder_views}

\subsubsection{Number of Training Subjects.}
We study the effect of training data scale in~\cref{tab:supplementary-num-subjects}. All metrics improve log-linearly as the number of subjects increases from 250 to 2K, with PSNR rising by over 5\,dB across this range. Beyond 4K subjects, gains begin to saturate on this validation dataset: increasing from 4K to 10K yields a 0.43\,dB improvement in PSNR.

\subsubsection{Number of Input Views.}
\cref{fig:6_view_scaling_comparison} compares how our method and Avat3r scale with the number of input views. Avat3r's reliance on heavy foundation models (DUSt3R~\cite{wang2024dust3r}, Sapiens~\cite{khirodkar2024sapiens}) and its per-pixel Gaussian prediction incur substantial memory overhead that grows quadratically with the number of image tokens. This limits Avat3r to at most $N{=}6$ views on a single A100/H100 GPU (80\,GB VRAM, batch size 1) before exceeding memory during training. Avat3r's reconstructions are blurry and lack high-frequency details because (1) it relies on predicted point maps that introduce geometric discontinuities (as noted in FastGHA~\cite{ji2026fastgha}), and (2) its confidence-based pruning can aggressively remove valid foreground regions.

In contrast, our architecture decouples the output Gaussian count from the input resolution and view count through the UV parameterization (discussed in Sec. 3.1 of the main paper), enabling efficient scaling to $N{=}16$ views and beyond with minimal per-view memory overhead. As shown in \cref{fig:6_view_scaling_comparison}, our method produces consistently sharper reconstructions with better-preserved identity details at every view count, including the monocular setting. We use $N{=}10$ views for the Internal10k dataset, as quality saturates beyond this point (see \cref{tab:supplementary-num-views} in the Supplementary Material and Fig.~4b in the main paper). For Ava256, we use $N{=}16$ views, as we find it provides a good trade-off of viewpoint coverage and training time.

\subsubsection{Mesh Type.}
We compare using a fixed neutral mesh versus an expression-tracked mesh as the UV parameterization substrate in~\cref{tab:supplementary-mesh-type}. The fixed neutral mesh outperforms expression-tracked meshes by a large margin across all metrics (+2.12\,dB PSNR, $-$0.088 LPIPS, $-$0.81 AKD). We attribute this to the fact that expression-tracked meshes introduce noisy per-frame vertex displacements that the model must account for, whereas a fixed neutral mesh provides a stable canonical surface that allows the Gaussian decoder to focus entirely on modeling appearance variation.

\subsubsection{Vertex Loss.}
We ablate the vertex position regularization loss ($\mathcal{L}_{\text{pos}}$) in~\cref{tab:supplementary-vertex-loss}. Although incorporating $\mathcal{L}_{\text{pos}}$ results in meaningful quantitative improvements, particularly in PSNR and LPIPS, this regularization primarily serves to stabilize early training and accelerate warm-up. The model converges to a similar visual quality without explicit vertex supervision across all metrics, demonstrating that our use of tracked meshes is a lightweight prior for enhanced training stability rather than a fundamental limitation of the proposed method.

\subsubsection{Latent and Gaussian UV Resolution.}
\cref{tab:supplementary-latent-and-uv-res} disentangles the effect of latent resolution (controlling model capacity) from Gaussian UV resolution (controlling the number of output Gaussians). Increasing the latent size from $16{\times}16$ to $128{\times}128$ yields consistent improvements across all metrics, with PSNR rising from 27.59 to 29.66\,dB. In contrast, doubling the Gaussian UV resolution from 256 to 512 at any fixed latent size provides only marginal gains. This indicates that model capacity, rather than Gaussian count, is the primary bottleneck for reconstruction quality in our architecture.

\subsubsection{Number of Target Views.}
We vary the number of target views used for supervision during training in~\cref{tab:supplementary-num-target-views}. Increasing from 1 to 8 target views improves all metrics, with PSNR rising from 28.89 to 29.54\,dB and AKD decreasing from 3.20 to 2.97. Supervising with more target views per training step enforces multi-view consistency and reduces geometric ambiguities, as the model must produce Gaussians that render correctly from diverse viewpoints simultaneously.

\subsubsection{Number of Transformer Blocks.}
\cref{tab:supplementary-transformer-blocks} ablates the number of transformer blocks in the encoder. Performance improves steadily from 2 to 8 blocks, with PSNR increasing from 28.24 to 28.89\,dB. Adding further blocks beyond 8 yields no additional improvement: the 12-block model achieves identical PSNR and LPIPS while slightly degrading AKD. We therefore use 8 blocks as the default, balancing reconstruction quality with computational cost.

\subsubsection{High-Res Finetuning: Region-Specific Losses.}
\cref{tab:supplementary-region-specific-losses} evaluates the contribution of each component in our high-resolution finetuning stage. Removing the eye-region loss degrades eye crop PSNR by 1.56\,dB, while removing the mouth-region loss reduces mouth crop PSNR by 1.97\,dB, confirming that region-specific supervision is critical for faithfully reconstructing fine details in these perceptually important areas. Removing the half-resolution loss causes the largest overall degradation ($-$3.55\,dB full-image PSNR), as it provides the coarse-to-fine gradient signal that stabilizes the high-resolution training. The full model achieves the best trade-off, with near-optimal performance across all regions.

\input{tables/6_ablation_combined}

\section{Additional Visualizations}
\label{sec:additional_visualizations}

\subsubsection{Video Visualizations.}
The \href{https://apple.github.io/ml-headsup/}{project webpage} contains interactive video results organized into the following sections:
\begin{itemize}[leftmargin=*,itemsep=2pt,topsep=4pt]
    \item \textbf{Ava-256 Validation Renders.} Per-subject reconstruction results with multi-expression renderings from novel viewpoints, comparing our method against Avat3r across different encoder view counts.
    \item \textbf{Ava-256 Comparisons.} Side-by-side rendering comparisons on held-out validation subjects, highlighting fine facial details such as pores, hair strands, and specular highlights.
    \item \textbf{Internal10K Renders.} Results on our internal multi-camera capture dataset.
    \item \textbf{Blendshape Rigging.} Interactive demonstrations of blendshape-driven deformation (neutral, smile, raised brows) showing consistent Gaussian field deformation across subjects.
\end{itemize}

\subsubsection{Novel Identity Generation.}
\begin{figure*}[t]
  \centering
  \includegraphics[width=\linewidth]{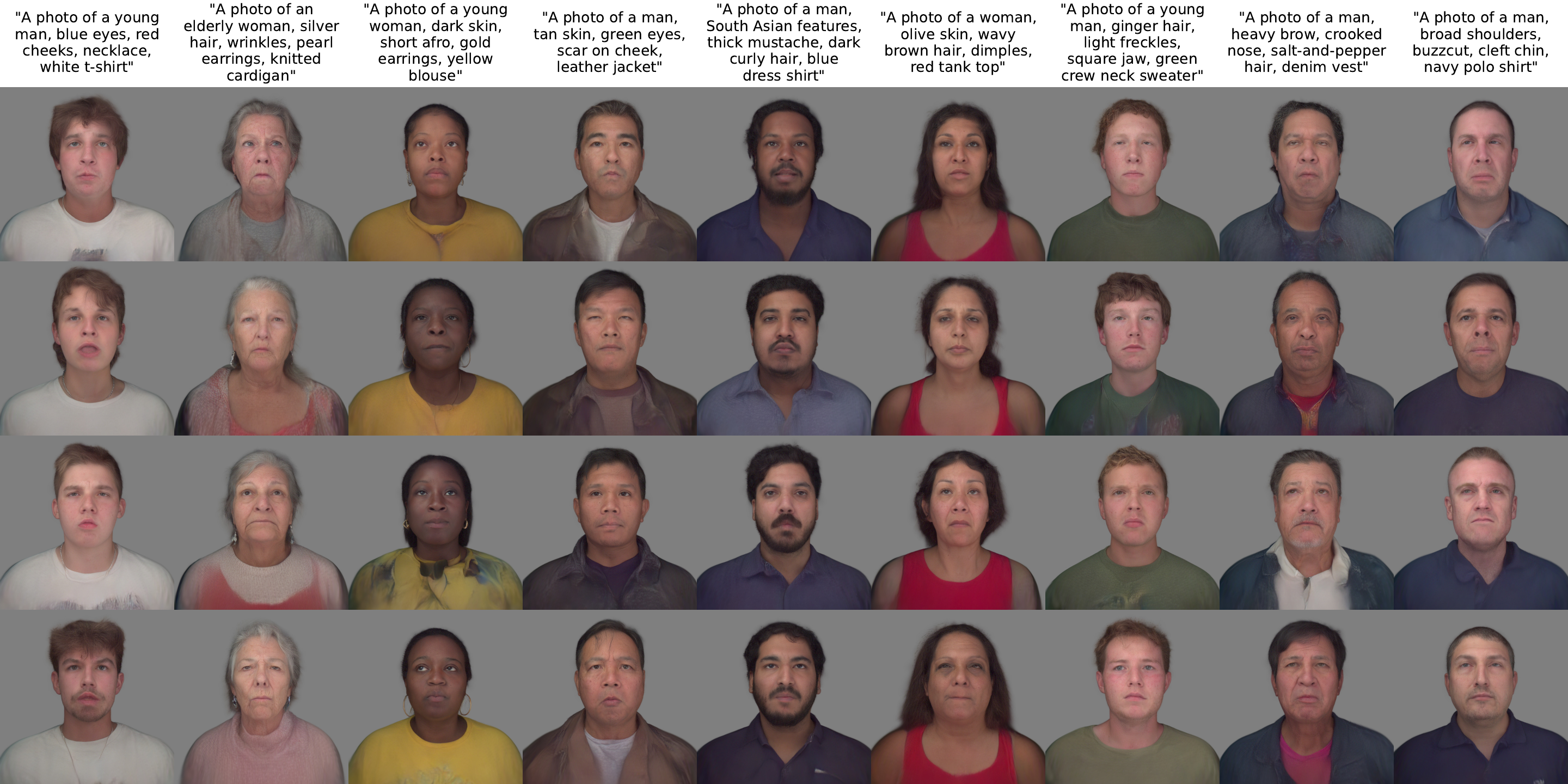}\\[2pt]
  \caption{\textbf{Additional Generated Novel Identities.} A diverse set of novel identities generated by a text-conditioned diffusion model trained on our learned head latents. Each row shows different generated subjects, demonstrating the diversity in age, gender, ethnicity, hairstyle, and facial features. These generated subjects do not exist in our training set, as confirmed by nearest-neighbor face similarity. Our supplementary webpage also provides rendered videos of these novel subjects.}
  \label{fig:supplementary_dit_samples}
\end{figure*}

In~\cref{fig:supplementary_dit_samples}, we provide more examples of generated novel identities discussed in Sec. 4.6 of the main paper.

\subsubsection{Predicted Background Gaussians.}
\begin{figure*}[t]
  \centering
  \includegraphics[width=\linewidth]{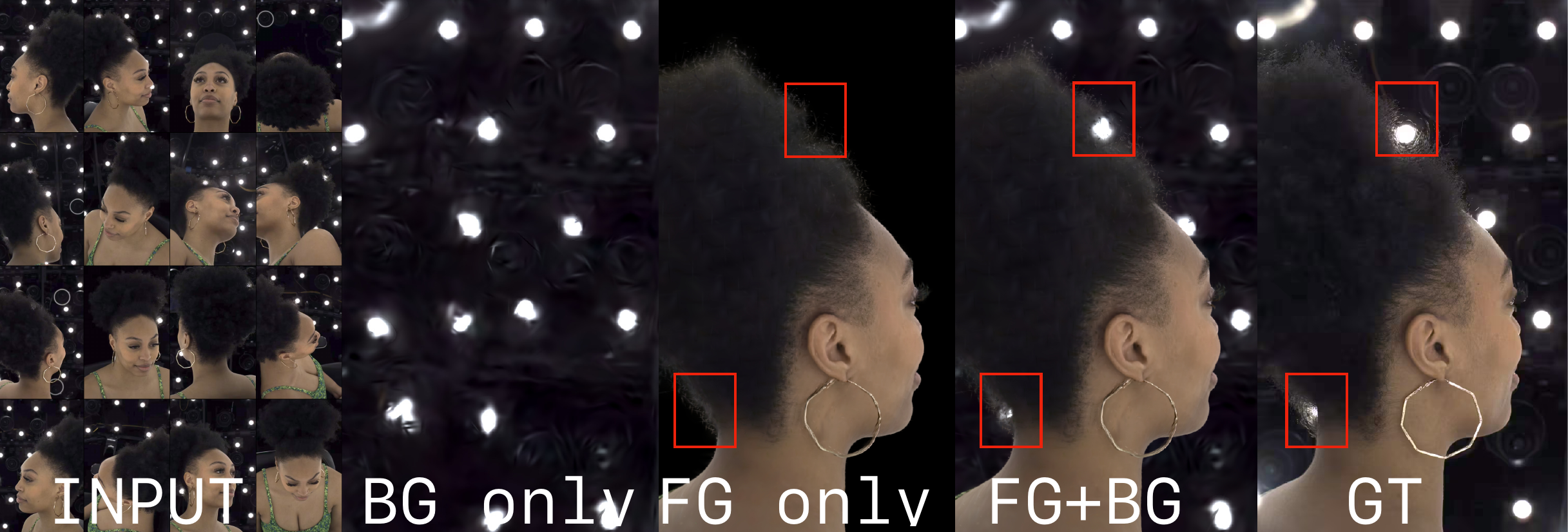}\\[2pt]
  \caption{\textbf{Background (BG) Gaussians.} Our lightweight BG model learns the rig background, eliminating the need for image matting. This auxiliary branch is discarded during inference. Notice the successful removal of rig lights from the foreground (FG).}
  \label{fig:supplementary_bg_gaussian}
\end{figure*}

We render the BG Gaussians predicted by our lightweight BG model in \cref{fig:supplementary_bg_gaussian}. Note that the capture rig's lights are modeled by BG Gaussians instead of FG Gaussians. The lightweight BG model (32-channel bottleneck) does not require any ground-truth, but it assumes a fixed-BG rig and is not designed for in-the-wild arbitrary BG removal. We aim to separate the BG during training, so that we do not rely on imprecise image matting.

\section{Implementation Details}

\subsection{Architectural Details}
We provide detailed specifications for each component of our architecture. All hyperparameters are summarized in~\cref{tab:hyperparameters}.

\subsubsection{Foreground Encoder.}

Each input image $\mathbf{I}_i \in \mathbb{R}^{3 \times H \times W}$ is patchified with a $7{\times}7$ convolution (stride~$7$) into patch embeddings of dimension $d{=}256$ (the image is first resized to be compatible with the patch size). Patch embeddings are then concatenated with $6$-dimensional Pl\"ucker ray embeddings~\cite{sitzmann2021lfns} encoding the camera geometry. A convolutional network with one downsampling stage ($2$ residual blocks at $512$ channels) followed by $4$ bottleneck residual blocks at $512$ channels maps the image patch tokens to foreground feature maps $\mathbf{F}^{\text{fg}}_i \in \mathbb{R}^{512 \times h_f \times w_f}$.
\subsubsection{Background Encoder.}
A separate, lighter convolutional encoder produces per-view background features $\mathbf{F}^{\text{bg}}_i \in \mathbb{R}^{256 \times h_f \times w_f}$, using two downsampling stages ($2$ residual blocks at $128$ channels and $2$ residual blocks at $64$ channels). These features are aggregated across views via global average pooling followed by a two-layer MLP, yielding a compact background latent $\mathbf{z}_{\text{bg}} \in \mathbb{R}^{d_{bg}}$. The background latent requires less capacity because the background is largely static across frames.

\subsubsection{Cross-Attention Transformer.}
The foreground features from all $N$ input views are flattened into a set of key-value tokens. A transformer with $8$ blocks, $8$ attention heads, hidden dimension $d_z{=}512$, and MLP dimension $1024$ maps a $2$D grid of $h_z{\times}w_z{=}64{\times}64$ learnable query tokens to the foreground latent $\mathbf{Z} \in \mathbb{R}^{512 \times 64 \times 64}$ via cross-attention. Each block applies layer normalization, multi-head cross-attention, and a feed-forward network with GELU activations. The query grid provides a fixed spatial structure that the decoder can directly reshape into a UV map, while cross-attention aggregates information from an arbitrary number of views without quadratic view-count scaling.

\subsubsection{Foreground Decoder.}
The latent $\mathbf{Z}$ is decoded into the Gaussian UV map $\mathbf{U} \in \mathbb{R}^{256 \times 256 \times 23}$ via a pre-activation residual network~\cite{he2016deep}. The decoder applies two $2{\times}$ nearest-neighbor upsampling stages with channel dimensions $[512, 256]$, each containing two pre-activation residual blocks using $3{\times}3$ convolutional kernels and learned residual branch scaling. Nearest-neighbor upsampling avoids the checkerboard artifacts of transposed convolutions. A final $3{\times}3$ convolution projects to $32$-dimensional per-texel features, from which the $23$ Gaussian attributes are regressed. Following GRM~\cite{xu2024grm}, we apply sigmoid activations for opacities and $L_2$ normalization for rotation quaternions. Position offsets use $\tanh$ scaled by $\delta_{\max}$ to bound displacements from the template mesh, scales use an exponential activation, and SH color coefficients are output directly without activation. This yields $256{\times}256 \approx 65$K foreground Gaussians.

\subsubsection{Background Decoder.}
The background latent $\mathbf{z}_{\text{bg}}$ is decoded into a UV map of $512{\times}512 \approx 262$K background Gaussians anchored to a sphere template fitted to the capture rig. Unlike the foreground decoder, the background decoder uses a residual network architecture with LeakyReLU activations and BatchNorm, progressively upsampling from $4{\times}4$ resolution through $7$ stages with channel dimensions $[512, 256, 128, 64, 32, 32, 32, 32]$. A tighter position offset bound ($\delta_{\max}{=}10\,\text{mm}$ vs.\ $200\,\text{mm}$ for foreground) constrains Gaussians near the rig geometry.

\subsection{Multi-Scale Perceptual Loss}
\label{subsec:loss_functions}
Our perceptual loss $\mathcal{L}_{\mathrm{LPIPS}}$ is implemented as a multi-scale LPIPS loss~\cite{zhang2018unreasonable} using the official pretrained AlexNet-based LPIPS network. Rather than computing the perceptual similarity at a single resolution, we evaluate it at three scales to capture both fine detail and global structure:
\begin{equation}
\mathcal{L}_{\mathrm{LPIPS}} = \sum_{k=0}^{2} \text{LPIPS}\!\left(\text{down}_{2^k}(I),\;\text{down}_{2^k}(I_{\mathrm{gt}})\right),
\end{equation}
where $\text{down}_{2^k}$ denotes $2^k{\times}$ spatial downsampling via bilinear interpolation, so the three scales correspond to the native resolution ($1{\times}$), $2{\times}$ downsampled, and $4{\times}$ downsampled.

\subsection{Downstream Applications}
Here, we provide more details on the downstream applications we discussed in Sec. 4.6 of the main paper: 

\subsubsection{Text-driven Identity Generation.}
To sample novel identities, we train a latent diffusion model that operates directly in the HeadsUp latent space. The HeadsUp encoder maps each identity to a latent tensor, which we treat as the diffusion target. Our denoising network is a DiT~\cite{peebles2023scalable} with $10$ transformer blocks, an embedding dimension of $512$, an MLP hidden dimension of $2048$. For text-conditioned generation, we encode the input prompt using a frozen Flan-T5-XXL encoder~\cite{chung2024scaling} with a maximum sequence length of $64$ tokens, followed by a token-wise linear projection from $4096$ to $512$ dimensions to match the transformer embedding size. We apply classifier-free guidance~\cite{ho2021classifierfree} by independently dropping text embeddings with probability $0.15$ and the full conditioning vector with probability $0.05$ during training.
The model is trained with SiD2 loss~\cite{hoogeboom2025simpler} (sigmoid shift $-3$) using the Adam optimizer with a learning rate of $2 \times 10^{-4}$ and a batch size of~$16$ for $300$K iterations on a single GPU. We initialize training from a pretrained checkpoint to accelerate convergence. The training data consists of HeadsUp latents extracted from our multi-view facial capture dataset, paired with automatically generated text captions describing subject appearance attributes.

At inference time, we sample from the learned distribution using a DPM solver~\cite{lu2022dpm} with $25$ denoising steps. When a text prompt is provided, classifier-free guidance steers the generation toward the described attributes. The sampled latent is then decoded by a frozen pretrained HeadsUp decoder into 3D Gaussian parameters (positions, rotations, scales, opacities, and spherical harmonics color coefficients), producing a complete head avatar that can be rendered in real time. 

\subsubsection{Blendshape-driven Latent Animation}
\label{subsubsec:blendshape_rigging}

\begin{figure*}[!h]
  \centering
  \includegraphics[width=\linewidth]{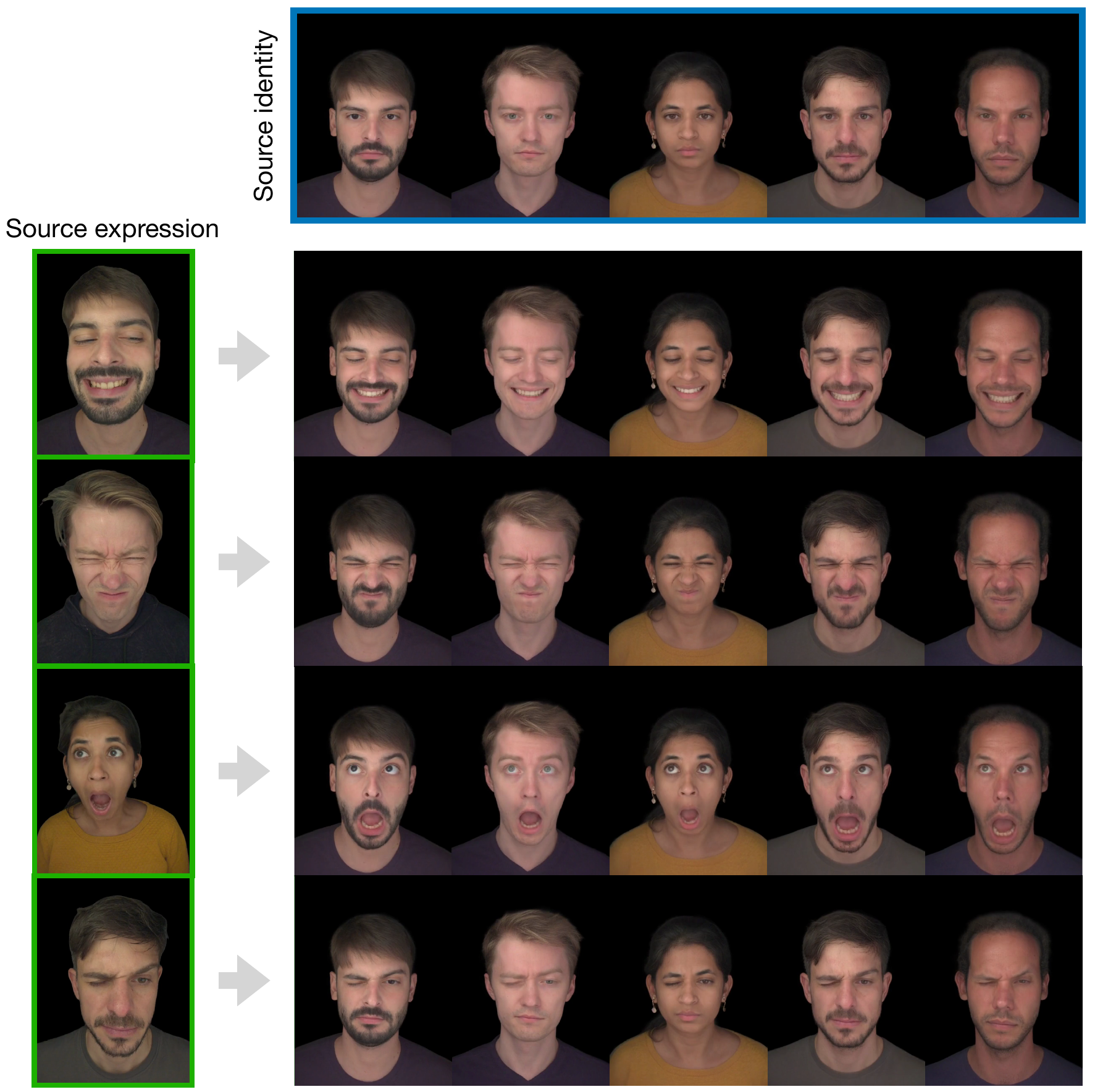}\\[2pt]
  \caption{\textbf{Blendshape rigging.} Our reconstructed Gaussian heads are riggable via blendshape parameters. We extract the source expression from the input image as blendshape coefficients to drive and animate our latent. %
  }
  \label{fig:supplementary_blendshape_rigging}
\end{figure*}

As demonstrated in the main text, HeadsUp's latent space supports temporally consistent and controllable facial animation driven by blendshape coefficients. The rigging results presented in ~\cref{fig:supplementary_blendshape_rigging} and the videos on the project webpage demonstrate that our blendshape-driven animation enables fine-grained control (e.g., eye gaze, asymmetric expressions) and identity-preserving expression transfer from reference performances. Here we explain the architecture used for this experiment.

We obtain neutral latents $Z_n$ by passing a subject's neutral-expression frames through our encoder. For the neutral head template, we use an internal UV template. However, we emphasize that our approach is completely topology-agnostic and can readily adopt the FLAME template. The $51$ blendshapes used for our animation task are derived from an internal face-tracking pipeline comparable to ARKit and FLAME. Given a neutral-expression latent and a target blendshape vector, our rigging network predicts a residual that is added to the neutral latent to produce the target expression. Each blendshape value is independently embedded using Fourier features (4 frequency bands), concatenated with a 32-dimensional learnable identifier to distinguish blendshape indices, and projected via a two-layer MLP. The resulting tokens serve as keys and values for an $8$-layer, $8$-head cross-attention transformer (hidden dimension $1024$). The queries are formed by flattening the neutral latent into $1024$ tokens. The transformer's output is then reshaped back to the spatial dimensions of the latent and added to the original neutral representation.

We train the latent animation network end-to-end by passing the predicted latent through a frozen pretrained decoder. Supervision is provided by a ground-truth target-expression latent extracted by the encoder. The training objective is a combination of an L1 loss directly on the predicted latent, L1 and multi-scale LPIPS perceptual losses on rendered images, and L1 losses on the decoded 3D Gaussian attributes (positions, colors, opacities, rotations, and scales). To preserve high-frequency details, particularly in the eyes and mouth, we employ region-specific LPIPS losses on camera-aware crops and a hinge-style adversarial loss (with a perceptual discriminator) on random $256 \times 256$ crops, which is activated after $50$K steps. The model is trained using Adam with a learning rate of $10^{-4}$ and batch size of $80$ for 200K steps on 8 A100 GPUs. 

\section{In-the-Wild Generalization}

\begin{figure*}[t]
  \centering
  
  \begin{minipage}{0.25\linewidth}
    \centering
    \small \textbf{Input}
  \end{minipage}%
  \begin{minipage}{0.75\linewidth}
    \centering
    \small \textbf{Renders}
  \end{minipage}\\[4pt] %
  
  \includegraphics[width=\linewidth]{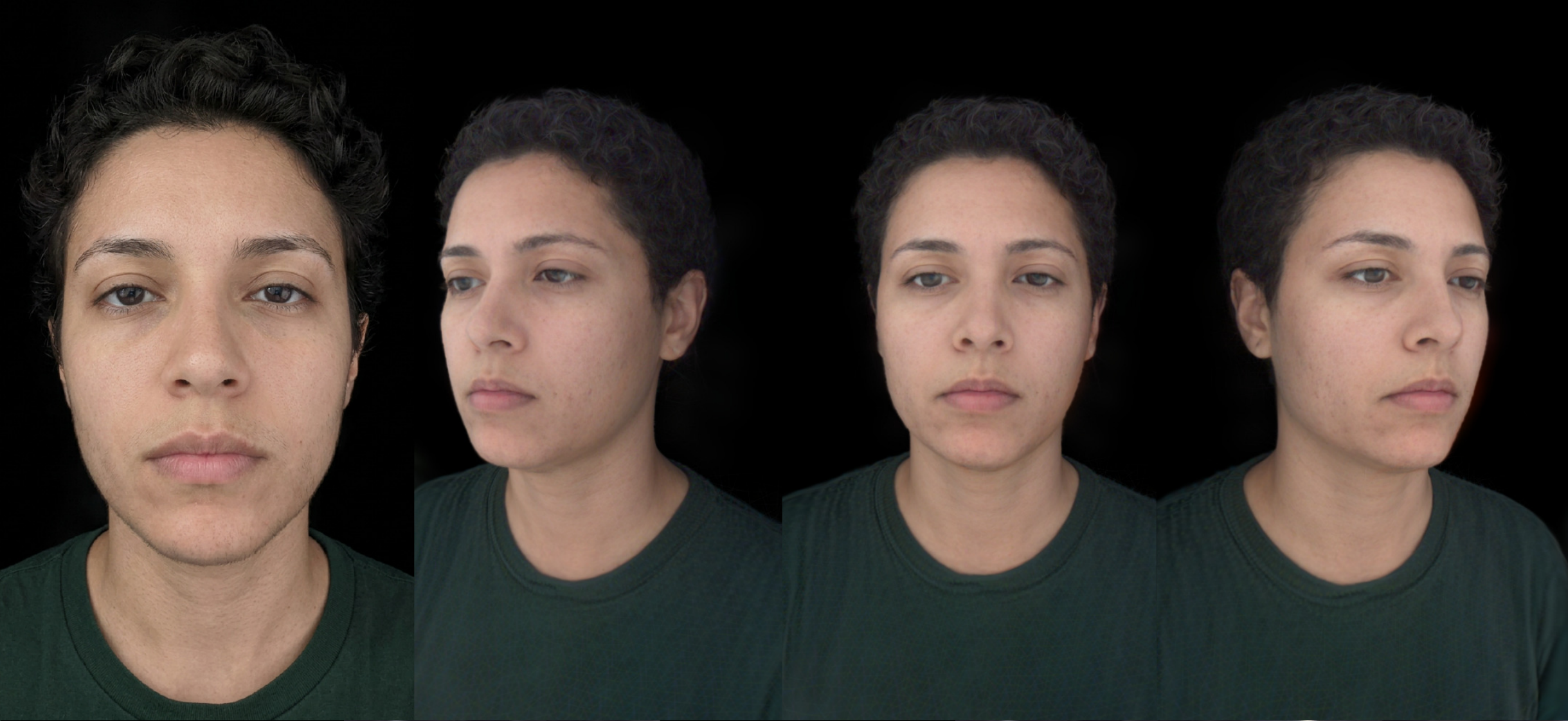}\\[2pt]
  
  \caption{\textbf{Reconstruction from single-view AI-generated image}. Despite relying on synchronized multi-view training data, our model shows early signs of generalization to novel domains. Using the monocular model variation ($N=1$), this figure demonstrates a successful 3D reconstruction derived from a single AI-generated image~\cite{googledeepmind_nanobanana_2026}, highlighting a promising avenue for future research in unconstrained settings.}
  \label{fig:supplementary_itw_monocular}
\end{figure*}

While our method is designed for and trained on calibrated studio captures, we provide an initial exploration of its in-the-wild generalization. ~\cref{fig:supplementary_itw_monocular} shows that our model can successfully reconstruct a subject in high detail from a single AI-generated image~\cite{googledeepmind_nanobanana_2026}. For this result, we use the monocular model variant ($N=1$) detailed in Section 4.5. Although the model is not yet robust to arbitrary in-the-wild data, this proof-of-concept demonstrates the potential to extend our approach to unconstrained settings in future work.

\section{HeadsUp Training Details}

\subsection{Dataset Processing}
\subsubsection{Internal10K Processing.}
We use our internal multi-view head dataset containing over \num{10000} subjects recorded with \num{16} calibrated RGB cameras. We sample \num{100} frames per subject for maximum expression diversity. We compute per-view foreground segmentation masks using an internal segmentation model.

\subsubsection{Ava-256 Processing.}
Following Avat3r~\cite{kirschstein2025avat3r}, we compute foreground matting masks for the entire dataset using BackgroundMattingV2~\cite{lin2021backgroundmattingv2} and color correct the images to non-linear sRGB.

\subsection{Viewpoint Sampling}
We select a fixed set of $N{=}10$ input cameras from the 16 available views with broad coverage of the face. For models trained or evaluated with fewer than 10 input views (e.g., Avat3r with $N{=}4$ or $N{=}6$, or ablated HeadsUp models), we use a subset of these 10 cameras selected to maximize face coverage. We prioritize frontal and near-frontal viewpoints before adding side views.

\subsubsection{Ava-256.}
For Ava-256, which provides 80 calibrated cameras, we sample $N{=}16$ input views via farthest-point sampling on camera positions to ensure maximal viewpoint diversity.

\subsubsection{Stage~1: Low-Resolution Training.}
Our model is trained on $2\times$ downsampled images at a resolution of $500 \times 375$ for $900\text{K}$ steps. We utilize a batch size of $64$ and provide $10$ input views per training sample. Optimization is performed with Adam~\cite{kingma2014adam} with a learning rate of $2 \times 10^{-4}$ and \texttt{bfloat16} mixed precision. To ensure stable initialization, we detach the gradients for the opacity and scale parameters during an initial $1\text{K}$-step warm-up phase. Furthermore, the position regularization weight, $\lambda_\mathrm{pos}$, is linearly annealed from $1.0$ to $0.01$ over the first $100\text{K}$ steps, while the silhouette loss weight, $\lambda_\mathrm{mask}$, is also annealed from $2.0$ to $0.1$ over the same interval. Finally, the adversarial loss, $\mathcal{L}_\mathrm{adv}$, is activated at $240\text{K}$ steps.

\subsubsection{Stage~2: High-Resolution Finetuning.}
We subsequently continue training at the resolution of $1000 \times 750$ for $200\text{K}$ steps, using a reduced batch size of $32$. During this phase, we introduce region-specific LPIPS perceptual losses applied to eye and mouth crops. To maintain optimization stability, the global LPIPS and discriminator losses are computed on $2\times$ downsampled renders (see Section~3.4 in the main paper). All other loss weights remain identical to those used in Stage~1. We train on the Internal10K dataset until validation metrics plateau using $16$ H100 GPUs. A comprehensive summary of all loss weights is provided in~\cref{tab:hyperparameters}.

\subsection{Ava-256 Finetuning}
Finally, we use the $4\,\text{TB}$ version of the Ava-256 dataset~\cite{martinez2024codec}, which comprises $256$ subjects, $80$ cameras, and approximately $5000$ frames per person. Following the experimental protocol established by Avat3r~\cite{kirschstein2025avat3r}, we train on $244$ subjects using $1000$ frames per subject, and evaluate our method on $12$ held-out validation subjects. We fine-tune the model that was pre-trained on the Internal10K dataset for $200\text{K}$ steps, utilizing $16$ input views sampled from the full set of $80$ cameras. This fine-tuning stage converges in less than one day using $16$ H100 GPUs.

\subsection{Evaluation Details}

\subsubsection{Internal10K.}
We evaluate on 50 held-out validation subjects with 20 frames each, using the same expression-diverse sampling as training. For each frame, we use a fixed set of 10 input views and evaluate on all remaining camera views. All metrics are computed at full resolution ($1000 \times 750$) on composite images (foreground + background). AKD is computed from 2D facial keypoints estimated by PIPNet~\cite{jin2021pixel}, and CSIM is computed from ArcFace~\cite{deng2019arcface} identity embeddings.

\subsubsection{Ava-256.}
We evaluate on 12 held-out validation subjects with approximately 2000 sampled frames in total, following Avat3r's evaluation protocol~\cite{kirschstein2025avat3r}. For each frame, we use 16 input views selected via farthest-point sampling and evaluate on all remaining cameras. All metrics are computed at full resolution ($1024 \times 667$). AKD and CSIM are computed identically to Internal10K.

\section{Potential Negative Societal Impacts}
\label{sec:negative-societal-impacts}

While our framework advances creative industries and telepresence, it presents potential risks regarding the synthesis and manipulation of photorealistic human avatars. High-fidelity 3D head reconstruction lowers the barrier for creating convincing digital humans. Specifically, our blendshape-driven latent animation enables highly controllable rigging of faces into arbitrary expressions. Although our method requires high-quality multi-view studio captures, downstream generative applications could be exploited to generate deepfakes for misinformation, fraud, or non-consensual harassment. To mitigate these risks, we advocate for robust watermarking of synthetic media and emphasize that an individual's digital likeness must only be created with explicit consent.

Regarding data ethics, our model is trained on over $10\,000$ subjects, all of whom provided written informed consent and received financial compensation. Personally identifiable information is securely managed in compliance with data protection regulations, and all individuals depicted herein explicitly consented to image reproduction. To protect privacy, we verify via nearest-neighbor face similarity that generated subjects do not replicate any training identities. Finally, we acknowledge that demographic imbalances in training data may cause asymmetrical reconstruction quality, disproportionately affecting underrepresented groups. We encourage dataset curation to mitigate such biases in future work.

\section{Hyperparameters}
\label{sec:hyperparameters}

For reproducibility, we list important hyperparameters in~\cref{tab:hyperparameters}.

\input{tables/6_hyperparameters}

\clearpage

%% file: tables/supp_avat3r_fastgha_headsup_4views.tex
\begin{table*}[b]
    \centering
    \caption{
    \textbf{Quantitative Comparison.} We compare our method against FastGHA~\cite{ji2026fastgha} and Avat3r~\cite{kirschstein2025avat3r} on the Internal10K dataset at half resolution ($500 \times 375$) using $4$ input views. Similarly to full-resolution data, our method delivers improvements across most metrics.
    }\label{tab:supp_avat3r_fastgha_headsup_4views}
    \setlength{\tabcolsep}{6pt}
    {%
    \setlength{\aboverulesep}{0pt}%
    \setlength{\belowrulesep}{0pt}%
    \renewcommand{\arraystretch}{1.2}%
    \begin{tabular}{l ccccc}
      \toprule
      \rowcolor{tableheadergray}
      & PSNR $\uparrow$ & SSIM $\uparrow$ & LPIPS $\downarrow$
      & AKD $\downarrow$ & CSIM $\uparrow$ \\
      \midrule
      Avat3r  &         23.54  & \textbf{0.819} &         0.364  &        6.50 &           0.729 \\
      FastGHA &         22.76  &         0.807  &         0.222  &        7.70 &           0.709 \\
      Ours    & \textbf{26.51} &         0.783  & \textbf{0.135} & \textbf{3.91} & \textbf{0.817} \\
      \bottomrule
    \end{tabular}%
    }
  \end{table*}

%% file: figures/supp_avat3r_fastgha_headsup_4views.tex
\begin{figure*}[!ht]
    \centering
    \setlength{\tabcolsep}{0pt}
    \renewcommand{\arraystretch}{0}
    \begin{tabular}{@{}ccccc@{}}
    Input & Avat3r & FastGHA & HeadsUp & GT \\[4pt]
    \includegraphics[width=0.201\linewidth]{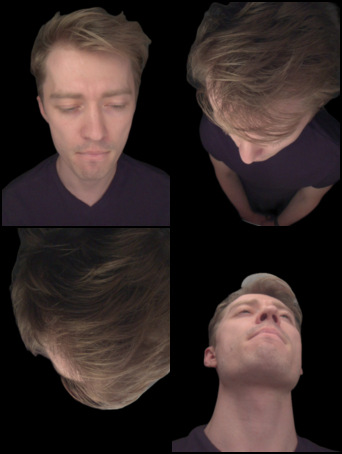} &
    \includegraphics[width=0.2\linewidth]{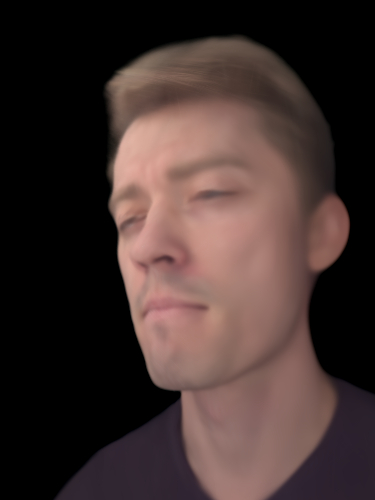} &
    \includegraphics[width=0.2\linewidth]{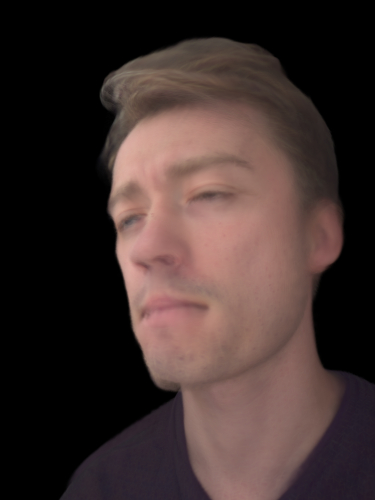} &
    \includegraphics[width=0.2\linewidth]{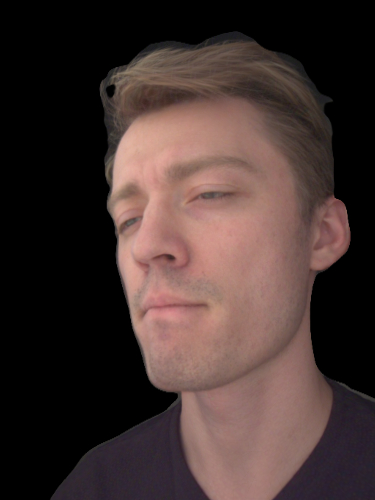} &
    \includegraphics[width=0.2\linewidth]{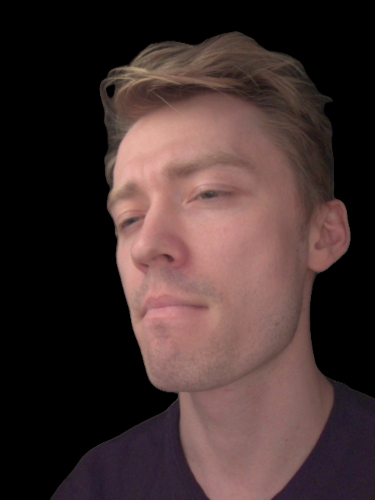} \\
    \includegraphics[width=0.2008\linewidth]{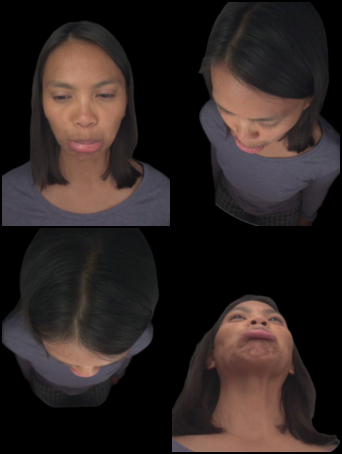} &
    \includegraphics[width=0.2\linewidth]{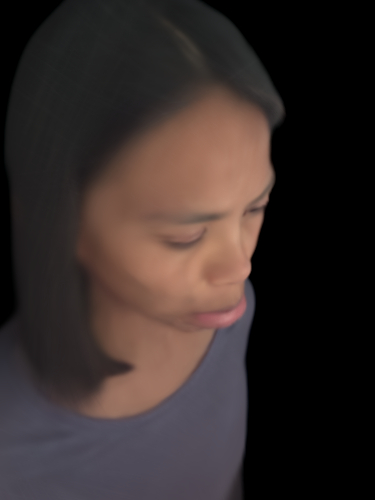} &
    \includegraphics[width=0.2\linewidth]{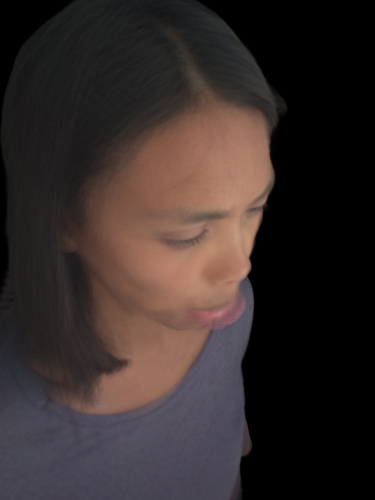} &
    \includegraphics[width=0.2\linewidth]{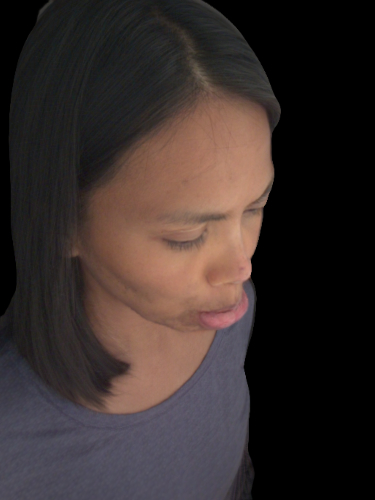} &
    \includegraphics[width=0.2\linewidth]{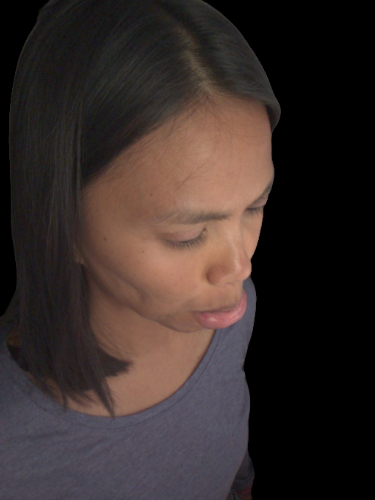} \\
    \includegraphics[width=0.2008\linewidth]{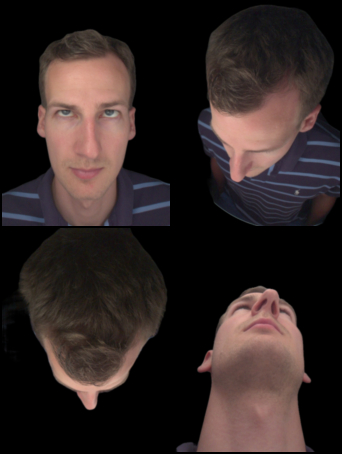} &
    \includegraphics[width=0.2\linewidth]{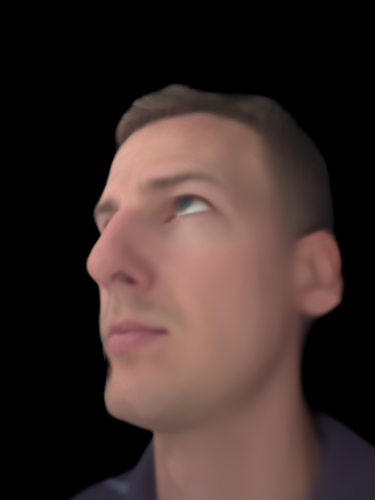} &
    \includegraphics[width=0.2\linewidth]{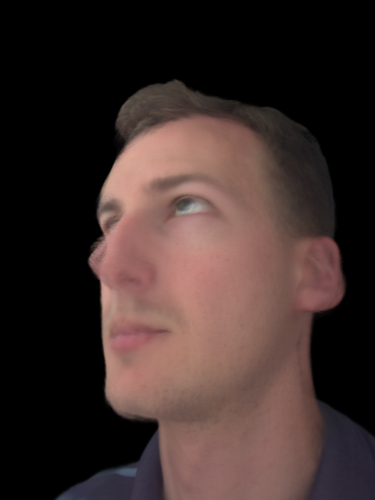} &
    \includegraphics[width=0.2\linewidth]{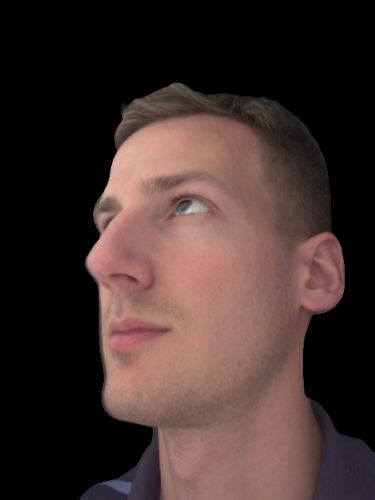} &
    \includegraphics[width=0.2\linewidth]{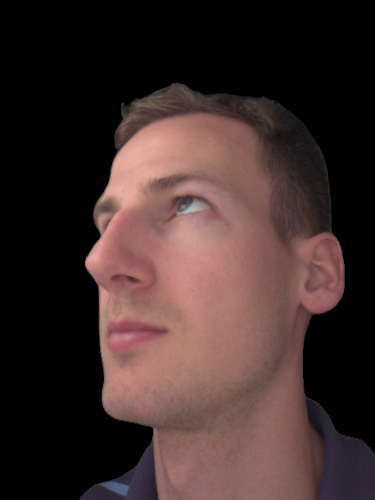} \\
    \includegraphics[width=0.2008\linewidth]{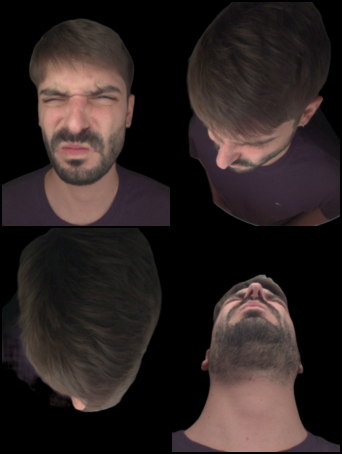} &
    \includegraphics[width=0.2\linewidth]{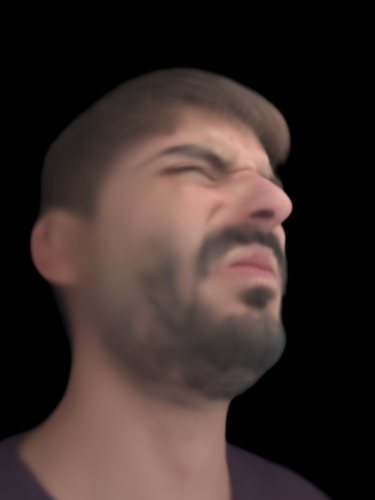} &
    \includegraphics[width=0.2\linewidth]{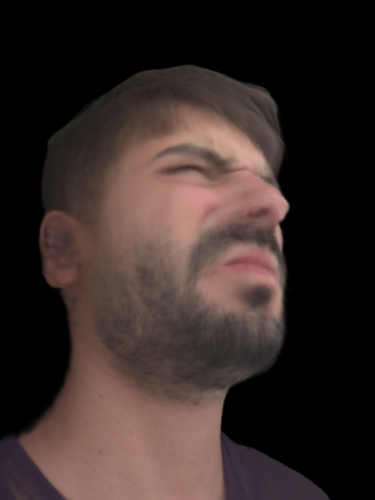} &
    \includegraphics[width=0.2\linewidth]{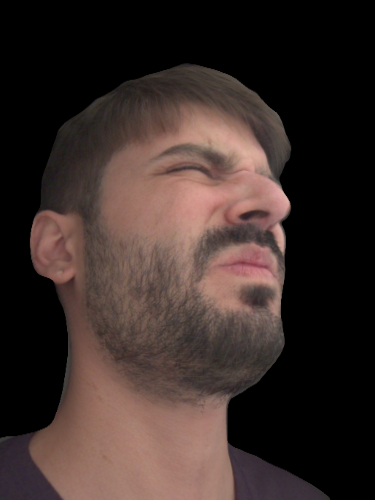} &
    \includegraphics[width=0.2\linewidth]{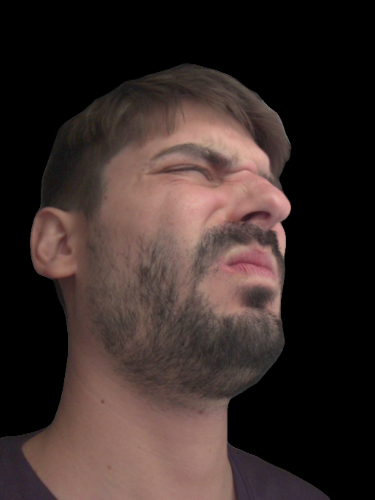} \\    
    \end{tabular}%
    \caption{\textbf{Visual Results on Internal10K (half-resolution).} 
    In our evaluation, Avat3r~\cite{kirschstein2025avat3r} produces geometrically faithful but lacks sharpness; 
    FastGHA~\cite{ji2026fastgha} improves on fine facial details but distorts appearance, most notably in mouth and eye regions. 
    In contrast, our method is able to accurately reconstruct geometry and appearance, yielding sharp renderings across subjects and facial expressions.}
    \label{fig:supp_avat3r_fastgha_headsup_4views}
  \end{figure*}

%% file: figures/6_ablation_avat3r_headsup_num_encoder_views.tex
\begin{figure*}[t]
\centering
\setlength{\tabcolsep}{0pt}
\renewcommand{\arraystretch}{0}
\footnotesize
\newcommand{\oomblock}[1]{%
\begin{tikzpicture}[inner sep=0pt, outer sep=0pt]
\node[inner sep=0pt, outer sep=0pt] (img) {\includegraphics[width=0.135\linewidth]{#1}};
\fill[black] (img.south west) rectangle (img.north east);
\node[gray, font=\bfseries\scriptsize, align=center] at (img.center) {Out of\\[4pt]Memory};
\end{tikzpicture}}
\resizebox{\linewidth}{!}{%
\begin{tabular}{@{}c@{\hspace{4pt}}ccccccc@{}}
& $N=1$ & $N=2$ & $N=4$ & $N=6$ & $N=8$ & $N=16$ & GT \\[4pt]
\rotatebox{90}{\hspace{0.2em}Avat3r} &
\includegraphics[width=0.135\linewidth]{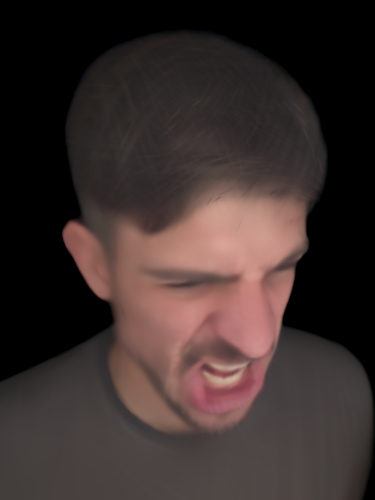} &
\includegraphics[width=0.135\linewidth]{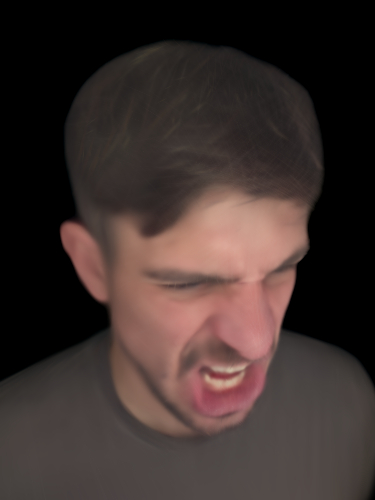} &
\includegraphics[width=0.135\linewidth]{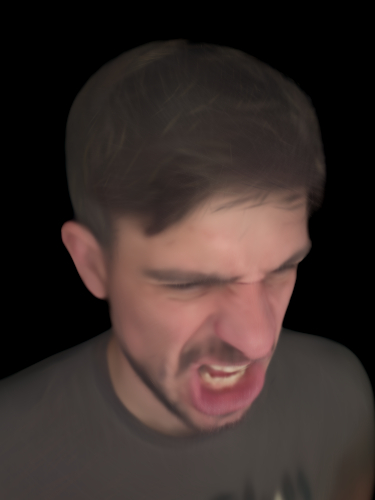} &
\includegraphics[width=0.135\linewidth]{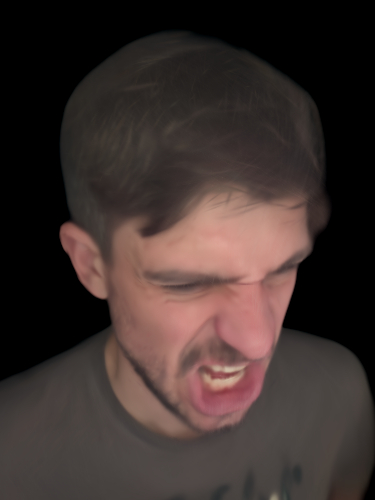} &
\oomblock{images/NumInputViewsComparison/Vittorio/N6_Avat3r_500K.jpg} &
\oomblock{images/NumInputViewsComparison/Vittorio/N6_Avat3r_500K.jpg} &
\includegraphics[width=0.135\linewidth]{images/NumInputViewsComparison/Vittorio/GT.jpg} \\
\rotatebox{90}{\hspace{0.2em}Ours} &
\includegraphics[width=0.135\linewidth]{images/NumInputViewsComparison/Vittorio/N1.jpg} &
\includegraphics[width=0.135\linewidth]{images/NumInputViewsComparison/Vittorio/N2.jpg} &
\includegraphics[width=0.135\linewidth]{images/NumInputViewsComparison/Vittorio/N4.jpg} &
\includegraphics[width=0.135\linewidth]{images/NumInputViewsComparison/Vittorio/N6.jpg} &
\includegraphics[width=0.135\linewidth]{images/NumInputViewsComparison/Vittorio/N8.jpg} &
\includegraphics[width=0.135\linewidth]{images/NumInputViewsComparison/Vittorio/N16.jpg} &
\includegraphics[width=0.135\linewidth]{images/NumInputViewsComparison/Vittorio/GT.jpg} \\[4pt]
\rotatebox{90}{\hspace{0.2em}Avat3r} &
\includegraphics[width=0.135\linewidth]{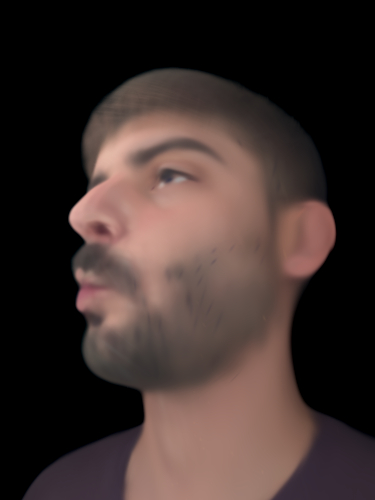} &
\includegraphics[width=0.135\linewidth]{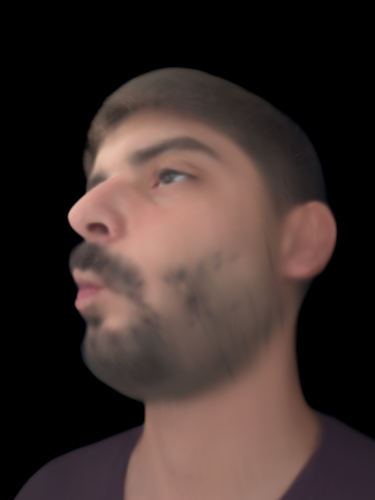} &
\includegraphics[width=0.135\linewidth]{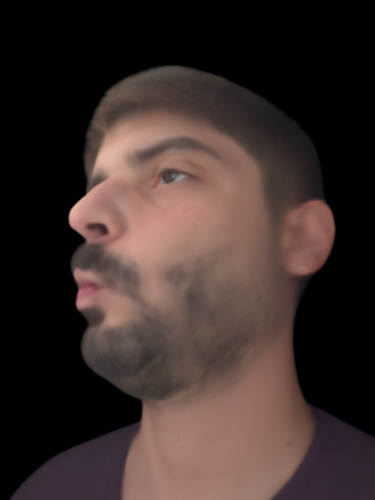} &
\includegraphics[width=0.135\linewidth]{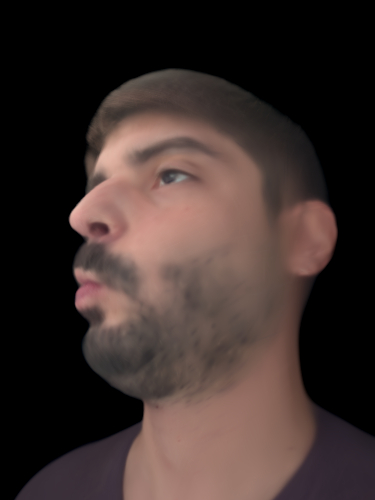} &
\oomblock{images/NumInputViewsComparison/Evan/N6_Avat3r_500K.jpg} &
\oomblock{images/NumInputViewsComparison/Evan/N6_Avat3r_500K.jpg} &
\includegraphics[width=0.135\linewidth]{images/NumInputViewsComparison/Evan/GT.jpg} \\
\rotatebox{90}{\hspace{0.2em}Ours} &
\includegraphics[width=0.135\linewidth]{images/NumInputViewsComparison/Evan/N1.jpg} &
\includegraphics[width=0.135\linewidth]{images/NumInputViewsComparison/Evan/N2.jpg} &
\includegraphics[width=0.135\linewidth]{images/NumInputViewsComparison/Evan/N4.jpg} &
\includegraphics[width=0.135\linewidth]{images/NumInputViewsComparison/Evan/N6.jpg} &
\includegraphics[width=0.135\linewidth]{images/NumInputViewsComparison/Evan/N8.jpg} &
\includegraphics[width=0.135\linewidth]{images/NumInputViewsComparison/Evan/N16.jpg} &
\includegraphics[width=0.135\linewidth]{images/NumInputViewsComparison/Evan/GT.jpg} \\
\end{tabular}}
\caption{\textbf{Ours vs.\ Avat3r: Impact of the number of input views.} Avat3r runs out of memory for $N > 6$ input views, while our method scales to 16 or more views. When comparing on the same number of input views, our method outperforms Avat3r with sharper reconstructions and better identity preservation. All methods were trained for 500K steps with 10K subjects at $500 \times 375$ resolution.
\label{fig:6_view_scaling_comparison}}
\end{figure*}

%% file: tables/6_ablation_combined.tex
\begin{table*}[p]
\centering
\caption{Ablation studies on a single-stage model trained for $500$K steps with $10$K subjects, $10$ input views, $32\times32$ latent and $256\times256$ Gaussian UV resolution unless mentioned otherwise.
\textbf{(a)}~Training data scaling: our model scales gracefully with the number of training subjects.
\textbf{(b)}~Input view scaling: quality improves with more input views, diminishing returns after 8.
\textbf{(c)}~Mesh type: a fixed neutral mesh outperforms expression-tracked meshes.
\textbf{(d)}~Vertex loss (evaluated at 360K steps): regularizing mesh vertices improves geometric consistency.
\textbf{(e)}~Model capacity: increasing the latent size yields better performance than increasing the number of Gaussians.
\textbf{(f)}~Supervision density: more target views improve geometric consistency.
\textbf{(g)}~Transformer blocks: performance saturates at 8 blocks.
\textbf{(h)}~High-resolution finetuning: region-specific losses for eyes and mouth are critical. \label{tab:supplementary-ablations}}
\footnotesize
\setlength{\tabcolsep}{4pt}
\captionsetup{skip=3pt}
\setlength{\aboverulesep}{0pt}
\setlength{\belowrulesep}{0pt}
\renewcommand{\arraystretch}{1.05}
\begin{minipage}[t]{0.48\linewidth}
\centering

\begin{subtable}[t]{\linewidth}
\centering
\caption{Number of Training Subjects \label{tab:supplementary-num-subjects}}
\vspace{-1.0em}
\label{tab:data_scaling}
\begin{tabular}{rccc}
\toprule
\rowcolor{tableheadergray} & PSNR$\uparrow$ & LPIPS$\downarrow$ & AKD$\downarrow$ \\
\midrule
250  & 22.19 & 0.267 & 6.27 \\
500  & 24.07 & 0.219 & 4.77 \\
1K   & 25.92 & 0.164 & 3.97 \\
2K   & 27.74 & 0.118 & 3.29 \\
4K   & 28.46 & 0.103 & 3.23 \\
8K   & 28.76 & 0.098 & \cellcolor{tablecellgreen}\textbf{3.16} \\
10K  & \cellcolor{tablecellgreen}\textbf{28.89} & \cellcolor{tablecellgreen}\textbf{0.096} & 3.20 \\
\bottomrule
\end{tabular}
\end{subtable}

\vspace{0.3em}

\begin{subtable}[t]{\linewidth}
\centering
\caption{Number of Input Views \label{tab:supplementary-num-views}}
\vspace{-1.0em}
\begin{tabular}{rccc}
\toprule
\rowcolor{tableheadergray} & PSNR$\uparrow$ & LPIPS$\downarrow$ & AKD$\downarrow$ \\
\midrule
1  & 22.98 & 0.208 & 5.08 \\
2  & 24.37 & 0.174 & 4.75 \\
4  & 26.51 & 0.135 & 3.91 \\
8  & 28.26 & 0.105 & \cellcolor{tablecellgreen}\textbf{3.35} \\
16 & \cellcolor{tablecellgreen}\textbf{29.03} & \cellcolor{tablecellgreen}\textbf{0.096} & 3.36 \\
\bottomrule
\end{tabular}
\end{subtable}

\vspace{0.3em}

\begin{subtable}[t]{\linewidth}
\centering
\caption{Mesh Type \label{tab:supplementary-mesh-type}}
\vspace{-1.0em}
\begin{tabular}{rccc}
\toprule
\rowcolor{tableheadergray} & PSNR$\uparrow$ & LPIPS$\downarrow$ & AKD$\downarrow$ \\
\midrule
Expr.-Tracked  & 26.77 & 0.184 & 4.01 \\
Fixed Neutral  & \cellcolor{tablecellgreen}\textbf{28.89} & \cellcolor{tablecellgreen}\textbf{0.096} & \cellcolor{tablecellgreen}\textbf{3.20} \\
\bottomrule
\end{tabular}
\end{subtable}

\vspace{0.3em}

\begin{subtable}[t]{\linewidth}
\centering
\caption{Vertex Loss \label{tab:supplementary-vertex-loss}}
\vspace{-1.0em}
\begin{tabular}{rccc}
\toprule
\rowcolor{tableheadergray} & PSNR$\uparrow$ & LPIPS$\downarrow$ & AKD$\downarrow$ \\
\midrule
w/o Vertex $\lambda$ & 27.68 & 0.114 & 3.65 \\
w/ Vertex $\lambda$  & \cellcolor{tablecellgreen}\textbf{28.46} & \cellcolor{tablecellgreen}\textbf{0.100} & \cellcolor{tablecellgreen}\textbf{3.31} \\
\bottomrule
\end{tabular}
\end{subtable}

\end{minipage}
\hfill
\begin{minipage}[t]{0.48\linewidth}
\centering

\begin{subtable}[t]{\linewidth}
\centering
\caption{Latent / Gaussian UV Resolution \label{tab:supplementary-latent-and-uv-res}}
\vspace{-1.0em}
\begin{tabular}{rccc}
\toprule
\rowcolor{tableheadergray} & PSNR$\uparrow$ & LPIPS$\downarrow$ & AKD$\downarrow$ \\
\midrule
16/256  & 27.59 & 0.121 & 3.59 \\
16/512  & 27.43 & 0.119 & 3.51 \\
32/256  & 28.89 & 0.096 & 3.20 \\
32/512  & 28.90 & 0.093 & 3.28 \\
64/256  & 29.21 & 0.089 & \cellcolor{tablecellgreen}\textbf{3.00} \\
64/512  & 29.46 & 0.086 & 3.21 \\
128/256 & 29.52 & 0.087 & 3.07 \\
128/512 & \cellcolor{tablecellgreen}\textbf{29.66} & \cellcolor{tablecellgreen}\textbf{0.083} & 3.04 \\
\bottomrule
\end{tabular}
\end{subtable}

\vspace{0.3em}

\begin{subtable}[t]{\linewidth}
\centering
\caption{Number of Target Views \label{tab:supplementary-num-target-views}}
\vspace{-1.0em}
\begin{tabular}{rccc}
\toprule
\rowcolor{tableheadergray} & PSNR$\uparrow$ & LPIPS$\downarrow$ & AKD$\downarrow$ \\
\midrule
1 & 28.89 & 0.096 & 3.20 \\
2 & 29.00 & 0.095 & 3.19 \\
4 & 29.28 & 0.094 & 3.14 \\
8 & \cellcolor{tablecellgreen}\textbf{29.54} & \cellcolor{tablecellgreen}\textbf{0.093} & \cellcolor{tablecellgreen}\textbf{2.97} \\
\bottomrule
\end{tabular}
\end{subtable}

\vspace{0.3em}

\begin{subtable}[t]{\linewidth}
\centering
\caption{Number of Transformer Blocks \label{tab:supplementary-transformer-blocks}}
\vspace{-1.0em}
\begin{tabular}{rccc}
\toprule
\rowcolor{tableheadergray} & PSNR$\uparrow$ & LPIPS$\downarrow$ & AKD$\downarrow$ \\
\midrule
2  & 28.24 & 0.102 & 3.40 \\
4  & 28.51 & 0.102 & 3.22 \\
6  & 28.80 & 0.099 & 3.27 \\
8  & \cellcolor{tablecellgreen}\textbf{28.89} & \cellcolor{tablecellgreen}\textbf{0.096} & \cellcolor{tablecellgreen}\textbf{3.20} \\
12 & \cellcolor{tablecellgreen}\textbf{28.89} & \cellcolor{tablecellgreen}\textbf{0.096} & 3.36 \\
\bottomrule
\end{tabular}
\end{subtable}

\end{minipage}

\vspace{0.3em}

\begin{subtable}[t]{\linewidth}
\centering
\caption{High-Res Finetuning: Region-Specific Losses \label{tab:supplementary-region-specific-losses}}
\vspace{-1.0em}
\label{tab:highres_finetuning}
\begin{tabular}{rcccccc}
\toprule
\rowcolor{tableheadergray} & \multicolumn{2}{c}{Full Image} & \multicolumn{2}{c}{Eye Crop} & \multicolumn{2}{c}{Mouth Crop} \\
\cmidrule(lr){2-3} \cmidrule(lr){4-5} \cmidrule(lr){6-7}
\rowcolor{tableheadergray} & PSNR$\uparrow$ & LPIPS$\downarrow$ & PSNR$\uparrow$ & LPIPS$\downarrow$ & PSNR$\uparrow$ & LPIPS$\downarrow$ \\
\midrule
w/o Eye        & 28.33 & 0.191 & 30.38 & 0.093 & \cellcolor{tablecellgreen}\textbf{32.27} & \cellcolor{tablecellgreen}\textbf{0.081} \\
w/o Mouth      & 28.20 & 0.188 & 31.78 & 0.069 & 30.17 & 0.118 \\
w/o HalfRes    & 24.80 & 0.267 & 28.26 & 0.122 & 30.54 & 0.118 \\
Full Model     & \cellcolor{tablecellgreen}\textbf{28.35} & \cellcolor{tablecellgreen}\textbf{0.190} & \cellcolor{tablecellgreen}\textbf{31.94} & \cellcolor{tablecellgreen}\textbf{0.070} & 32.14 & 0.083 \\
\bottomrule
\end{tabular}
\end{subtable}

\end{table*}

%% file: tables/6_hyperparameters.tex
{
\small
\setlength{\tabcolsep}{4pt}
\begin{longtable}{@{}l r@{}}
    \caption{\textbf{Hyperparameters.}}
    \label{tab:hyperparameters} \\
    \toprule
    \textbf{Hyperparameter} & \textbf{Value} \\
    \midrule
    \endfirsthead
    \toprule
    \textbf{Hyperparameter} & \textbf{Value} \\
    \midrule
    \endhead
    \midrule
    \multicolumn{2}{r@{}}{\textit{Continued on next page}} \\
    \endfoot
    \bottomrule
    \endlastfoot
    \multicolumn{2}{@{}l}{\textit{Input and Output}} \\[2pt]
    \quad Image resolution (Internal) & $1000 \times 750$ \\
    \quad Image resolution (Ava-256) & $1024 \times 667$ \\
    \quad Input views (Internal) & 10 \\
    \quad Input views (Ava-256) & 16 \\
    \quad Foreground Gaussians &  65K \\
    \quad Background Gaussians & 262K \\
    \midrule
    \multicolumn{2}{@{}l}{\textit{Feature Extraction}} \\[2pt]
    \quad Patch size & $7 \times 7$ \\
    \quad Patch embedding dimension $d$ & 256 \\
    \quad Foreground feature dimension $c$ & 512 \\
    \quad UV latent resolution $h_f \!\times\! w_f$ & $64 \times 64$ \\
    \quad Background feature dimension $c'$ & 256 \\
    \midrule
    \multicolumn{2}{@{}l}{\textit{Cross-Attention Transformer}} \\[2pt]
    \quad Transformer blocks & 8 \\
    \quad Attention heads & 8 \\
    \quad Hidden dimension $d_z$ & 512 \\
    \quad MLP dimension & 1024 \\
    \quad Latent resolution $h_z \!\times\! w_z$ & $64 \times 64$ \\
    \midrule
    \multicolumn{2}{@{}l}{\textit{Decoder}} \\[2pt]
    \quad Number of upsampling stages & 2 \\
    \quad Residual blocks per upsampling stage & 2 \\
    \quad Upsampling Block Channel Dimensions  & [512, 256] \\
    \quad Convolutional Kernel Size & $3 \times 3$ \\
    \midrule
    \multicolumn{2}{@{}l}{\textit{Activation Functions}} \\[2pt]
    \quad Position offsets & $\tanh \cdot \delta_{\max}$ \\
    \quad Opacities & sigmoid \\
    \quad Scales & exponential \\
    \quad Rotations & $L_2$ normalization \\
    \quad Colors (SH coefficients) & identity \\
    \midrule
    \multicolumn{2}{@{}l}{\textit{Optimization}} \\[2pt]
    \quad Optimizer & Adam \\
    \quad Learning rate & $2 \times 10^{-4}$ \\
    \quad Precision & \texttt{bfloat16} \\
    \quad Stage 1 iterations & 900K \\
    \quad Stage 1 batch size & 64 \\
    \quad Stage 2 iterations & 200K \\
    \quad Stage 2 batch size & 32 \\
    \midrule
    \multicolumn{2}{@{}l}{\textit{Losses}} \\[2pt]
    \quad Warm-up iterations (opacity\,/\,scale) & 1\,000 \\
    \quad Adversarial loss activation iteration & 240K \\
    \quad Discriminator crop size & $256 \times 256$ \\
    \quad SH degree $L$ & 1 \\
    \quad Position offset bound $\delta_{\max}$ (fg) & 200\,mm \\
    \quad Position offset bound $\delta_{\max}$ (bg) & 10\,mm \\
    \quad $\lambda_{\mathrm{L1}}$ & 1.0 \\
    \quad $\lambda_{\mathrm{LPIPS}}$ & 0.1 \\
    \quad $\lambda_{\mathrm{adv}}$ & 0.25 \\
    \quad $\lambda_{\mathrm{pos}}$ & $1.0 \to 0.01$ (linearly annealed over $100\,K$ steps) \\
    \quad $\lambda_{\mathrm{mask}}$ & $2.0 \to 0.1$ (linearly annealed over $100\,K$ steps) \\
    \quad $\lambda_{\mathrm{TV}}$ & 10.0 \\
\end{longtable}
}